# Gen-PINNs: Generative Adversarial Physics Informed Neural Networks for solving partial differential equations

Muhammad M. Akmal [a,*], Kamy Sepehrnoori [a], Michael J. Pyrcz [a,b]

[a] Hilderbrand Department of Petroleum and Geosystems Engineering, The University of Texas at Austin

[b] Department of Earth and Planetary Sciences, The University of Texas at Austin

*Corresponding author email address: muakmal@utexas.edu

## Abstract

Physics-Informed Neural Networks (PINNs) are a widely used data-free method for solving Partial Differential Equations (PDEs) using machine learning. With recent advances in Generative Adversarial Networks (GANs), adversarial learning has shown strong capabilities for modeling complex data-driven problems; however, the use of GANs in deterministic physics-informed PDE solutions remains limited. In this work, we first identify limitations of standard PINNs for solving PDEs, including spectral bias, loss imbalance, and optimizer stagnation. We then propose Generative Adversarial Physics-Informed Neural Networks (Gen-PINNs), a unified deterministic residual-adversarial framework designed to improve data-free solutions of PDEs with sharp or shock-front behavior. The generator learns the underlying PDE solution using dynamically weighted physics-informed loss components, while separate discriminators evaluate complementary PDE residual features against ideal zero-residual states. The framework further develops and adapts several methodological components, including a Fourier representation for resolving high-frequency spatial content, an orthonormal spectral diagnostic for quantifying frequency-dependent solution errors, and a modified gradient-based dynamic weighting system for physics, initial-condition, boundary-condition, and adversarial loss objectives. Gen-PINNs is tested against standard PINNs on nonlinear and higher-order PDEs, including the Burgers, Allen-Cahn, and Kuramoto-Sivashinsky equations. The results demonstrate substantial improvements in accuracy and convergence across sharp-front, stiff, and higher-order PDE solutions, highlighting the potential of deterministic residual-adversarial learning as an effective approach for solving challenging nonlinear PDEs.

*Keywords: Physics Informed Neural Networks, Generative Adversarial Networks, Partial Differential Equations, Computational Physics, Deep Learning, Scientific Machine Learning*

## 1. Introduction

The accurate and efficient modeling of Partial Differential Equations (PDEs) is a key component of scientific research and engineering design. PDEs underpin a plethora of applications in heat transfer, fluid dynamics, climate models, and numerical reservoir simulation (Bird, Stewart, & Lightfoot, 2014; Ertekin, 2005; Hansen et al., 1983). Classically, numerical solvers such as finite difference methods, finite element methods, and spectral methods are used to solve PDEs, providing robust and accurate solutions. However, the computational cost, in terms of memory and time, scales quickly in high-dimensional, non-linear, or multi-scale problems where a dense mesh or domain-specific discretization strategies are needed (Lu et al., 2021; Markidis, 2021; Sepehrnoori et al., 2020). This, along with advances in machine learning and GPU processing, has motivated many researchers over the years to investigate alternate PDE solvers (Dissanayake and Phan-Thien, 1994; Lagaris et al., 1997).

Raissi et al. (2017a, 2017b) introduce Physics Informed Neural Networks (PINNs), which have emerged as a solution-independent alternative to traditional PDE solvers. In PINNs the PDE equation is checked at observation locations and times, along with the initial and boundary conditions, which are embedded into a loss function, and using an optimizer, the best fit solution to satisfy the PDE conditions is found (Cuomo et al., 2022; Raissi et al., 2019). PINNs have been shown to perform well across various Scientific Machine Learning (SciML) applications (Hao et al., 2023; Jalili et al., 2024; Liu and Pyrcz, 2022; Morales et al., 2025; Nabian et al., 2021; Zhang et al., 2024; Zhu et al., 2022; Zucker et al., 2025). Despite the success in inference-based problems, PINNs tend to suffer from training stiffness (small changes in the network parameters cause large changes in PDE solution error) resulting in convergence issues, often resorting to trivial solutions of shock-front bearing (stiff) PDEs (Leiteritz and Pflüger, 2021; Wang et al., 2022). This typically manifests as a zero solution across the domain to satisfy the PDE and a matching solution only at known initial or boundary conditions. Moreover, PINNs tend to suffer from spectral bias, often favoring low-frequency over high-frequency solution regions (Rohrhofer et al., 2021; Tancik et al., 2020).

To overcome the limitations in the convergence of PINNs, many variations of the standard (vanilla) PINNs are proposed (Raissi et al., 2024). These PINN extensions include adaptive training methods such as self-adaptive point weighting in SA-PINNs by McClenny and Braga-Neto (2020), adaptive transfer learning in AtPINNs by Liu et al. (2023), and failure-informed adaptive methods in FI-PINNs by Gao et al. (2023). Furthermore, improved decomposition strategies include XPINNs for extended space-time areas by Jagtap and Karniadakis (2020) and FBPINNs by Dolean et al. (2024), where multilevel frequency-based decomposition shows a marked improvement over the standard PINNs method. Improved NN architectures include transformer-based approaches such as in PINNsFormer by Zhao et al. (2023), multiscale Fourier feature spatio-temporal architecture in Wang et al. (2021), and point neural PINNs by Bi and Abhayapala (2024) with strict solution enforcement of a PDE solution and Wasserstein-driven generative adversarial, GAN, PINNs with gradient penalty in WGAN-GP by Yang et al. (2018). Despite many PINN-based variants, whether adaptive in weights, network structure, or input parametrization, there is a notable lack of generalizable approaches (Torres et al., 2025). Many of the PINN extensions aim to address individual PDE modeling challenges one at a time – such as stiffness, high-frequency solution regions, and long-time solution integration – while not considering other modeling difficulties simultaneously (Nasir et al., 2025). This often results in longer training times, complex loss landscapes, poor optimizer performance, and exaggerated hyperparameters' sensitivity (Luo et al., 2025). Therefore, there is a gap in the generic modeling of PDEs using current PINN-based extensions, and a method that overcomes multiple PDE modeling challenges together is needed.

Moreover, many physical systems modeled by PDEs exhibit one or more unique structural features in the solution space. These include discontinuities, moving shock-fronts, periodic boundaries, and complex spatial-temporal oscillations (LeVeque, 1992). Capturing all the distinct features is not just an aesthetic requirement but a fundamental necessity for accurate, robust, and interpretable modeling in Scientific Machine Learning, SciML (Thiyagalingam et al., 2022). Normally, performance improvements of an approach are evaluated by benchmarking new PINNs extensions against the standard PINNs for a series of select PDEs. These PDEs typically include the Burgers equation, Allen-Cahn equation, advection-diffusion equation, Buckley-Leverett equation, and, most recently, the Kuramoto-Sivashinsky equation (Braga-Neto, 2022; Coutinho et al., 2023; Rodriguez-Torrado et al., 2022; Wang et al., 2024). Despite the continued progress, most PINN extensions tend to perform well for a particular set of such PDEs, at fixed conditions; however, they struggle to maintain accuracy when the underlying solution characteristics vary significantly. Hence, there exists a need for a more adaptable modeling paradigm where the method does not require redesigning for each new instance of a PDE, whether it is a different PDE, initial or boundary conditions, or simply coefficients within the same PDE.

In a different yet similar machine learning setting, Generative Adversarial Networks (GANs), demonstrate an excellent ability in generating realistic, complex high-dimensional data in the form of images starting from random noise inputs (Goodfellow et al., 2014). More recently, GANs were also shown to perform well in modeling many tasks, including sharp discontinuities, oscillations in target distributions, and multi-scale structures in the output space; features which are commonly encountered in PDE solutions (Karras et al., 2018; Ledig et al., 2016; Wang et al., 2018). Proposed by Yang et al. (2018), physics-informed GANs (PI-GANs) are shown to perform well for high-dimensional stochastic PDEs. A few other researchers worked on GAN-based PINN extensions, including developing Physics-Informed Enhanced Super-Resolution GANs (PIESRGANs), by Bode et al. (2021), quantifying uncertainty in physics-informed Wasserstein GANs (WGANs) by Gao and Ng (2022), and using a hybrid physics-informed-data-driven approach to train a GAN model by Ciftci and Hackl (2024). Despite the performance improvements of using GANs with a physics-based loss, research in this area remains scarce. Nonetheless, GAN-based PINN extensions represent a potential direction for general-purpose SciML modeling for PDEs.

In this work, we first illustrate the limitations of standard PINNs for modeling shock-front bearing PDEs. These include spectral bias towards resolving low-frequency solution modes, complex loss landscapes with constant loss-weight multipliers, poor convergence with 1st order optimizers, and limited sensitivity of the mean square error to heterogeneous error distributions Building on this analysis, we propose a multi-physics adversarial framework where a deterministic coordinate conditioned generator approximates the PDE solution while multiple discriminators evaluate complementary multi-physics residual features against their ideal zero-residual states.  Several supporting mechanism adapted from concepts introduced in prior PINN extensions are reformulated into the Gen-PINNs framework to address specific representation and optimization pathologies including Fourier input feature embeddings, dynamic loss weights, and staged optimization. We also examine multiple formulations of the training loss function and develop a strategy tailored for generator-discriminator  training algorithms for solving PDEs. The resulting  Generative Adversarial Physics-Informed Neural Network (Gen-PINNs) provides a flexible framework for deterministic PDE solution while integrating physical constraints of standard PINNs. Finally, we demonstrate the improved accuracy of the Gen-PINNs approach against standard PINNs across non-linear and higher order PDE configurations, demonstrating improved accuracy and convergence in multiple challenging solution regimes

The paper is structured in the following way. In Section 2, we summarize standard PINNs going over fundamental information needed for the proceeding analysis. We then use a benchmark PDE to examine several limitations of standard PINNs, including spectral bias, imbalance between the loss terms, and optimizer stagnation. We also develop Dirichlet-compatible spectral-error and per-mode convergence diagnostics to track the representation and learning of individual solution frequencies during training.  In Section 3, we describe the methodology used in the Gen-PINNs, including the deterministic residual adversarial approach and the representation and optimization components used with the framework, and discuss the relevant mathematical implications. In Section 4, we discuss the training and optimization approach used for solving various PDEs, including a structured approach for selecting components and hyperparameter tuning in Gen-PINNs summarized in Algorithm. 1. and apply the Gen-PINNs method against a series of PDEs with increasing solution complexity, examine its performance across non-linear, stiff and high order PDEs and compare the resulting solutions with standard PINNs. Finally, in Section 5, we summarize the findings of the work and discuss the limitations and potential opportunities of future research for residual adversarial physics informed methods.

## 2. Physics Informed Neural Networks (PINNs)

Introduced by Raissi et al. (2017a, 2017b), PINNS are the first SciML method to solve PDEs using a machine learning approach with only the initial, boundary conditions, and the PDE residual as the training data, without using a numerical solution as prior information. In this problem setup, we use PINNs for inference of PDE solutions as illustrated in Raisssi et al. (2019). The PDE residual is calculated at collocation points, and the initial and boundary losses are calculated at set points and then are embedded into Neural Networks (NN) as an optimization loss function to satisfy the governing equation and specific PDE conditions, respectively.

### 2.1. Problem Description

Use of standard PINNs to solve PDEs is described as follows, based on the approach in Raissi et al. (2017a). Consider a spatiotemporal domain denoted by,

$$\Omega \subset \mathbb{R}^d \times [0, T] \tag{1}$$

with spatial and temporal coordinates given by $x \in x\mathbb{R}^d$ and $t \in [0, T]$ and consider a PDE of the form,

$$\mathcal{N}[u(x,t); \lambda] = 0, (x,t) \in \Omega, \tag{2}$$

where $\mathcal{N}[\cdot]$ is a linear or a non-linear differential operator which is parametrized by $\lambda$ and $u: \Omega \to \mathbb{R}^m$ is the unknown solution field in space and time. The PDE is then subject to the initial conditions (ICs) and boundary conditions (BCs),

$$u(x, 0) = u_0(x), x \in \Omega_x, \tag{3}$$

$$\mathcal{B}[u(x,t)] = 0, (x,t) \in \partial\Omega_x \times [0, T], \tag{4}$$

where $\mathcal{B}[\cdot]$ represents the boundary operator, including Dirichlet, Neumann, Robin, and Periodic boundary conditions and $\Omega_x$ represents the spatial domain of the PDE. Consisting of deep neural networks, PINNs take $(x, t)$ as input and approximate the actual solution $u(x, t)$, as the NN output, $u_\theta(x, t)$, by training learnable parameters (weights and biases) $\theta$.

This is achieved by using automatic differentiation where the derivative of $u_\theta$ is calculated with respect to $x$ and $t$ and the respective values for the different derivatives are entered into the PDE operator, $\mathcal{N}[\cdot, \lambda]$, yielding the PDE residual function,

$$f_\theta(x,t) \coloneqq \mathcal{N}[u_\theta(x,t); \lambda] \tag{5}$$

The training objective function then minimizes a composite that is a function of the PDE loss and the initial and boundary conditions loss as a Mean Squared Error (MSE). The initial conditions loss is calculated as,

$$\mathcal{L}_{\mathrm{IC}} = \frac{1}{N_{\mathrm{IC}}} \sum_{i=1}^{N_{\mathrm{IC}}} \left| u_\theta\left(x_{\mathrm{IC}}^i, t_{\mathrm{IC}}^i\right) - u_{\mathrm{IC}}^i \right|^2 \tag{6}$$

where $\left(x_{\mathrm{IC}}^i, t_{\mathrm{IC}}^i\right)$ and $u_{\mathrm{IC}}^i$ are the coordinates and target values at initial conditions. The boundary conditions loss is calculated as,

$$\mathcal{L}_{\mathrm{BC}} = \frac{1}{N_{\mathrm{BC}}} \sum_{i=1}^{N_{\mathrm{BC}}} \left| \mathcal{B}\left[u_\theta\left(x_{\mathrm{BC}}^i, t_{\mathrm{BC}}^i\right)\right] - u_{\mathrm{BC}}^i \right|^2 \tag{7}$$

where $\mathcal{B}[\cdot]$ represents the boundary operator and $\left(x_{\mathrm{BC}}^i, t_{\mathrm{BC}}^i\right)$ and $u_{\mathrm{BC}}^i$ are the coordinates and target values at the boundary conditions. For the special case of periodic boundary conditions, $\mathcal{B}[\cdot]$ is used to impose equivalence of solution and/or the derivative(s) of the solution at opposing boundaries such that,

$$u_\theta(x_a, t) = u_\theta(x_b, t) \tag{8}$$

$$\frac{\partial u_\theta}{\partial x}(x_a, t) = \frac{\partial u_\theta}{\partial x}(x_b, t) \tag{9}$$

hold where $x_a$ and $x_b$ denote spatial domain endpoints. This is a commonly occurring condition for problems with function or flux value continuity requirement at endpoints.

Next, the PDE residual loss is calculated at selected collocation points, $(x_f^i, t_f^i)$, as

$$\mathcal{L}_{\mathrm{PDE}} = \frac{1}{N_f} \sum_{i=1}^{N_f} \left| f_\theta\left(x_f^i, t_f^i\right) \right|^2 \tag{10}$$

where $N_f$ is the number of collocation points. Finally, the total optimization loss is calculated as a weighted sum of the losses as

$$\mathcal{L} = \lambda_{\mathrm{IC}} \mathcal{L}_{\mathrm{IC}} + \lambda_{\mathrm{BC}} \mathcal{L}_{\mathrm{BC}} + \lambda_{\mathrm{PDE}} \mathcal{L}_{\mathrm{PDE}} \tag{11}$$

where $\lambda_{\mathrm{IC}},\ \lambda_{\mathrm{BC}},\ \lambda_{\mathrm{PDE}}$ are the weight hyperparameters used to control the influence of the initial, boundary, and PDE losses respectively. Many PINN extensions specifically aim at improving the loss function, including the use of adaptive points sampling, adaptive hyperparameter weighting, and, more recently, curriculum learning PINN variants (Dwivedi and Srinivasan, 2020; Gao et al., 2025; Kharazmi et al., 2019; Monaco and Apiletti, 2023; Skukla et al., 2021; Wang et al., 2024). Again, despite the significant progress over standard PINNs in terms of improved convergence and testing accuracy, current PINN extensions are task specific. There is indeed a lack of general frameworks that are capable of simultaneously modeling different PDE training challenges. Moreover, standard PINNs as described in Raissi et al. (2017a) often fail to converge with unstable optimization loss landscapes for PDEs containing higher-order derivatives (Basir and Senocak, 2022; Song et al., 2024). This often leads to either very long training times, rendering the method impractical for real-world application, or resorting to a near-trivial solution that only satisfies the PDE at initial and boundary conditions and constant values elsewhere (Leiteritz and Pflüger, 2021). This is investigated next in detail.

## 2.2. Standard PINNs limitations

To illustrate the inherent limitations of standard PINNs we look at an example of the one-dimensional Burgers' PDE (Eq. 12-14). The non-linear PDE with a shock front is used to identify the main factors hindering fast and reliable convergence of standard PINNs,

$$\frac{\partial u}{\partial t} + u\frac{\partial u}{\partial x} = \nu\frac{\partial^2 u}{\partial x^2};\ t \in [0,1], x \in [-1,1]; \nu = 0.01 \tag{12}$$

$$u(x,0) = -\sin(\pi x) \tag{13}$$

$$u(-1,t) = u(1,t) = 0 \tag{14}$$

The PINNs model approximates the solution $u(x,t)$ of the PDE using a deep neural network that outputs with trainable parameters, $\theta$, where the output is denoted by $u_\theta(x,t)$. The NN is implemented using JAX as a fully connected NN with 4 hidden layers, and each layer consists of 128 neurons (Bradbury et al., 2018). The hyperbolic tangent (tanh) is used as the activation function. A value of 1 is used for the ICs, BCs, and PDEs loss weighting to avoid any bias loss term. The domain is discretized by $N_x \times N_t$ where $N_x = 256$ , while $N_t = 100$. For the ICs and BCs, 500 points each are randomly selected at the edges where data is known. For the PDE, the entire 25600 uniform grid mesh is used as the collocation points to better capture the gradient change intricacies. The equation is next solved using the fully implicit backward Euler method time-stepping scheme, along with central differencing in space, which is unconditionally stable. The resulting non-linear system is solved using Newton-Raphson iterations forming the numerical solution $u_{ref}(x,t)$ . The general equation for the semi-discrete system is given by Eq. 5 where $L$ is the linear operator, $\Delta t$ is the time step, and $N(\cdot)$ is the non-linear term corresponding to standard fully implicit backward Euler method. To evaluate the accuracy of the PINNs solution, the relative $L^2$ error, $\varepsilon_{L^2}$, is calculated using Eq. 16 at all grid points, matching the discretization of the numerical solution,

$$\frac{u(t+\Delta t) - u(t)}{\Delta t} = Lu(t+\Delta t) + N(u(t+\Delta t)) \tag{15}$$

$$\varepsilon_{L^2} = \frac{\left(\sum_{i=1}^{N_x} \sum_{j=1}^{N_t} \left|u_\theta(x_i,t_j) - u_{ref}(x_i,t_j)\right|^2\right)^{1/2}}{\left(\sum_{i=1}^{N_x} \sum_{j=1}^{N_t} \left|u_{ref}(x_i,t_j)\right|^2\right)^{1/2}} \tag{16}$$

Next, using the ADAM optimizer with a learning rate of $1 \times 10^{-3}$ the PINNs model is trained for 500 epochs (Kingma and Ba, 2014). The number of epochs is limited to investigate factors constraining the PINNs model convergence. As shown in Figure 1, the convection-dominated solution forms a narrow viscous shock-front layer centered at $x = 0$ which is roughly captured by the PINNs model. However, the standard PINNs model struggles to capture the smooth trend on either side of the transition zone, leading to a large average relative $L^2$ error of 35.41%, where the highest percentage error of 59.6% is observed at $x = -0.035$, $t = 0.515$. The line graph trend in Figure 2 shows where the PINNs model fails to capture the trend, mainly at inflection points near $x = 0$. Unfortunately, this is a common occurrence in PDE modeling using standard PINNs, where suboptimal performance is observed when capturing shock fronts across a plethora of PDEs as documented in Abbasi et al. (2025).

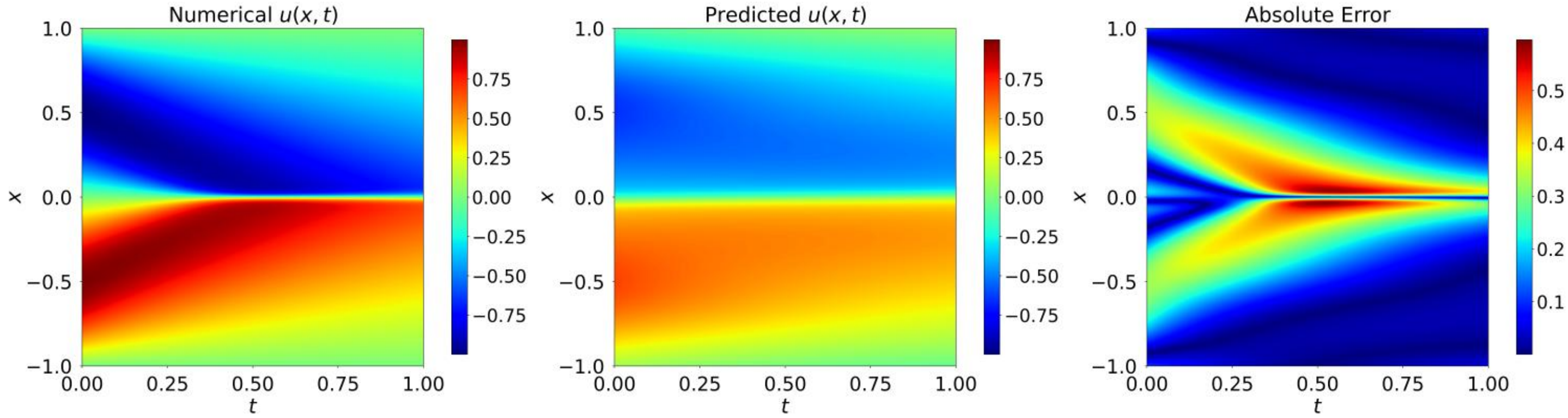

*Figure 1: Left: Numerical solution, Center: PINNs prediction, Right: Absolute error in PINN prediction of u(x,t) for the Burgers' PDE. A $L^2$-relative error of 35.41% is observed.*

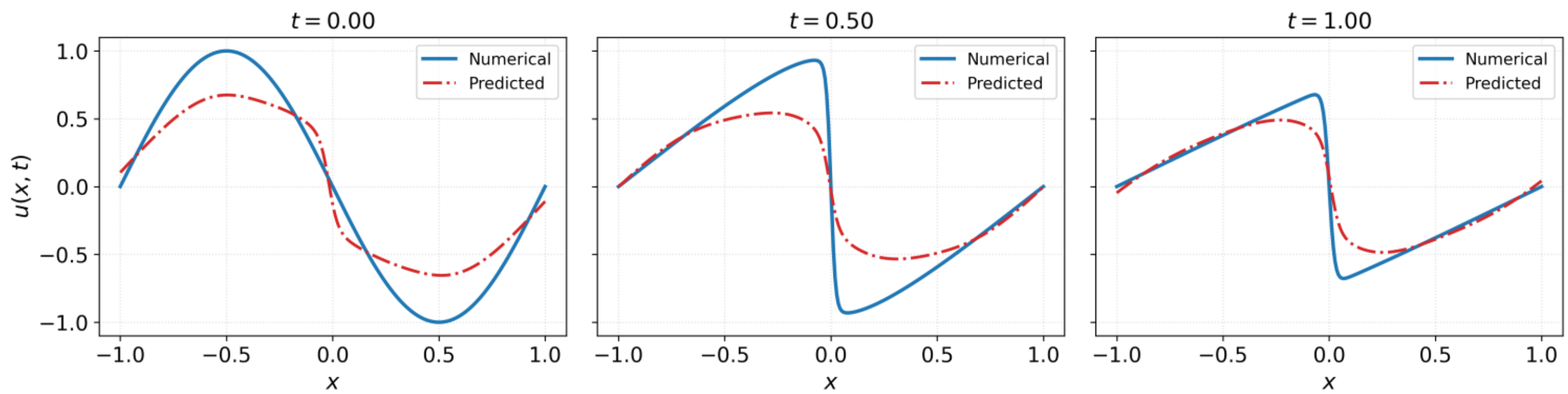

*Figure 2: Left: t=0, Center: t=0.5, Right: t=1 line plot comparison of the numerical solution (blue) and PINNs prediction (red) for the Burgers' PDE*

There are several potential causes for the slow convergence of standard PINNs when modeling shock-front bearing stiff-PDE solutions such as the Burgers' equation (Mishra and Moseley, 2024). Firstly, a poor representation of higher frequency modes in the solution space leads to a spectral bias towards learning lower frequency modes (relatively long-term phenomenon) first, as opposed to sharp, and in many cases, repeated gradient changes in time (Moseley et al., 2023; Rahaman et al., 2019; Wang et al., 2021). This primarily stems from the implicit regularization effect in gradient descent which favors the resolution of less complex low-frequency errors first, and the limited real data at the domain edges, ICs, and BCs, which lacks the ability to overcome this pathology in NN training. Second, as the standard PINNs loss function is a constant unscaled addition of losses, including the IC vs PDE residual loss, gradient imbalance of the losses tends to occur as each term in the loss function decays at an inherently different rate (Basir and Senocak, 2022a). Furthermore, higher-order PDEs tend to ill-condition the loss landscape and first-order optimizers, such as ADAM, may fail to converge sufficiently in time and suffer from stability issues (Rathore et al., 2024; C. Song et al., 2024; Wang et al., 2022). Lastly, the standard PINNs approach uses an MSE loss that does not consider the variability in the solution space and penalizes collocation points equally regardless of the local error distribution (Yang et al., 2021). To better understand these training pathologies that are a potential cause of slow convergence in PINNs, we look at the posited limitations in more detail.

**Standard PINNs suffer from representation pathologies:** To determine representation limitations in conventional Neural Networks, known as Multi-Layer Perceptron (MLP), Jacot et al. (2018) propose the Neural Tangent Kernel (NTK) approach, which shows that for an infinite-width limit NN, a NN with an infinite number of neurons in the hidden layers, gradient descent essentially resolves to a kernel gradient descent with a fixed architecture-dependent kernel. The NTK then remains almost constant during training, and the errors associated with the largest eigenfunctions decay faster than those of the lower eigenfunctions. This results in low-frequency spectral bias where NNs tend to prioritize learning less complex low-frequency features first before potentially resolving complex high-frequency features during training (Cao et al., 2019; Mustajab et al., 2024; Qin et al., 2024; Rahaman et al., 2019; Xu et al., 2019). Furthermore, as investigated by Wang et al. (2021), when applied to PINNs, the NTK essentially becomes a fixed deterministic kernel such that the training error follows the eigenfunction as well. Each term in the loss function decays at a rate set by the corresponding eigenvalue. This not only indicates a different convergence rate for each frequency mode but also a disparate rate of convergence for each loss term. This contributes to the poor convergence in standard PINNs when modeling PDEs with multi-scale solutions.

To consider the role of spectral bias in standard PINNs, we look at the error distribution across frequencies after training. Qin et al. (2024) implement an energy spectral error analysis for a periodic BC PDE while recent work groups spectral power over wavenumber bins comparing errors across different spatial scales using line plots (Chakraborty et al., 2026). Using a Discrete Sine Transform (DST-1), we extend the approach for the Dirichlet BCs in the current Burgers' equation and implement a relative spectral error method with the SciPy scientific and technical computing Python package (Virtanen et al., 2020). To apply the approach to a Dirichlet BCs PDE, suppose $u_{\mathrm{ref}}(x,t)$ is the reference numerical solution and $u_\theta(x,t)$ is the standard PINNs model prediction. The grid is then uniformly divided into $x_i = x_{\min} + i\Delta x, i = 0, \dots, N_x - 1$, and time indices $t_j = t_{\min} + j\Delta t, j = 0, \dots, N_t - 1$. As Dirichlet BCs apply, $u(x_{\min}, t) = u(x_{\max}, t) = 0$ and we work on the interior points $i = 1, \dots, M$ where $M = N_x - 2$ and define the error field as

$$e(x,t) = u_\theta(x,t) - u_{\mathrm{ref}}(x,t) \tag{17}$$

The Laplacian eigenfunctions on $[x_{\min}, x_{\max}]$ with homogeneous Dirichlet BCs is calculated as

$$\phi_n(x) = \sin\left(\frac{n\pi}{L}(x - x_{\min})\right), k_n = \frac{n\pi}{L}, n = 1,2,3, \dots, \tag{18}$$

where $L = x_{\max} - x_{\min}$. Next, the fields are expanded along the orthonormal DST-1 basis on the interior points (avoiding ICs and BCs). Then for a spatial signal $v_i = v(x_i, t_j)$ its sine coefficients are

$$\hat{v}_n(t_j) = \mathrm{DST} - \mathrm{I}_{\mathrm{ortho}}\left[v_{1:M}(t_j)\right], n = 1, \dots, M \tag{19}$$

As DST-1 is orthonormal, its discrete sine transform matrix $S$ satisfies the $S^T S = I$ and the Parseval's Identity holds exactly on the interior points as

$$\sum_{i=1}^{M} \left|v_i(t_j)\right|^2 = \sum_{n=1}^{M} \left|\hat{v}_n(t_j)\right|^2 \tag{20}$$

$$\hat{v}(t_j) = S v(t_j) \tag{21}$$

where the spatial $L^2$ energy in $v(\cdot, t_j)$ equals the sum of squared DST modal coefficients (model energy) for the same time. Next, we apply the same orthonormal transform to the prediction vector (Eq. 22), the reference vector (Eq. 23) and the error vector $e$ defined earlier (Eq. 24) as,

$$\hat{u}_\theta(t_j) = Su_\theta(t_j) \tag{22}$$

$$\hat{u}_{\text{ref}}(t_j) = Su_{\text{ref}}(t_j) \tag{23}$$

$$\hat{e}(t_j) = Se(t_j) = \hat{u}_\theta(t_j) - \hat{u}_{\text{ref}}(t_j). \tag{24}$$

The per-mode energy for the error and the reference solution are aggregated over time, and the average is taken respectively as,

$$E_{\text{err}}(n) = \frac{1}{N_t}\sum_j \left|\hat{e}_n(t_j)\right|^2 \tag{25}$$

$$E_{\text{ref}}(n) = \frac{1}{N_t}\sum_j \left|\hat{u}_{\text{ref},n}(t_j)\right|^2 \tag{26}$$

To calculate the relative spectral error ratio, we first bin nearby wavenumber groups allowing for a smoother and more interpretable graph. The (angular) wavenumber $k_n$ for the corresponding Dirichlet sine mode is calculated as,

$$k_n = \frac{n\pi}{L}, N = 1,2, \dots, M \tag{27}$$

Next, let $\{\mathcal{B}_b\}_{b=1}^{B}$ be the collection of disjoint wavenumber bins portioning the index set $N$. The representative wavenumber for each logarithmic bin is then calculated as a geometric mean of the bin edges as,

$$k_{\text{center}}(\mathcal{B}_b) = \exp\left(\frac{1}{2}[\ln k_{min}(\mathcal{B}_b) + \ln k_{max}(\mathcal{B}_b)]\right) \tag{28}$$

Finally, for each bin $\mathcal{B}_b$, we define the log-binned energy-weighted relative spectrum as the ratio of the sum of the model energies in Eq. 27 and is then plotted against $k_{\text{center}}$ to obtain the relative error spectrum vs wavenumber in Fig. 3 for the 1-D Burgers' PDE,

$$R_{\text{bin}}(\mathcal{B}_b) = \frac{\sum_{n\in\mathcal{B}_b} E_{\text{err}}(n)}{\sum_{n\in\mathcal{B}_b} E_{\text{ref}}(n)} \tag{29}$$

As confirmed from the plot, the standard PINNs model possesses a low-to-high learning signature such that as $k$ increases, $R_{\text{bin}}$ also increases showing that the fraction of missed energy grows with wavenumber as also indicated by other diagnostic methods such as the F-principle (Rahaman et al., 2018; Xu et al. 2019). Therefore, the standard PINNs approach prioritizes learning lower frequency features compared to higher frequency features first. This inherent spectral bias becomes one of the main factors of poor convergence of standard PINNs in modeling complex details with high $k$. As documented by the NTK method, standard PINNs focus on resolving errors along eigenfunctions with larger eigenvalues (lower $k$) before errors with lower eigenvalues (higher $k$)(Wang et al., 2021).

**Standard PINNs struggle to converge to an ideal solution:** Second, as the standard PINNs' combined loss term with constant loss multipliers is unscaled, the sum of ICs, BCs, and PDE loss terms struggles to converge simultaneously. Each term consists of a different unit, derivative order and is sampled in a different density. This leads to different loss gradients for each term, hampering the optimizer from smoothly converging all terms together for a valid physics-based solution (Basir and Senocak, 2022b). For many PDEs, although the PINN may satisfy the values at ICs/BCs, the model may fail to propagate it in the solution domain over space and time. PINNs often become stuck in a local loss minimum, and many times fail to converge to the global loss minimum for the problem (Elhamod et al., 2020). A common practice is to use a higher multiplier on the IC's loss to help it to converge first while waiting for the PDE and BCs losses to converge later. However, for higher-order stiff PDEs with rapidly changing gradients in space and time, this may lead to trivial solutions.

Furthermore, stiff-PDEs consist of features with different scales spanning the spatiotemporal domain. First-order optimizers such as the Adaptive Moment Estimation (ADAM) used in standard PINNs are based on fixed diagonal preconditioning, where the gradients are rescaled independently per parameter dimension without modeling the off-diagonal structures of the local Hessian matrix (Kingma and Ba, 2015). Consequently, the diagonal rescaling fails to adapt the step direction to match the underlying changes in the anisotropic loss landscape of the complex PDE being solved. This often leads to either a stall in training when using a small learning rate or a possible overshooting of the local minimum when using a higher learning rate, resulting in limit cycles (oscillations) and therefore a suboptimal convergence of the total loss function (Bock and Weiß, 2022). Combined, the unscaled loss function and the first-order optimizer significantly hamper the training time and convergence of standard PINNs when solving stiff PDEs.

To demonstrate this, we construct a 2D heat plot showing the variation of the relative error across different solution frequencies over time. Following the approach of Rahaman et al. (2019) and Xu et al. (2019), we use the earlier defined variables $u_\theta(x,t)$ and $u_{\text{ref}}(x,t)$ to denote the PINN and numerical solution, respectively. Then, for each time step $t_j$, we calculate the discrete Fourier coefficients, $\check{u}$, of the solution slices for $u_\theta(x,t)$ and $u_{\text{ref}}(x,t)$ using Eq. 30, where $k = 0,1,\dots,\left\lfloor \frac{N_x}{2} \right\rfloor$,

$$\check{u}(k,t_j) = \frac{1}{N_x} \sum_{i=0}^{N_x-1} \left( u(x_i,t_j) - \bar{u}(\cdot,t_j) \right) e^{-\frac{2\pi \mathrm{i} k i}{N_x}} \tag{30}$$

Next, at each $t_j$ the largest $K$ number of Fourier modes, wavenumbers, for $u_\theta(x,t)$ and $u_{\text{ref}}(x,t)$, is selected. The wavenumbers are then ranked according to magnitude, where the largest value wavenumbers are stored for each time slice for a set $K_{t_j}$ using Eq. 31 where $\left| K_{t_j} \right| = K$,

$$K_{t_j} = \underset{k\geq 1}{\arg\operatorname{topK}} \left| \check{u}(k,t_j) \right| \subset \left\{ 1, \dots, \left\lfloor \frac{N_x}{2} \right\rfloor \right\}. \tag{31}$$

We use a value of $K = 5$ for a concise presentation of the results. Each of the $K_{t_j}$ sets are collected across time steps, and a union of values is taken as,

$$\mathcal{K} = \bigcup_{j=1}^{J} K_{t_j} \tag{32}$$

The wavenumbers are then sorted in ascending order where each $\kappa \in \mathcal{K}$ represents a unique wavenumber index using,

$$\mathcal{K} = \{\kappa_1, \kappa_2, \dots, \kappa_{K_{\text{all}}}\}, 1 \leq \kappa_1 < \kappa_2 < \cdots < \kappa_{K_{\text{all}}} \leq \left\lfloor \frac{N_x}{2} \right\rfloor \tag{33}$$

where $K_{\text{all}} := |\mathcal{K}|$ is the total number of unique Fourier modes after the union. Next, for each training iteration of the PINN model, the set of times where $\kappa$ is selected is denoted by $J_\kappa = \{t_j : \kappa \in K_{t_j}\}$. The relative error is then calculated by aggregating over time for each mode using Eq. 34 where $\breve{u}_{\theta_e}(\kappa, t_j)$ represents the PINN model output solution's Fourier coefficient at training epoch, e, and $\breve{u}_{\text{ref}}(\kappa, t_j)$ represents numerical solution Fourier coefficients.

$$\Delta F_e(\kappa) = \frac{1}{|J_\kappa|} \sum_{t_j \in J_\kappa} \frac{\left|\breve{u}_{\theta_e}(\kappa, t_j) - \breve{u}_{\text{ref}}(\kappa, t_j)\right|}{\left|\breve{u}_{\text{ref}}(\kappa, t_j)\right| + \varepsilon}. \tag{34}$$

Finally, to plot a visual representation of the errors for the different wavenumbers as training progresses, the $\Delta F_e(\kappa_m)$ are calculated for all the top wavenumbers $k = 1, \dots, K_{\text{all}}$ at all the training epochs $e = 1, \dots, E$ to form the per-mode training relative error matrix $H \in \mathbb{R}_+^{K_{\text{all}} \times E}$ as

$$H[k, e] = \Delta F_e(\kappa_m) \tag{35}$$

The relative error matrix $H$ plotted against $k, e$ in Fig. 3 demonstrates the lack of convergence for $k > 1$ in training where higher frequency modes oscillate in $\Delta F(\kappa)$ and sustain a high error value during training. It is posited that the inflexibility in loss weight multipliers and the first-order optimizers lead to the stagnation of the training loss. The standard PINN approach is unable to shift focus to resolving fine-scale errors, represented by higher $k$ bands, possibly from poor gradient upgrades by the current optimizer. Moreover, plotting the loss terms from Eq. 6, 7, and 10 against training iteration in Fig. 3, although the IC and BC loss terms decrease in training, the PDE loss remains relatively high and unresolved. As training proceeds beyond the current 500 epochs, there is a larger focus on resolving the ICs where gradient update is possibly rapid, while the PDE loss, critical to convergence, remains largely ignored because gradient flow from the optimizer becomes limited. The constant loss terms weights and the first-order ADAM optimizer may therefore need to be changed to improve the standard PINN convergence rate for solving PDEs.

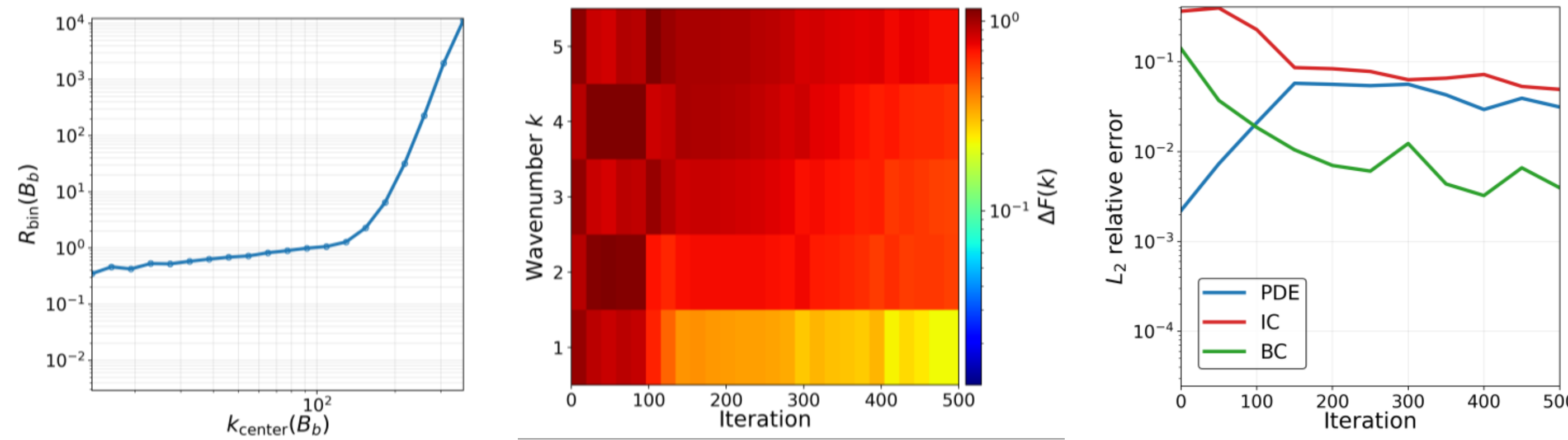

*Figure 3: Left: Log-binned energy-weighted relative spectrum $R_{bin}(\mathcal{B}_b)$ vs wavenumber $k_{center}(\mathcal{B}_b)$ plot showing rapid increase in relative error for high frequency modes, Center: Per-mode relative error heat plot showing a rapid decay in convergence for higher wavenumbers k even as training progresses, Right: Stagnating relative $L^2$ errors for the PDE, IC, and BC loss component as training progresses in standard PINNs.*

Lastly, standard PINNs use a uniform MSE loss for training that prioritizes all collocation points equally. The MSE loss may not account for the variability in the solution space entirely. For stiff PDEs, this may lead regions containing shock-fronts and high-frequency solutions to receive the same importance as smooth and easy-to-resolve regions during training. As a result, a large portion of the high PDE error region(s) may remain under-resolved. This again may lead to a trivial solution where the model may only converge at the known solution data, ICs, and BCs while failing to learn the interior solution in the entire spatiotemporal domain. There are a few studies on the potentially different ways of improving sampling and consequently the loss function, all showing improvements over the standard PINNs approach (Daw et al., 2022; McClenny & Braga-Neto, 2020; Yang et al., 2021).

The basic MSE loss function is hence another limiting factor in the convergence of standard PINNs when solving PDEs. Combined with the inherent NN spectral bias toward lower frequency modes, fixed unity weights on each of the loss components, and the poor convergence of first-order optimizers, standard PINNs struggle to converge sufficiently in time as the PDE complexity grows. Therefore, there is a need for a multi-faceted approach that can overcome standard PINN limitations. This encourages the development of adaptable, error-aware, and integrated PINN models to overcome limitations of standard PINNs. In the next section, we introduce a novel approach leveraging unique solution-finding abilities of multiple PINN extensions into a single generative-adversarial NN framework that is shown to improve modeling of PDEs using physics-based methods.

## 3. Generative Adversarial Physics-Informed Neural Networks (Gen-PINNs)

Motivated by the well-documented limitations of standard PINNs in literature for solving different PDEs, where spectral bias and rigid optimization techniques make it difficult to converge effectively, we propose a different modeling paradigm in the form of Generative Adversarial Physics Informed Neural Networks or Gen-PINNs. The novel PINN extension is developed as a deterministic, coordinate conditioned adversarial framework that employs an explicit generative-adversarial training objective for training. The supporting components are selected to correspond directly to the representation, loss-balancing, temporal-propagation, and optimization limitations identified in Section 2, rather than as independent additions to the framework. Instead of targeting one type of limitation at a time, Gen-PINNs merges multiple documented improvements into a single high-fidelity game based forward PDE solving NN model. Furthermore, by leveraging improved pattern recognition ability of Least-Square Generative Adversarial Networks (LS-GANs), the method seeks to converge across different PDE solution trends using a single adaptable algorithmic approach (Mao et al., 2016).

In Gen-PINNs, the generator $G$, a fully connected NN operating in 64-bit, receives input features, $(x, t)$, and enhances the input using Fourier feature mapping to $\Phi(x, t) = [x, t, \sin(kx), \cos(kx)]_{k=1}^{n_{\text{freq}}}$, where $n_{\text{freq}}$ is the number of embedded Fourier dimensions, to propose the output solution field, $u_\theta(x, t)$. Next, one or more discriminators $\{D_s\}_{s=1}^{S}$ evaluate whether $u_\theta(x, t)$ it is valid across physics-based metrics by inspecting multiple physics loss term collections $\phi_s[u_\theta]$, which go to 0 for the correct PDE solution, and are aggregated into an adversarial loss term. The discriminator adversarial loss includes the PDE residual $f_\theta$, the Sobolev-$H^1$ residual density $(\partial_x f_\theta)^2$, the signed cubic residual, $f_\theta^3$ and the quartic residual $f_\theta^4$. Each $D_s$ takes in $[\phi_s[u_\theta](x, t), x, t]$ and uses an LS-GAN approach where, over the interior points, the real solution corresponds to the zero tensor, while the synthetic generator output is evaluated using the calculated $\phi_s[u_\theta]$ loss term for the respective $D_s$. As shown by Isola et al. (2016), conditional GANs such as Gen-PINNs, where each residual evaluation point is paired with its respective conditional spatiotemporal coordinates, can train with little to no effect of added noise, allowing the generator to learn the deterministic PDE solution without an independent latent input requirement.

The generator is then trained on a combined loss function consisting of the ICs, BCs, PDE, and the adversarial loss supplied by the discriminators.. The discriminators are updated at a set frequency against generator updates to avoid possible gradient saturation. Finally, an L-BFGS optimization stage is added in the second half of training to improve model convergence (Liu and Nocedal, 1989). Gen-PINNs thus offers a deterministic yet flexible design that leverages adversarial NN learning and incorporates complementary regression mechanisms to improve PDE modeling. Using hyperparameter tuning, the method is shown to solve PDEs with a variety of different solution patterns. Below, the architecture of Gen-PINNs is detailed in depth, followed by application to Eq. 12, to illustrate improvement in modeling, and then a generic use algorithm is constructed for further applications of the method.

### 3.1. Gen-PINNs Architecture

The Gen-PINNs architecture combine residual-adversarial objective with supporting representation and optimization components designed to address the PINN limitations identified in Section. 2. . These include LS-GAN physics-based adversarial loss, Fourier input embeddings, dynamic loss weighting, a mixed use of optimizers, variance-informed MSE loss metric, a delayed discriminator learning schedule, and 64-bit training protocol. These components are incorporated to address the complementary representation and optimization limitations identified in Section 2, including spectral bias, loss imbalance, heterogeneous residual errors, and optimizer stagnation. . Below, the components of the novel PINN extension are detailed

**FP64 training regime:** Standard PINNs use 32-bit floating point arithmetic, FP32, in all the arithmetic calculations. The FP32 approach tends to introduce significant round-off and truncation errors across the course of training from repeated gradient updates and round-off errors. These round-off errors can accumulate numerical errors and lead to significant differences between the actual and predicted solutions, even when, theoretically, according to the universal approximation theorem, the network is wide enough to solve a problem (Hornik et al., 1989). As observed by Xu et al. (2025), using FP32 in training prematurely triggers optimizer convergence in training as there is an overflow of digits beyond the recorded 7 after decimal digits that are not captured. This is overcome when the precision is increased to 64-bit floating point arithmetic, FP64, with around 16 decimal digits. FP64 thereafter can significantly reduce the accumulation of round-off errors and improve the stability and accuracy of the automatic differentiation calculations. This becomes especially critical when working with higher-order derivatives and PDEs with shock-front where small changes in gradient calculations can affect the residual calculations and hence the final output of the PINN model. Consequently, to reduce round-off errors, Gen-PINNs uses FP64, including in the generator, discriminators, all loss terms, and Fourier feature embeddings to improve PDE residual modeling as observed in ablation testing. We next look at feature augmentation used to add more information inside the input to the Gen-PINNs model.

**Fourier input embeddings:** To effectively capture high-frequency function features in PDE modeling, we incorporate Fourier feature embeddings, enabling faster convergence for multi-scale solution trends. As noted by Tancik et al. (2020), using a random Fourier feature mapping with an appropriate scale allows coordinate-driven standard NNs to improve high-frequency modeling for low-dimensional problems, including 3-D. Using a Fourier-enhanced input allows injecting sinusoidal features with high-frequency basis functions directly into the PINN model, which accelerates the learning of fine-scale features with an improved optimization landscape for functions with a wide range of solution spectra and reduced spectral bias (Rahaman et al., 2019). Furthermore, as demonstrated by Dong and Ni (2021), properly scaled Fourier feature embeddings can encode periodic boundary structure by mapping periodically equal boundary locations to the same representations, which in turn can eliminate the need for a separate BC loss enforcing that periodic value requirement.

In the Burgers' PDE case, the approach still holds. It improves modeling of the shock front occurring at $x = 0$ by overcoming spectral bias for higher frequency terms. For the spatial coordinates, $x, t$, the Fourier embedding is defined as,

$$\gamma(x,t) = [x, t, \sin(\omega_1 x), \cos(\omega_1 x), \sin(\omega_2 x), \cos(\omega_2 x), \dots, \sin(\omega_m x), \cos(\omega_m x)]^\top \tag{36}$$

where $m = n_{\text{freq}}$ is the number of specified Fourier features. The angular frequency, $\omega_k$, is related to the physical domain length, $L_x$, where periodicity is calculated as,

$$\omega_k = \frac{2\pi k}{L_x}, k = 1,2, \dots, m \tag{37}$$

As $L_x$ for the equation is 2, $\omega_k = \pi k$. The generator $G$ receives an enhanced input tensor such that $u_\theta(x_f^i, t_f^i) = G\left(\gamma(x_f^i, t_f^i)\right)$. The temporal coordinates remain as a raw coordinate as no specific value is imposed at the two ends of the simulation. The periodic frequency ensures that the value of the function is equal at the two ends by satisfying $\sin(\omega_k(x + L_x)) = \sin(\omega_k x)$ and $\cos(\omega_k(x + L_x)) = \cos(\omega_k x)$ for all real numbers of k, which improves the modeling of the trivial Dirichlet BCs where $u(\pm 1, t) = 0$. For the homogeneous Dirichlet Burgers' case 1, the endpoint compatibility complements the explicit BC loss used to impose the prescribed zero boundary values. Moreover, the purpose of using Fourier mapping in Gen-PINNs however, is to enhance input representation of higher frequency features, as discussed next. The Dirichlet BCs are strictly imposed through a separate BCs loss term ensuring that the feature is primarily used for modeling fine scale solution structures.

As shown by Rahimi and Recht (2007), Fourier feature mapping is interpretable through Bochner's theorem as a stationary kernel. For the current 1D case, the kernel, $k_\gamma$, is calculated by multiplying the inner product of the embedding inputs from Eq. 36 as,

$$k_\gamma\big((x_1,t_1),(x_2,t_2)\big) = \gamma(x_1,t_1)^\top \gamma(x_2,t_2) \tag{38}$$

After calculating the above, the kernel becomes a sum of the product of the sine and cosine written as,

$$k_\gamma\big((x_1,t_1),(x_2,t_2)\big) = x_1x_2 + t_1t_2 + \sum_{k=1}^{m} [\sin(\omega_k x_1)\sin(\omega_k x_2) + \cos(\omega_k x_1)\cos(\omega_k x_2)] \tag{39}$$

Using trigonometric identity, $\cos(\alpha-\beta) = \cos\alpha\cos\beta + \sin\alpha\sin\beta$, Eq. 39 reduces to,

$$k_\gamma\big((x_1,t_1),(x_2,t_2)\big) = x_1x_2 + t_1t_2 + \sum_{k=1}^{m} \cos\big(\omega_k(x_1-x_2)\big) \tag{40}$$

where the sum, $h_\gamma(x_1-x_2) = \sum_{k=1}^{m} \cos\big(\omega_k(x_1-x_2)\big)$, is essentially a finite trigonometric polynomial representing a stationary kernel in the spatial dimension and that is constructing a set of spectral values at the specific frequencies $\{\omega_k\}$ (see Tancik et al. (2020) for details).

The effect of the feature map on the kernel is that the function space $\gamma\big(x_f^i, t_f^i\big)$ seen by the first layer of the generator already contains high-frequency Fourier modes as part of the input. As a result, $G$ no longer needs to construct high-frequency features through conventional optimization only. The optimizer can simply adjust the weight ($W$: $w_k^{(\sin)}, w_k^{(\cos)}, w_x, w_t$) and bias ($b$) of the first linear layer acting as pseudo-Fourier coefficients and capturing sharp gradient changes in the solution space with improved accuracy,

$$W\gamma(x,t) + b = \sum_{k=1}^{m} \Big(w_k^{(\sin)}\sin(\omega_k x) + w_k^{(\cos)}\cos(\omega_k x)\Big) + w_x x + w_t t + b \tag{41}$$

This helps to overcome the limitations of conventional MLPs in standard PINNs that inherently tend to optimize low frequency features first in training (Cao et al., 2019; Jacot et al., 2018). Moreover, combining the Fourier features with the explicit BC loss encourages a smooth transition of the solution in the interior points while matching the exact values at the boundaries. In the current Gen-PINNs implementation, $G$ receives $\gamma\big(x_f^i, t_f^i\big)$ while $D_s$ receives $\big(x_f^i, t_f^i\big)$ only to prevent early gradient starvation and avoid critic becomes too good early on in training. Moreover, it is important to note the difference between Fourier representation and the adversarial components. The goal of the Fourier mapping is to enrich the coordinate representation input to $G$, allowing better capture of high frequency solution space as demonstrated above. In contrast, each $D_s$evaluates a physics based residual-feature. $\beta_s\phi_s[u_\theta]$, along with the raw spatiotemporal coordinates as part of conditional GAN design. The learned response from the $D_s$ enters the $L_{\text{ADV}}$to modify the optimization pathway of $G$.

Thus, the Fourier mapping acts to improve availability of raw representation data to $G$, while $D_s$ act on leveraging physics-informed optimization pathways for improved convergence.

**Least Square Generative Adversarial physics-based losses:** Full grid PDE-based losses are constructed and optimized for the adversarial component of training on the LS-GANs method (Mao et al., 2016). A related residual-adversarial approach is DEQGAN, where the PDE is rearranged in a way such that the left hand side is calculated based on the generator dependent terms, while the right-hand side is treated as the physics consistent reference based on the discriminator (Bullwinkel et al., 2022). In contrast, Gen-PINNs retains the direct physics-informed loss terms and the tertiary causal loss term and then integrates the adversarial component which evaluates multiple residual-derived physics features using discriminators conditioned on the spatiotemporal coordinates as explained next.

Using the full-grid PDE residual, $f_\theta(x,t)$ from Eq. 5, over the interior collocation points $\{(x_f^i, t_f^i)\}_{i=1}^{N_f} \subset \Omega$, a number of physics-based metrics, S, for each discriminator, $D_s$, are calculated. Each physics discriminator $\{D_s\}_{s=1}^{S}$ then trains on a unique residual feature including: the regular PDE loss, $\mathcal{L}_{\mathrm{PDE-ADV}}$, representing governing equation violation (s=1)(Eq. 42), the $H^1$ residual density, $\mathcal{L}_{H^1}$, that penalizes roughness or lack of smoothness in the residual using the derivative of the residual (s=2)(Eq. 43), the directional imbalance cubic residual moment, $\mathcal{L}_{\mathrm{CUB}}$ , which preserves the residual sign while penalizing large residual errors(s=3)(Eq. 44) and the quartic residual moment, $\mathcal{L}_{\mathrm{QUART}}$, that penalizes large outlying residual terms independent of the sign (s=4)(Eq. 45). Each of these physics features, such as the $\mathcal{L}_{\mathrm{PDE-ADV}}$, should go towards 0 indicating that the PDE is satisfied over the collocation points for a physically accurate solution.

$$\mathcal{L}_{\mathrm{PDE-ADV}} = \sum_{i=1}^{N_f} \left(f_\theta(x_f^i, t_f^i)\right)^2 \tag{42}$$

$$\mathcal{L}_{H^1} = \sum_{i=1}^{N_f} \left(\partial_x f_\theta(x_f^i, t_f^i)\right)^2 \tag{43}$$

$$\mathcal{L}_{\mathrm{CUB}} = \sum_{i=1}^{N_f} \left(f_\theta(x_f^i, t_f^i)\right)^3 \tag{44}$$

$$\mathcal{L}_{\mathrm{QUART}} = \sum_{i=1}^{N_f} \left(f_\theta(x_f^i, t_f^i)\right)^4 \tag{45}$$

The residual features in Eq. 42-45 therefore using nonlinear transformations a set of informative underlying PDE residual fields in constructed. The squared, cubic, and quartic forms introduce progressively different sensitivity to the residual magnitude, while the $H^1$ based term incorporates spatial variation proportional to the first order derivative $\partial_x f_\theta$. Particularly, the local derivatives of the power-based features with respect to $f_\theta$ are $2f_\theta$, $3f_\theta^2$, and $4f_\theta^3$, respectively, such that larger residual values exert increasingly stronger influence in the higher-order residual fields. The transformed residual fields are then used as the physics-informed inputs to the corresponding $D_s$.

The discriminator specific physics-informed residual feature maps are thus $\phi_s[u_\theta] \in \{f_\theta^2, (\partial_x f_\theta)^2, f_\theta^3, f_\theta^4\}$ calculated using the Generator output, $u_\theta = G\left(\gamma\left(x_f^i, t_f^i\right)\right)$. For each discriminator $D_s$, the fake input is then constructed from $u_\theta$ by calculating the residuals $\phi_s[u_\theta]$ over all the collocation points using Eq. 42-45, denoted by the operator $F_s$. To allow for control over the contribution of each of the physic s feature on the adversarial loss, we scale each of the physics features with weights $\beta_s$. Next, the scaled physics feature residuals, $\beta_s\phi_s[u_\theta]\left(x_f^i, t_f^i\right)$, are concatenated with raw spatio-temporal coordinates, $x_f^i, t_f^i$, and used as the equivalent inputs to the respective $D_s$.

$$\beta_s F_s(G(\gamma\left(x_f^i, t_f^i\right))) := \left[\beta_s\phi_s[u_\theta]\left(x_f^i, t_f^i\right), x_f^i, t_f^i\right];\ \text{s=1,...,S, S=4} \tag{46}$$

Finally, using the LS-GAN implementation the objective functions for the generator and the discriminator are written as Eq. 47 and Eq. 48 respectively with a factor, $\omega$=0.5, added for scaling.

$$\mathcal{L}_{\text{ADV}} = \omega \sum_{s=1}^{S} \sum_{i=1}^{N_f} \left(D_s\left(\beta_s F_s\left(G\left(\gamma\left(x_f^i, t_f^i\right)\right)\right)\right) - 1\right)^2 \tag{47}$$

$$\mathcal{L}_D = \omega \sum_{s=1}^{S} \left[\sum_{i=1}^{N_f} \left(D_s\left(\left[0, x_f^i, t_f^i\right]\right) - 1\right)^2 + \sum_{i=1}^{N_f} \left(D_s\left(G\left(\gamma\left(x_f^i, t_f^i\right)\right)\right)\right)^2\right] \tag{48}$$

Using the equations, it is observable that the Generator $G$ seeks to minimize $\mathcal{L}_{ADV}$ in Eq. 47 by outputting accurate values $u_\theta(\gamma(x_f^i, t_f^i))$ such that the discriminators $D_s$ are unable to distinguish whether the physics residual field is in fact from $G$ or the real null-physics residual field. The $D_s$ output values of probability of a real solution using a value of 1, in this case proportional to how close the collocation value is to zero, or for a fake output, a value close to 0. When the $G$ is well trained, the value of $D_s\left(\beta_s F_s\left(G\left(\gamma\left(x_f^i, t_f^i\right)\right)\right)\right)$ for all the discriminators tends to reach 1, thus minimizing the adversarial loss component. Consequently, the discriminators seek to recognize the real physics residual tensors, $\left[0, x_f^i, t_f^i\right]$, by inducing a value close to 1 to minimize the first term and a value close to 0 for the fake generator tensors, $\left[\beta_s\phi_s[u_\theta]\left(x_f^i, t_f^i\right), x_f^i, t_f^i\right]$, using Eq. 48. Using this game approach, both $G$ and $D_s$ push the boundaries of the conventional single PINN model and improve overall prediction using the adversarial loss component. A detailed mathematical analysis of the residual feature sensitivity to the adversarial gradient and the generator parameter update through the discriminators is provided in Appendix B.

It is noted that the squared penalty terms in the LS-GAN approach are proportional to the error, help prevent oversaturation of gradients, and allow for continuous residual correction even when the discriminator becomes well trained. This trait is less observable in the vanilla GAN loss and the Wasserstein losses that were also tested (Arjovsky et al., 2017; Goodfellow et al., 2014). Next, we look at the tertiary causality-based PDE loss in addition to the adversarial PDE loss and the general PDE loss.

**Tertiary causality-based PDE loss:** A tertiary causality-based PDE loss is next included in Gen-PINNs framework to capture the temporal structure of certain PDE with strongly causal solution structures (Chen et al., 2024; Meng et al., 2020; Roy and Castonguay, 2024; Wang et al., 2024) Instead of using multiple time-evolving NNs where the next NNs IC is dependent on the previous NNs final time solution, a single Gen-PINNs model is used over the entire training time sequence. This, although theoretically less accurate, allows for a more efficient and flexible approach in the Gen-PINNs framework. Particularly, when solving stiff-PDEs with a strong dependence on the initial conditions for the final time solution, the causality-based loss proves useful.

Building on the causal weighting formulation of Wang et al. (2024), the causality PDE loss is added into the Gen-PINNs framework with several key modifications. First, it is important to note that only a single Gen-PINNs model is used over the entire spatiotemporal domain, where the ICs remains fixed at $t = 0$. Next, to improve the rate of convergence to the actual solution, we integrate it into the Variance Informed MSE (VIMSE) loss function with two possible pathways for gradient updates within the exponential function, depending on the complexity of the PDE (Hanna et al., 2025). Also, as there are two other explicit PDE based losses, including the regular global PDE loss $\mathcal{L}_{\text{PDE}}$ and the adversarial PDE loss $\mathcal{L}_{\text{PDE-ADV}}$, we add a dynamic loss weighting system, based on modified gradient-balancing approach motivated by Gradient Normalization (GradNorm) to dynamically scale all the loss multipliers allowing for improved convergence of the optimizer when combining contributions from the disparate loss terms (see section below) (Chen et al., 2017).

To implement the exponentially decaying causality-based PDE loss, $\mathcal{L}_{\text{PDE-exp}}$, the domain is divided into $N_t$ uniform bins such that each time step is a unique bin. Next, marching in time, each collocation point $\left(x_f^j, t_f^j\right)$ is assigned within a bin index, $b(j)$, starting from $t = 0$ to $t = 1$ as,

$$b(j) = \text{clip}\left(\left\lfloor \frac{t_f^j - t_{min}}{\Delta t} \right\rfloor, 0, N_t - 1\right); \Delta t = \frac{t_{max} - t_{min}}{N_t} \tag{49}$$

Thus, residual samples sharing the same bin index, $b(j) = n$, belong to same time level, $t_n$. Using Eq. 5, the PDE residual is calculated for each time bin $n = 0, \dots, N_t - 1$. Within each bin, the MSE, $\mu_n$, and variance, $\sigma_n^2$, is then calculated using Eq. 62 and 63, respectively. Next, the per-bin VIMSE, $v_n$, is calculated as,

$$v_n = \alpha\mu_n + (1 - \alpha)\sqrt{\sigma_n^2} \tag{50}$$

Next, each of the bin losses are accumulated and integrated into an exponential loss function where the contribution of each bin is controlled by $\alpha_{\text{exp}}$ (Eq. 51). As $\alpha_{\text{exp}}$ increases, a larger emphasis is added to earlier time bins such that the cumulative residual of all earlier bins must decrease before the next bin causal PDE weight is minimized. To initialize the bin weights, $w_n$, the first weight is set to $w_0 = 1$ and the remaining weighs for $n > 1$ are calculated sequentially using the previous bin weight as follows

$$w_n = w_{n-1}\exp\left(-\alpha_{\text{exp}} v_{n-1}\right), n = 1, \dots, N_t - 1 \tag{51}$$

$$\mathcal{L}^{V}_{\text{PDE-EXP}} = \frac{1}{N_t}\sum_{n=0}^{N_t-1} w_n v_n \tag{52}$$

The weights in the current implementation are updated at every training iteration. Using the per-bin weight at each time step, the causal PDE loss is aggregated over all time steps with Eq. 52. It is noted that there are at least two possible gradient pathways for the updates. In the causal formulation of Wang et al. (2024), the temporal weights are calculated using the preceding temporal residual losses and are made non-differentiable during backpropagation at each stage such that the optimizer only acts on the weighted residuals for each time bin of the respective NN. This prevents optimize saturation and we use this as the stabilized causal pathway in Gen-PINNs where a single predicting generator NN is used over the entire temporal space considered as one bin. A second fully differentiable pathway is also implemented as it was found to improve convergence better in certain PDEs cases. However, the unrestricted pathway can introduce additional optimization instability for higher-order PDEs with sharp solution features and if no additional optimization approaches are included. It is thus referred to as the unrestricted causal pathway in which the gradients are allowed to propagate through both the residuals and the temporal weights at the same time during backpropagation. In the Gen-PINNs framework, these causal pathways are incorporated within the single generator model and the VIMSE loss is also applied. Using dynamic loss weighting the tertiary causal loss is balanced with the standard PDE loss and the residual-adversarial PDE loss components in the final objective function. In the next subsection, we look at combining the contributions of all the losses via a dynamic weighting scheme. Note: per-time bin causal weights that are used to calculate a value for the $\mathcal{L}^{V}_{\text{PDE-EXP}}$ loss term and are distinct from the per-loss dynamic weights, which dynamically control the contribution of all loss terms, including $\mathcal{L}^{V}_{\text{IC}}$, $\mathcal{L}^{V}_{\text{BC}}$, $\mathcal{L}^{V}_{\text{PDE}}$, $\mathcal{L}^{V}_{\text{ADV}}$, to calculate a final value for optimization.

**Dynamic loss weighting:** In standard PINNs, a fixed unity weight controls the contribution from each weight to the respective loss term and remains constant during training. Consequently, in stiff-PDEs, the optimizer may focus on resolving on a select loss term with the highest magnitude. As the actual physical nature of the loss terms is different, for e.g. ICs vs PDE, the optimizer may mistakenly optimize a single loss term at the expense of others, resulting from a loss weighting discrepancy. There are several studies aimed at elevating the mismatch in the loss term multiplier weights and consist of dynamically updating the loss multiplier terms as the training progresses (Basir and Senocak, 2022b; Li and Feng, 2022; Y. Song et al., 2024; Xiang et al., 2022).

Gen-PINNs consist of several different losses, including the initial condition loss $\mathcal{L}_{\text{IC}}$, boundary condition loss $\mathcal{L}_{\text{BC}}$, PDE residual , $\mathcal{L}_{\text{PDE}}$ , and the adversarial loss $\mathcal{L}_{\text{ADV}}$ term. With static weights and loss multipliers, it is observed that one term may dominate in the loss landscape, hindering other terms' optimization, and promoting a non-ideal convergence to the PDE solution. Therefore, to reduce this effect, it is important that the weight terms, $w$: $w_{IC}, w_{\text{BC}}, w_{\text{PDE}}, w_{\text{ADV}}, w_{\text{PDE-EXP}}$, are adjusted at a set frequency as training proceeds. To achieve this, we employ a modified gradient balancing approach that is motivated by the GradNorm method developed in Chen et al. (2017). In Gen-PINNs, the approach is adapted and applied for a PINNs setting. First, the gradient norm of each component with respect to the generator parameters of each of loss term component, including ICs, BCs and the physics informed loss components, is calculated. Then a geometric mean of the active gradient norms is calculated. Finally, using a scaled ratio of the respective gradient norm of each component with respect to the generator parameters and the overall the geometric gradient mean, the weight of each term is updated in a gradual approach starting from a based value as explained next.

To implement the dynamic loss weighting, the gradient norm of each loss term is calculated with respect to the generator parameters as,

$$g_i = \|\nabla_\theta \mathcal{L}_i\|_2 \tag{53}$$

where $i \in \{\mathrm{IC}, \mathrm{BC}, \mathrm{PDE}, \mathrm{ADV}, \mathrm{PDE} - \mathrm{EXP}\}$. Next, the numerically stable geometric mean for $i = 5$ is calculated as,

$$g_{\mathrm{geo}} = e^{\frac{1}{5}(\log g_{\mathrm{IC}} + \log g_{\mathrm{BC}} + \log g_{\mathrm{PDE}} + \log g_{\mathrm{ADV}} + \log g_{\mathrm{PDE-EXP}})} \tag{54}$$

Next, the new weights, $\widetilde{w}_i$, are calculated at a set frequency using the earlier weight, $w_i^{\mathrm{prev}}$, according to the relative discrepancy from the geometric mean, $g_{\mathrm{geo}}$, using Eq. 55 where $\tau$ is an adjustable factor set to 0.5. For the first update, $w_i^{\mathrm{prev}}$ is equal to the initial base weights, $w_i^{\mathrm{base}}$, for the respective loss term,

$$\widetilde{w}_i = w_i^{\mathrm{prev}} \left(\frac{g_{\mathrm{geo}}}{g_i}\right)^{\tau} \tag{55}$$

Once the updated weight is calculated it is rescaled to match the initial total weight of all loss term multipliers (Eq. 56). $w_i^{\mathrm{base}}$ are set for all terms as anchor point to set a heuristic weight values to a loss term to aid in optimization.

$$\widetilde{w}_i \leftarrow \widetilde{w}_i \frac{\sum_i w_i^{\mathrm{base}}}{\sum_i \widetilde{w}_i} \tag{56}$$

Finally, to allow for a stable change in weights which allows the optimizer to smoothly converge, the final weight is a ratio of the previous and newly calculated weight preventing large abrupt changes in gradients (Eq. 57). $\eta$ is set to 0.3 ensuring that 70% of the weight is carried from the previous value for a smooth update; the value is selected on empirical testing results.

$$w_i^{\mathrm{new}} = (1 - \eta) w_i^{\mathrm{prev}} + \eta \widetilde{w}_i \tag{57}$$

Using the dynamic loss weighting approach, Gen-PINNs avoids dominance of a single term in the loss gradients by balancing the respective parameter gradient contributions. As the weighting is based on the generator gradient norms, $g_i$, and not simply scalar loss magnitudes, as used in standard PINNs, the procedure is able to directly respond to imbalances in the optimization signal produced by the various loss terms. The balanced adjustment of loss weight scales leads to improved training convergence across tested PDEs.

**Gen-PINNs optimization :** Finally, to initiate training in Gen-PINNs the deterministic residual adversarial objective functions in constructed. Typically the MSE is the loss function used in standard PINNs to minimize the unweighted sum of loss terms. Although it performs well in most cases, it tends to average out large residual features or sharp shock fronts occurring in stiff PDEs. This may lead to reduced efficiency in convergence, possibly requiring wider NNs or longer training iterations before converging adequately. As proposed by Hanna et al. (2025), using a VIMSE helps to overcome high local errors and large standard deviation from the mean. In the Gen-PINNs model the VIMSE approach is applied to the active loss components where, the respective $L^1$ errors at the different evaluation points, $(x_i, t_i)$, for ICs, BCs, PDE, PDE-EXP residual, and the adversarial residual are calculated as $\varepsilon_i$ and collected in the error vector $\varepsilon = (\varepsilon_{IC}, \varepsilon_{BC}, \varepsilon_{PDE}, \varepsilon_{ADV})$ using Eq. 58-61.

$$\varepsilon_{IC} = u_\theta\left(x_{\mathrm{IC}}^i, t_{\mathrm{IC}}^i\right) - u_{\mathrm{IC}}^i \tag{58}$$

$$\varepsilon_{BC} = \mathcal{B}\left[u_\theta\left(x_{\mathrm{BC}}^i, t_{\mathrm{BC}}^i\right)\right] - u_{\mathrm{BC}}^i \tag{59}$$

$$\varepsilon_{PDE} = f_\theta\left(x_f^i, t_f^i\right) \tag{60}$$

$$\varepsilon_{ADV} = D_s\left(\beta_s F_s\left(G\left(\gamma\left(x_f^i, t_f^i\right)\right)\right)\right) - 1 \tag{61}$$

Using the mean, $\mu_\varepsilon$, and standard deviation, $\sigma_\varepsilon$, the VIMSE loss, $\mathcal{L}^V$, is calculated for each of the loss components: $\mathcal{L}_{\mathrm{IC}}^V$, $\mathcal{L}_{\mathrm{BC}}^V$, $\mathcal{L}_{\mathrm{PDE}}^V$, $\mathcal{L}_{\mathrm{ADV}}^V$ where $\alpha_{VIMSE} \in [0,1]$ is used to control the contribution of the mean and standard deviation in the respective loss term (Eq. 62-64). $\mathcal{L}_{\text{PDE-EXP}}^V$ is calculated earlier in Eq. 52.

$$\mu_\varepsilon = \frac{1}{N}\sum_{i=1}^{N} \varepsilon_i^2 \tag{62}$$

$$\sigma_\varepsilon = \sqrt{\frac{1}{N}\sum_{i=1}^{N} (\varepsilon_i^2 - \mu_\varepsilon)^2} \tag{63}$$

$$\mathcal{L}^V = (\alpha_{VIMSE})\mu_\varepsilon + (1 - \alpha_{VIMSE})\sigma_\varepsilon \tag{64}$$

In the Burgers' PDE application that follows, a fixed value of 0.9 is used for $\alpha_{VIMSE}$. Combining the active loss terms, the generator objective function is constructed as Eq. 65 where $w{:}\, w_{IC}, w_{\mathrm{BC}}, w_{\mathrm{PDE}}, w_{\mathrm{ADV}}, w_{\mathrm{PDE-EXP}}$ represent the corresponding weights of the respective loss component that are updated according to Eq. 53-57. The VIMSE formulations in Eq. 58-64 are applied to each of the loss components before multiplication with the dynamic weights.

$$\mathcal{L}_G = w_{\mathrm{IC}}\mathcal{L}_{IC}^V + w_{\mathrm{BC}}\mathcal{L}_{\mathrm{BC}}^V + w_{\mathrm{PDE}}\mathcal{L}_{\mathrm{PDE}}^V + w_{\mathrm{ADV}}\mathcal{L}_{\mathrm{ADV}}^V + w_{\mathrm{PDE-EXP}}\mathcal{L}_{\mathrm{PDE-EXP}}^V \tag{65}$$

Eq. 47-65 are used to define the coupled optimization structure used in Gen-PINNs. During a generator update, $G$ minimizes the global objective $\mathcal{L}_G$ in Eq. 65, while the discriminator parameters are kept fixed. During a discriminator update, each $D_s$minimizes the LS-GAN objective defined in Eq. 48 for the active residual feature defined in Eq. 42-46. The two updates are linked through the deterministic solution produced by $G$ where the produced solution is used to construct the residual adversarial features evaluated by $D_s$ which in turn is contributes back to the generator through $\mathcal{L}_{\mathrm{ADV}}^V$ using Eq. 47 and Eq. 65. Thus, the resulting optimization approach integrates direct physics constraints, adaptive loss weighting and a game residual adversarial feedback mechanism into a unified deterministic framework.

**Optimizer configuration:** Standard PINNs use the first-order ADAM optimizer for training, although it performs well in the initial stage of optimization, often stalls in the convergence of the loss function to an accurate solution (Taylor et al., 2022). Second-order accurate optimizers, on the other hand, such as L-BFGS, are based on calculating the Hessian matrix, $H_k$, of the loss function and are thus curvature aware (Liu and Nocedal, 1989). As observed in Eq. 66 for the L-BFGS function, the NN parameters at iteration $k$, $\theta_k$, undergo a quasi-Newton step to update to the new parameters, $\theta_{k+1}$, where the direction of parameter change is proportional to the gradient of the losses with respect to $\theta_k$ given by $\nabla\mathcal{L}(\theta_k)$ (first order information), the curvature given by $H_k^{-1}$ (second order information) and the internal learning rate given by $\rho_k$. Consequently, with using curvature-based optimization, L-BFGS is better able to traverse the loss landscape and converge more efficiently for fine-scale features in the solution space.

$$\theta_{k+1} = \theta_k - \rho_k H_k^{-1}\nabla\mathcal{L}(\theta_k) \tag{66}$$

Rathore et al. (2024) show that using ADAM alone leads to poor convergence. However, combining ADAM and L-BFGS shows significant improvements in modeling an accurate solution, as corroborated by recent studies (Kiyani et al., 2025; Urbán et al., 2025). Following this approach a two-step optimization process is adopted. First, the loss function is optimized with the ADAM optimizer, followed by L-BFGS. This helps save time and computational resources during initial optimization, where loss gradients guide the optimizer close to a global minimum while allowing for exact solution finding in the later stage. For the implementation, the discriminator is updated at a lower frequency than the generator at a rate of 1 discriminator per 10 generator updates. For ADAM, the learning rate is set at $10^{-3}$ while for the L-BFGS optimizer, the history size using parameter and gradient pairs for curvature calculation is set at 5, and the gradient tolerance as an early stop criterion for convergence is set at $10^{-6}$.

## 3.2. Hyperparameter tuning

To identify ideal hyperparameter settings for the multiple components in Gen-PINNs, structured configuration evaluations are performed. The contribution of individual components and parameter values is evaluated according to the resulting PDE solution behavior and convergence. Table.1 serves as a practical guide for selecting appropriate configuration of Gen-PINNs components based on theoretical understanding and practical results. It is organized into the baseline framework and NN architecture and training for general starting choices and problem-informed loss configurations for spectral, temporal, boundary and residual characteristics of the PDE. Building on this guidance, Algorithm 1 serves as a structured approach for configuring and further refining problem-informed components for the Gen-PINNs model configuration. The procedure links the observed PDE and training characteristics with the congruent Gen-PINNs components defined in the earlier sections and listed in Table. 1.

For the Burgers' PDE case 1 configuration, the number of Fourier features is specified to 10 so the total number of features in the first layer is raised to 22 based on Eq. 36. The number of neural layers and number of neurons in the $G$ is specified higher than the $D_s$ to avoid the discriminators from becoming too critical early on in training which may hinder convergence of $G$. Similarly, the $D_s$ NNs are updated once every 10 $G$ NN updates, enabling smooth adversarial feedback. Furthermore, to match the standard PINNs implementation, the total number of iterations is limited to 500, where half is split between ADAM followed by L-BFGS optimizers. A value of 0.9 is used for the $\alpha_{VIMSE}$ ,

allowing a 10% contribution from the variance of error to the total error. In this case, the dynamic weights are not applied, and the $\mathcal{L}_{\text{PDE}-\text{ADV}}$ is the only loss used inside the adversarial component for simplicity of demonstration. These settings follow the guidance in Table. 1 and Algorithm 1 with specific values used in the Burgers' PDE case 1 application listed in Table. 4 (Appendix A).

*Table 1: Gen-PINNs optimization hyperparameters*

| Category | Hyperparameter | Configuration guidance |
|---|---|---|
| **Baseline Framework** | Weight Initialization | 'Xavier' for $G$ and $D_s$ NNs |
| | Activation type | 'Tanh' to allow smooth automatic differentiation for higher order PDEs |
| | $G$: $D_s$ update frequency | Update $D_s$ less frequently than $G$ to prevent early discriminator dominance |
| | Optimizer Stage 1 | ADAM 1st order algorithm for initial optimization to enable faster convergence |
| | Optimizer Stage 2 | L-BFGS 2nd order algorithm for late-stage optimization for challenging loss landscapes |
| | L-BFGS History size | Small value for tracking (Default: 5) |
| **NN architecture and training** | Fourier frequencies in $G$ | Increase for sharp fronts, oscillatory solutions, or unresolved high-frequency modes |
| | $G$ NN structure | Increase capacity for more complex or highly multiscale solution fields |
| | $D_s$ NN structure | Keep $D_s$ smaller than $G$ to limit discriminator dominance |
| | ADAM Learning rate | Common starting rate across PDE cases; reduce only if ADAM training becomes unstable |
| | Iterations Stage 1 | Increase until ADAM losses continue to decrease (not plateau) before switching to L-BFGS |
| | Iterations Stage 2 | Increase when L-BFGS continues to improve the converged solution for all losses |
| **Problem-informed loss configuration** | $\alpha_{VIMSE}$ | Increase variance contribution when large localized residual errors remain under-resolved |
| | Dynamic Weights | Generally enable for improved gradient stabilization across the different loss metrics. Decrease $\eta$ for higher stability if slower weight update is required (Defaults: $\eta = 0.3$) |
| | $w_{IC}^{\text{base}}$ | Recommend a higher value for the IC weight for establishing convergence on known data |
| | $w_{BC}^{\text{base}}$ | Use a moderate weight multiplier to ensure convergence on BCs for the Dirichlet PDEs |
| | $w_{PDE}^{\text{base}}$ | Reduce when the global PDE loss competes with a more effective causal PDE formulation |
| | $w_{ADV}^{\text{base}}$ | Maintain sufficient adversarial signal without allowing it to dominate direct physics losses |
| **Problem-informed temporal configuration** | Exponential PDE Loss | Use if the PDE solution is strongly causal, avoid using for simple problems |
| | $\alpha_{\text{exp}}$ | Increase or decrease by an order of magnitude when the PDE solution regime is strongly dependent or less dependent on the initial PDE solution respectively (Default: $1 \times 10^2$) |
| | Gradient pathway | Use stabilized causal gradients when unrestricted propagation through the causal weights destabilizes training as only a single $G$ NN is used |
| | $w_{PDE-EXP}^{\text{base}}$ | Increase when the PDE solution is strongly causal to allow for faster convergence |
| **Problem-informed adversarial configuration** | Adversarial weight type | Use the direct PDE residual as the primary feature ($\beta_1$), Add Sobolev-type features if there are rapidly varying residuals ($\beta_2$), Add signed cubic features if there are one-sided residual peaks ($\beta_3$), and add quartic features if there are large localized residuals ($\beta_4$) |
| | $\beta_s$ | Balance adversarial residual-feature weights based on the respective contribution to convergence |

**Algorithm 1: Gen-PINNs configuration and component selection procedure**

**Input:** PDE $\mathcal{N}[u] = 0$, ICs, BCs and Baseline framework settings in Table 1
**Expected Output:** Selected Gen-PINNs configuration and PDE solution $u_\theta(x, t)$
**Procedure:**

1. **Initialize** Generator $G$, Discriminators $D_s$, Base loss weights $w_i^{\text{base}}$, and the Baseline framework and NN architecture and training settings from Table 1.
2. **Set** Fourier input mapping $\gamma(x, t)$ according to Eq. **36–41**.
   **if** sharp gradients, oscillations, or high-frequency modes are insufficiently resolved **then**
      increase $n_{\text{freq}}$ and include Fourier features in the required $x$ or $t$ directions.
3. **Construct** the adversarial residual features $\Psi_s(r_\theta)$ using Eq. **42–45** and adversarial objectives using Eq. **46–48**.
4. **Select** active $D_s$ adversarial features and corresponding $\beta_s$ multiplier values where:
   $s = 1$: Unweighted PDE residual,
   $s = 2$: Sobolev-type PDE residual for rapidly varying residuals,
   $s = 3$: Signed-cubic PDE residual for one-sided peaks,
   $s = 4$: Quartic PDE residual for large localized residuals.
5. **if** the PDE exhibits strong temporal causality **then**
      activate $L_{\text{PDE-EXP}}$ and set $w_{\text{PDE-EXP}}^{\text{base}} > 0$ using Eq. **49–52**
      Select $\alpha_{\text{exp}}$ over approximately $10^1 - 10^3$
      **if** later-time propagation is suppressed **then**
         decrease $\alpha_{\text{exp}}$
      **if** early-time errors are insufficiently prioritized **then**
         increase $\alpha_{\text{exp}}$
      **else**
         set $w_{\text{PDE-EXP}}^{\text{base}} = 0$.

Note: excessive causal emphasis can prevent later-time propagation as only a single Gen-PINN is propagating the solution from start to end of time.

6. **if** competes with **then**
      reduce $w_{\text{PDE}}^{\text{base}}$, including $w_{\text{PDE}}^{\text{base}} = 0$ when needed.
7. **if** unrestricted causal gradients destabilize optimization in higher order PDEs **then**
      use the **stabilized causal** gradient pathway
   **else**
      retain the unrestricted causal pathway.

Note: This distinction is defined earlier in the methodology and is important in the Allen–Cahn PDE analysis.

8. **if** gradients of the active losses $L_i$ are imbalanced or struggling to converge **then**
      activate dynamic weights $w_i$ using Eq. **53–57**;
   **else**
      retain $w_i = w_i^{\text{base}}$

9. **Apply** VIMSE to the active loss components using Eq. **58–64** and selected $\alpha_{\text{VIMSE}}$.
10. **Assemble** the final generator loss $\mathcal{L}_G$ according to Eq. **65**.
11. **Train** $G$ and $D_s$ using the selected $G$: $D_s$ update frequency and ADAM learning rate based on Table 1.
12. **Continue** the optimization where ADAM left off with L-BFGS for smaller number of iterations according to Eq. **66** when ADAM convergence begins to stagnate.
13. **Evaluate** the corresponding IC, BC, PDE-residual, PDE-temporal and PDE adversarial convergence over the training period.
14. **if** a dominant unresolved characteristic remains **then**
    update only the corresponding configuration in Steps **2–9** and retrain.

**Return** the selected Gen-PINNs configuration and the PDE solution $u_\theta(x, t)$.

### 3.3. Gen-PINNs application to the Burgers' benchmark PDE

Using the hyperparameters listed in Table 1, the Gen-PINNs model is trained over the benchmark Burgers' PDE and compared against the numerical solution (Eq. 12-14). The newly predicted $u(x, t)$ closely matches the numerical solution, including an accurate match of the viscous shock front layer at $x = 0$ (Fig. 4). Accordingly, the relative error decreases from 35.41% in standard PINNs to 2.99% in Gen-PINNs across all time steps (Fig. 5). Moreover, all important trends in the predicted solution including minimum, maximum values as well as occurrence of inflection points are matched accurately. This indicates that in a short iteration span, 500 epochs, the generator, guided by the various physics loss components and the discriminator, accurately captures the global structure and sharp gradient change, avoiding large errors as seen in the standard PINN prediction.

Next, plotting the spectral and training diagnostics shows that the Gen-PINNs approach significantly lowers the spectral bias previously seen in the standard PINNs approach (Fig. 6). Observing the relative error spectrum plot, there is a much smaller increase in relative error spectrum $R_{\text{bin}}\,(\mathcal{B}_b)$ as the average binned wavenumber $k_{\text{center}}$ increases. This indicates a lower fraction of missed energies for higher frequency features compared to the standard PINNs trend, where the $R_{\text{bin}}\,(\mathcal{B}_b)$ rises start at an inherently high value of 0.5 compared to a value of $0.5\times10^{-2}$. There is also a more consistent rise in $R_{\text{bin}}\,(\mathcal{B}_b)$ as $k_{\text{center}}$ increases, unlike standard PINNs, where there is a rapid rise in error when $k_{\text{center}}$ is around $10^{2,}$ indicating a large bias against high frequencies (Fig. 3).

The pattern is repeated for the per-mode relative error heat plot for the wavenumber pattern vs iteration number (Fig. 6). Previously, in standard PINNs, only wavenumber 1 converged, leaving higher frequency features relatively unresolved with a $\Delta F(\kappa)$ value of 1 for wave number 2-5. With Gen-PINNs approach leveraging multiple frequency resolving methods, there is a large drop in $\Delta F(\kappa)$ falling below 0.1 for all wavenumbers. Moreover, the wavenumber 2-5 shows a significant reduction in error as the number of iterations increases, indicating a reduced spectral bias. Finally, the IC, BC, and PDE error histories are plotted. Under the same training process there is much larger decrease in the individual errors over time compared to standard PINNs (Fig. 6). The reduction of the log-binned energy-weighted relative spectrum at larger wavenumbers, the per-mode errors for the higher dominant modes and the larger loss term resolution indicated that Gen-PINNs is able to better resolve low and high frequency features and reduce the governing equation mismatch simultaneously. The pattern significantly differs from standard PINNs behavior in Fig. 3, where high frequency features remain under resolved and the different losses decrease less and oscillate in magnitude failing to converge in the set training conditions. It is important to note; however, that further improvements are possible by increasing the number of iterations for all PINN models. However, as the purpose of analysis is to identify and improve current limitations in standard PINNs, the total number of training iterations is capped to make it easier to note changes as components in Gen-PINNs are analyzed. In the next section, we use the Gen-PINNs extension to benchmark a series of physics based PDE and compare against the numerical and standard PINNs solutions.

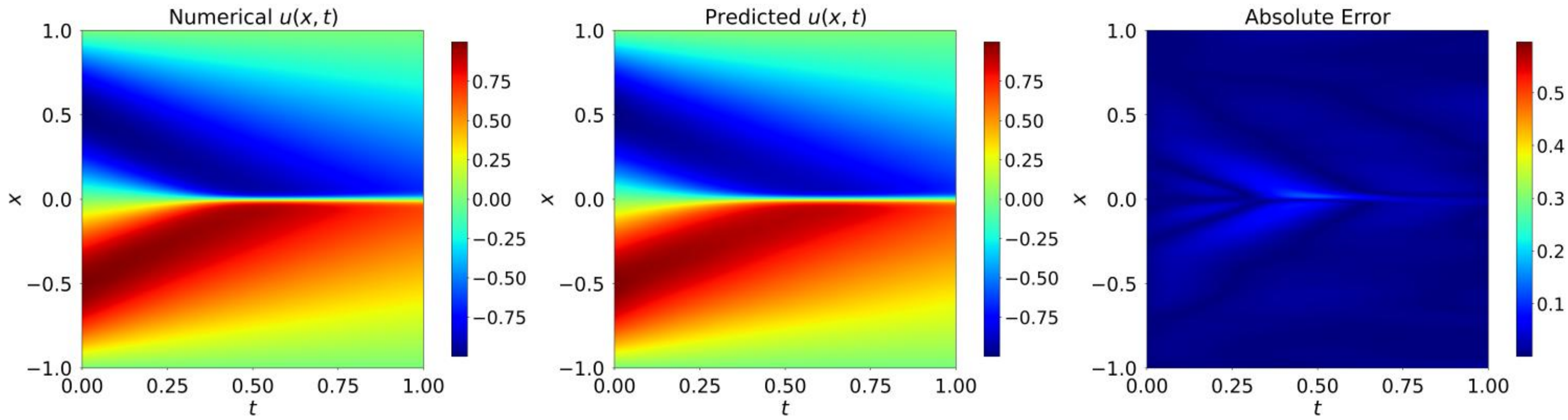

*Figure 4: Left: Numerical solution, Center: Gen-PINNs prediction, Right: Absolute error in Gen-PINNs prediction of u(x,t) for the Burgers' PDE. A $L^2$-relative error of 2.99% is observed.*

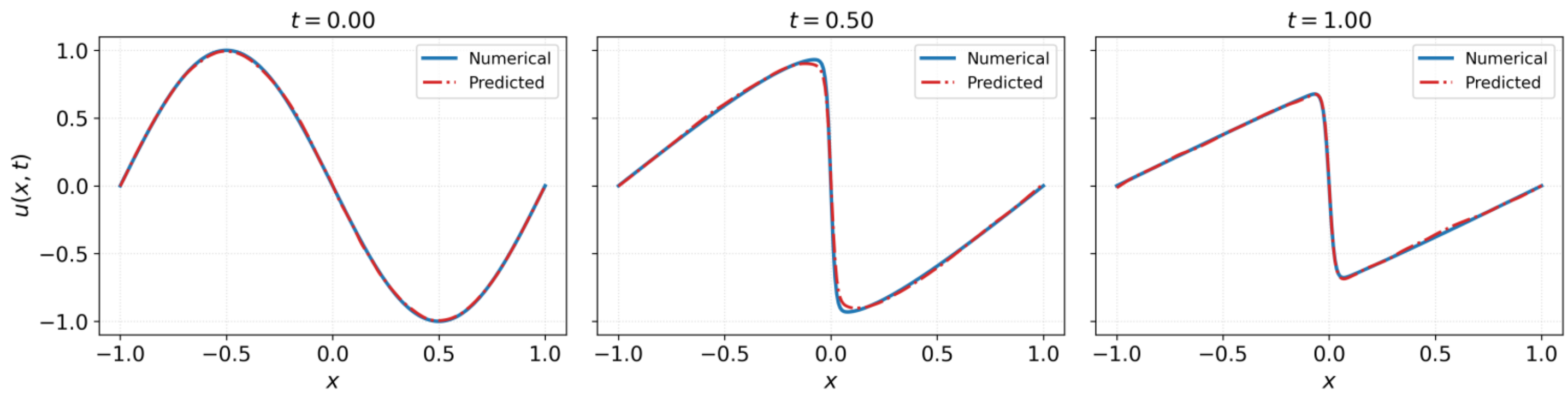

*Figure 5: Left: t=0, Center: t=0.5, Right: t=1 line plot comparison of the numerical solution (red) and Gen-PINNs (blue) prediction for the Burgers' PDE*

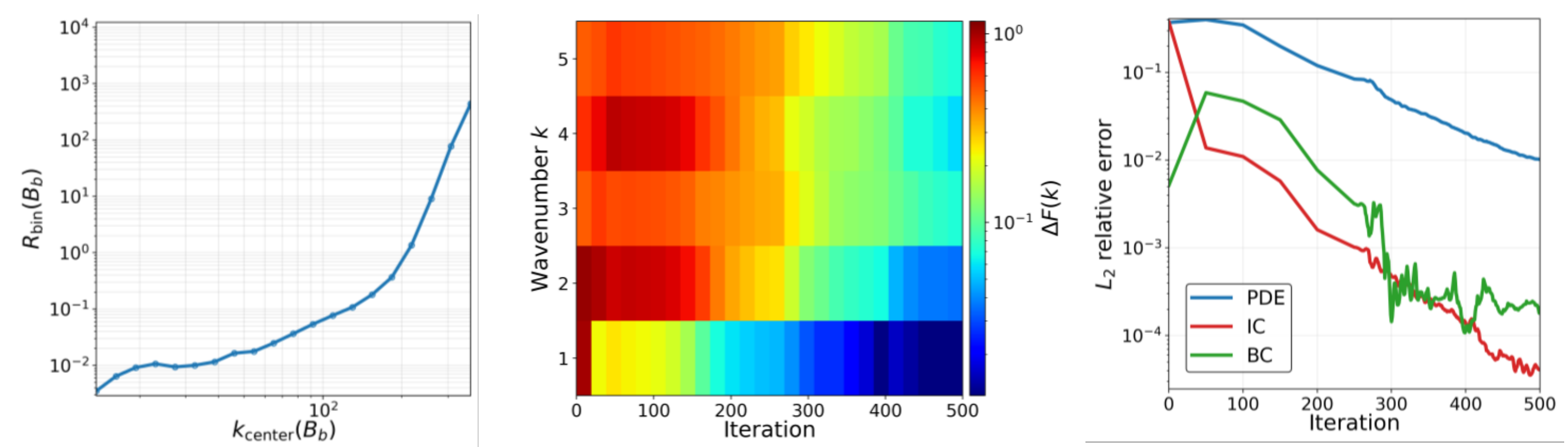

*Figure 6: Left: Log-binned energy-weighted relative error spectrum $R_{bin}$ ($\mathcal{B}_b$) vs wavenumber $k_{center}$ ($\mathcal{B}_b$) plot a steadier yet lower increase in relative error for high frequency modes compared to standard PINNs, Center: Per-mode relative error heat plot showing reduction in the error for all wavenumbers k as training progresses albeit higher reduction in error for lower wavenumbers, Right: Rapidly decreasing relative $L^2$ errors for the PDE, IC and BC loss component as training progresses in Gen-PINNs*

## 4. PDE Experiment Results

To demonstrate the effectiveness of Gen-PINNs, a series of PDE experiments is conducted. The benchmark set focuses on experimenting on one dimensional forward PDE problems selected to span distinct nonlinear, sharp-front, stiff-interface, temporal, spectral, and higher-order solution scenarios. These PDE solutions contain sharp gradient changes, multi-scale features, and are sensitive to initial and boundary conditions. In these challenging regimes, issues such as spectral bias, optimizer stagnation and different gradients on the losses affect standard PINNs convergence. Specifically, problems with shock-fronts are experimented with, including two more Burgers' Equations under different ICs and BCs, an Allen-Cahn equation with a small diffusion coefficient, and a Kuramoto-Sivashinsky equation with small linear hyperviscosity and diffusion coefficients. Focused benchmark suites of this scale are also common in many PINN studies, where a limited number of PDEs are used to evaluate performance of a PINN extension under specific solution characteristics. For example, the standard PINN formulation of Raissi et al. (2017a) was introduced on three PDEs, Wight and Zhao (2020) considered two PDEs, and Wang et al. (2024) also considered two PDEs together with the Lorenz dynamical system. Accordingly, as in many studies the problems are solved using the fully implicit method or the Exponential Time Differencing Rungga-Kutta 4$^{th}$ order (ETDRK4) method in case of the Kuramoto-Sivashinsky equation to form an accurate reference (numerical) solution. The solution output of standard PINNs and Gen-PINNs is compared against the numerical solution to evaluate the accuracy of prediction using the relative $L^2$ error metric.

Standard PINNs are used as the common reference baseline for comparison throughout the PDE experiments to provide a direct controlled comparison of the complete Gen-PINNs formulation that uses a single NN for making forward predictions.. Moreover, the standard PINNs models and Gen-PINN models are trained using similar hyperparameter settings consistent with other PINN extension studies for a fair comparison. These include using the hyperbolic tangent (Tanh) activation functions, a learning rate of $1\times10^{-3}$ for the ADAM training phase, and the same specification of predicting NNs with weights initialized using the Xavier scheme (Glorot and Bengio, 2010). The stochastic model initialization and sampling are set using a fixed JAX pseudorandom number generator seed available within the corresponding notebook of each PDE. The total number of iterations for each case is also kept constant and the same IC, BC, and interior collocation points are used in training. The additional configuration settings used in the Gen-PINNs model are framework specific components including the residual Fourier enhanced input and adversarial losses that are absent in standard PINN formulation. The Fourier expanded input results in small increase in the number of NN parameters while discriminator networks contain additional NN framework parameters depending on the active adversarial features. For the Gen-PINNs models, Table 1 is first used to select the appropriate problem-informed components based on the temporal, spectral, residual, and solution characteristics of the PDE, after which Algorithm 1 provides the structured training procedure for the selected configuration. Throughout the PDE experiments the following components in Gen-PINNs are always included unless otherwise stated, including FP64 accuracy, Fourier input embeddings, global PDE adversarial loss, optimizer scheduling, and the VIMSE loss. The dynamic weighting scheme and components within the adversarial loss are adjusted as needed for better PDE modeling. A consolidated analysis of the component ablation and sensitivity evaluations across the PDE cases is provided in Section 4.5. It is important to note that the reported configurations in the individual PDE analysis are drawn from a broader component and sensitivity studies performed during the model development. The current analysis is used to illustrate how the residual adversarial Gen-PINNs framework responds to different PDE solution characteristics in training. The deterministic residual-adversarial framework incorporates a range of techniques adapted and developed from earlier PINN extensions, allowing the configuration to be effectively leveraged for multiple characteristics of PDE solutions. . The training hyperparameters are listed in Table. 4, Appendix A for all the PDEs. Once trained, the standard PINN and Gen-PINN model can predict the solution of a PDE at any discretized point within the confines of the spatio-temporal training domain. For testing, the entire discretized domain in the numerical solution is used to calculate the output of the standard PINNs and Gen-PINNs method.

### 4.1. Burgers' PDE case 2

For the first experiment, the Burgers' PDE with different ICs and BCs settings is analyzed. Using a fully implicit method, Eq. 12 is solved for $\nu$ = 0.01 over $t \in [0,1]$ and $x \in [0,1]$ where ICs and BCs are given by Eq. 67-68.

$$u(x, 0) = \sin(\pi x) \tag{67}$$

$$u(0, t) = u(1, t) = 0 \tag{68}$$

This PDE is similar to the first case; however, it is over half the $x$ domain and no longer symmetric about $x = 0$, which is challenging to model using PINNs. We first attempt to solve this using the same optimization hyperparameters as in Table 1, however, the Gen-PINN only converges as well as the standard PINN. To overcome this, a grid search is performed over the hyperparameters, and it is found that turning on the tertiary causal PDE loss, along with dynamic weighting over the loss terms enabled a better match within a total of 1000 iterations (Fig. 7-8). We use a high value for the causality parameters, $\alpha_{\text{exp}} = 1 \times 10^3$, that forces the model to resolve early time-mismatch before resolving later time errors. As indicated in several studies, enforcing a causal PDE loss prevents the optimizer from getting lost in converging multiple troublesome physics mismatches simultaneously, where it may fail to reach an accurate solution. Moreover, the dynamic weighting approach allows for the 5 different loss terms to balance over training, preventing over resolution of one of the loss terms while ignoring others. It also helps balance the mismatch physical meaning of the various losses, allowing for a better converged model. The total number of predicting NN parameters for the models is approximately equal, with 50,049 and 52,609 parameters in the standard PINNs and Gen-PINNs model respectively. The final $L^2$ relative error at evaluation is 50.63% for standard PINNs and 1.42% for Gen-PINNs, indicating a large improvement in convergence with the addition of the causal and dynamic weighting scheme.

To investigate the approach further, the number of total iterations is increased to 2000, while holding all other hyperparameters constant. This resulted in a drop $L^2$ to 0.91%. The corresponding standard PINNs relative $L^2$ decreased to 8.21% (Appendix C.1). Next, the remaining physics adversarial components ($\mathcal{L}_{H^1}, \mathcal{L}_{\text{CUB}}, \mathcal{L}_{\text{QUART}}$) are activated with multiplier values of $\beta_1$= 1 and $\beta_2, \beta_3, \beta_4$ = $1\times10^{-2}$ and the training is repeated. The relative $L^2$ subsequently decreases to 0.88%, highlighting a small improvement from activating the adversarial residual features at an already low-error stage of training (Appendix C.2). The weights, $\beta_2, \beta_3, \beta_4$, are increased to 1 to match the $\mathcal{L}_{\text{PDE-ADV}}$ multiplier, and the training is repeated. The relative $L^2$ error, however, increases to 1.38%, confirming that only a small contribution from the non-global PDE physics adversarial components ($\mathcal{L}_{H^1}, \mathcal{L}_{\text{CUB}}, \mathcal{L}_{\text{QUART}}$) is needed to improve testing accuracy. Finally, the number of iterations is increased to $1\times10^5$ , while restoring $\beta_1$= 1 and $\beta_2, \beta_3, \beta_4$ = $1\times10^{-2}$, training is rerun for the benchmark models. The relative $L^2$ error is now 0.87% for standard PINNs and 0.77% for Gen-PINNs. Thus, for certain PDEs the large accuracy differences observed at shorter training budgets are less pronounced if prolonged optimization is used with standard PINNs.

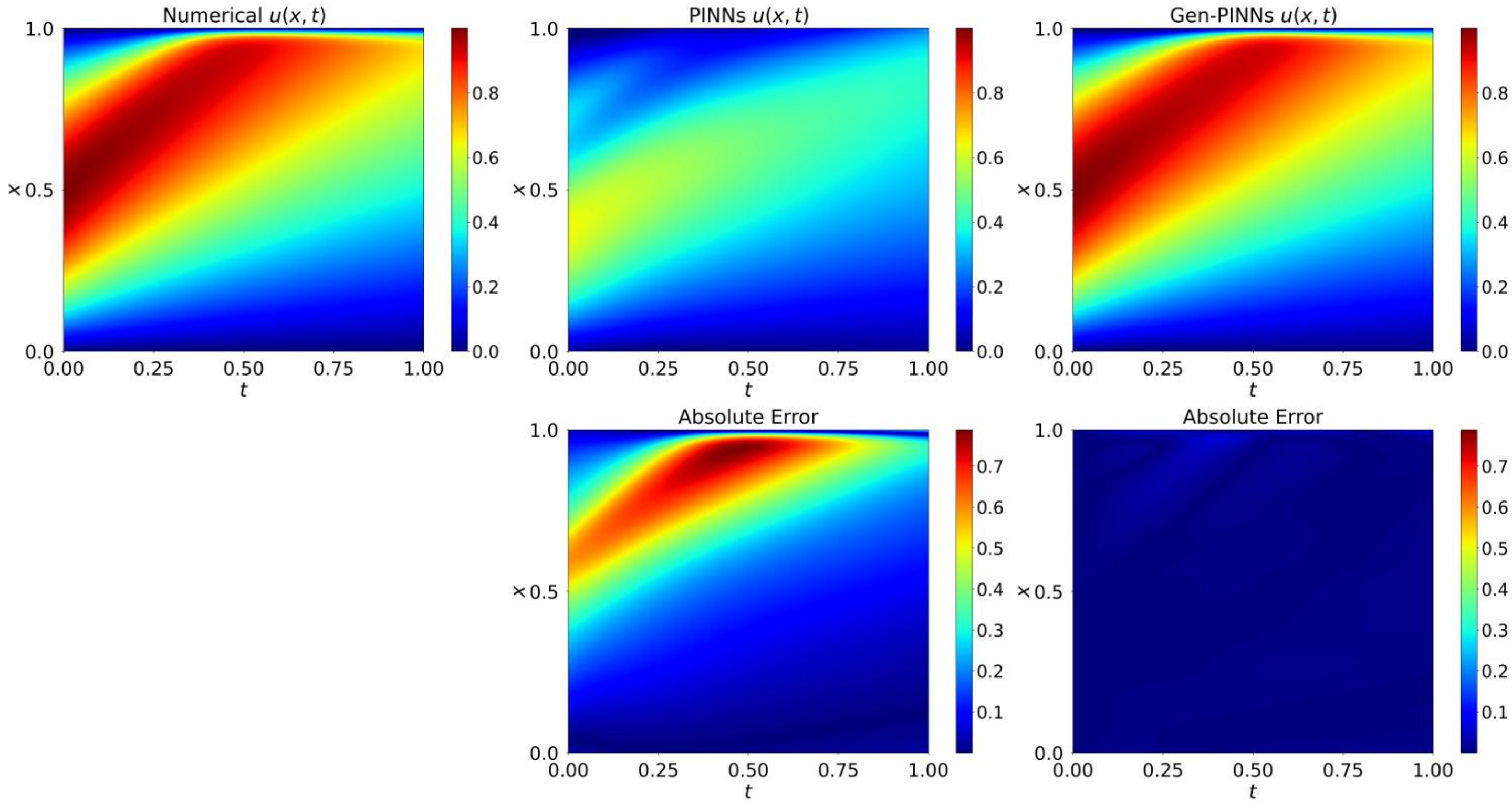

*Figure 7: Top Left: Numerical solution, Top Center: PINNs prediction, Top Right: Gen-PINNs prediction, Bottom Center: Absolute error in PINN prediction, Bottom Right: Absolute error in Gen-PINNs prediction for case 2 of the Burgers' PDE*

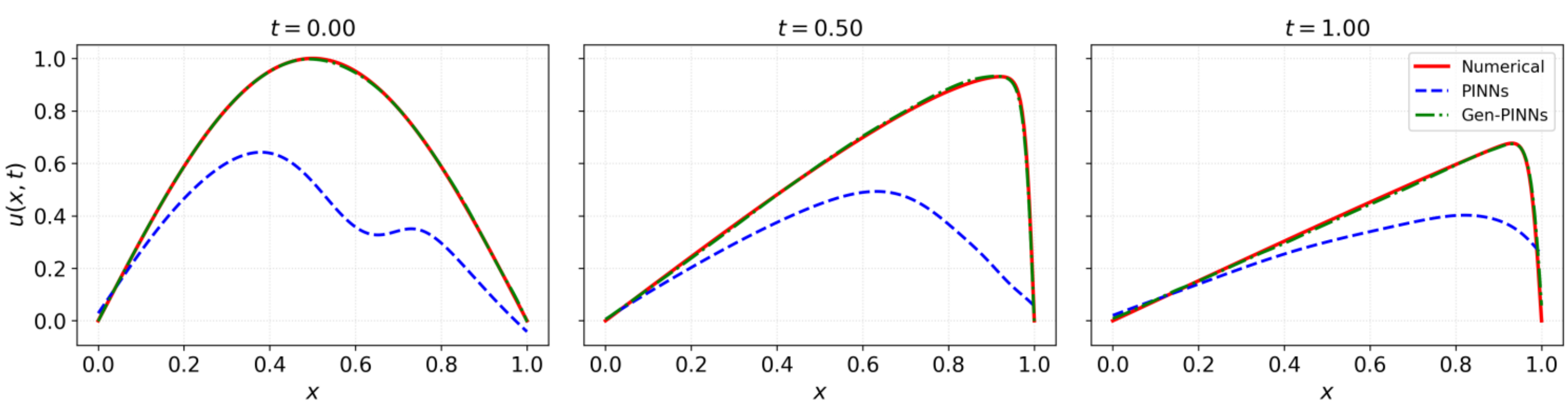

*Figure 8: Left: t=0, Center: t=0.5, Right: t=1 line plot comparison of the numerical solution (red), PINNs prediction (blue) and Gen-PINNs prediction (green) for case 2 of the Burgers' PDE*

## 4.2. Burgers' PDE case 3

In the next PDE experiment, a different version of the Burgers' Eq. 12 is solved where $\nu$ = 0.003 over $t \in [0,1]$ and $x \in [0,1]$ and ICs and BCs are given by Eq. 69-70. The exact analytical solution Eq. 71, with variables defined in Eq. 72, of the PDE constructs a more accurate reference compared to the numerical version. This is the reason for selecting at least one such case application to identify possible pathologies in PINN models before applying them to a more challenging case.

$$u(x,0) = \phi(x,0) \tag{69}$$

$$u(0,t) = \phi(0,t), u(1,t) = \phi(1,t) \tag{70}$$

$$u(x,t) = (0.1e^{-A} + 0.5e^{-B} + e^{-C})/(e^{-A} + e^{-B} + e^{-C}) \tag{71}$$

$$\begin{aligned} A &= \frac{50}{3}(x - 0.5 + 4.95t) \\ B &= \frac{250}{3}(x - 0.5 + 0.75t) \\ C &= \frac{500}{3}(x - 0.375) \end{aligned} \tag{72}$$

In this version of the Burgers' PDE, the BCs are non-constant and unequal. The aim is to see what Gen-PINNs model specifications converge to an accurate solution in the least number of iterations. To do this a grid search is again performed over the hyperparameters. Initially, we start off with the learnings from the last PDE by once again applying the causal PDE weight and running it with a similar hyperparameter setting as reported in Table 2 for 4.1. This surprisingly results in a large error in the Gen-PINNs model with a relative $L^2$ error of 11% compared to 2.02% for standard PINNs (Appendix D.1). After analyzing the space average errors plots across the predicting temporal domain, it is noted that the causality PDE loss imposes a strong bias towards resolving the initial time errors (Appendix D.2). This results in a strong focus on the losses at earlier times and if the PDE, as in the current case, is not strongly causal prevents the solution from propagating adequately to later times.

Consequently, the causal PDE loss is removed to explore other possible hyperparameter combinations. Continuing the tests, an additional Fourier input feature embedding is added in the $t$ direction (Appendix D.3). As the BCs for this PDE are varying, it is noticed through empirical testing that additional random Fourier frequencies in $x$ and $t$ directions improve convergence. It is also observed again that using large adversarial physics loss weights other than for the adversarial PDE loss, where $\beta_2, \beta_3, \beta_4$ = 1, results in a poorer performance compared to when the weights are smaller, $\beta_2, \beta_3, \beta_4$= $1\times10^{-2}$. Other combinations of $\beta$ are also tested; however, it is found that mainly the adversarial PDE loss $\mathcal{L}_{\text{PDE-ADV}}$ improves the convergence of the adversarial loss component. Using these findings, the number of random Fourier features is set to 8, raising the total input dimension to 34, and only the $\mathcal{L}_{\text{PDE-ADV}}$ is turned on inside the adversarial component. Next, the Gen-PINNs model and the standard PINNs are run for a total of 10000 iterations. Accordingly, the relative $L^2$ error for standard PINNs drops to 0.5% while for Gen-PINNs it drops to 0.015%. This is around a 30 times improvement using the Gen-PINNs model over the standard PINNs approach (Fig. 9-10). Moreover, comparing Fig. 22 in Appendix D.2 with Fig. 11 demonstrates that the optimized Gen-PINNs model overcomes the initial bias towards resolving earlier time trends and matches the solution well throughout the temporal dimension when looking at the space-averaged relative real and PDE errors.

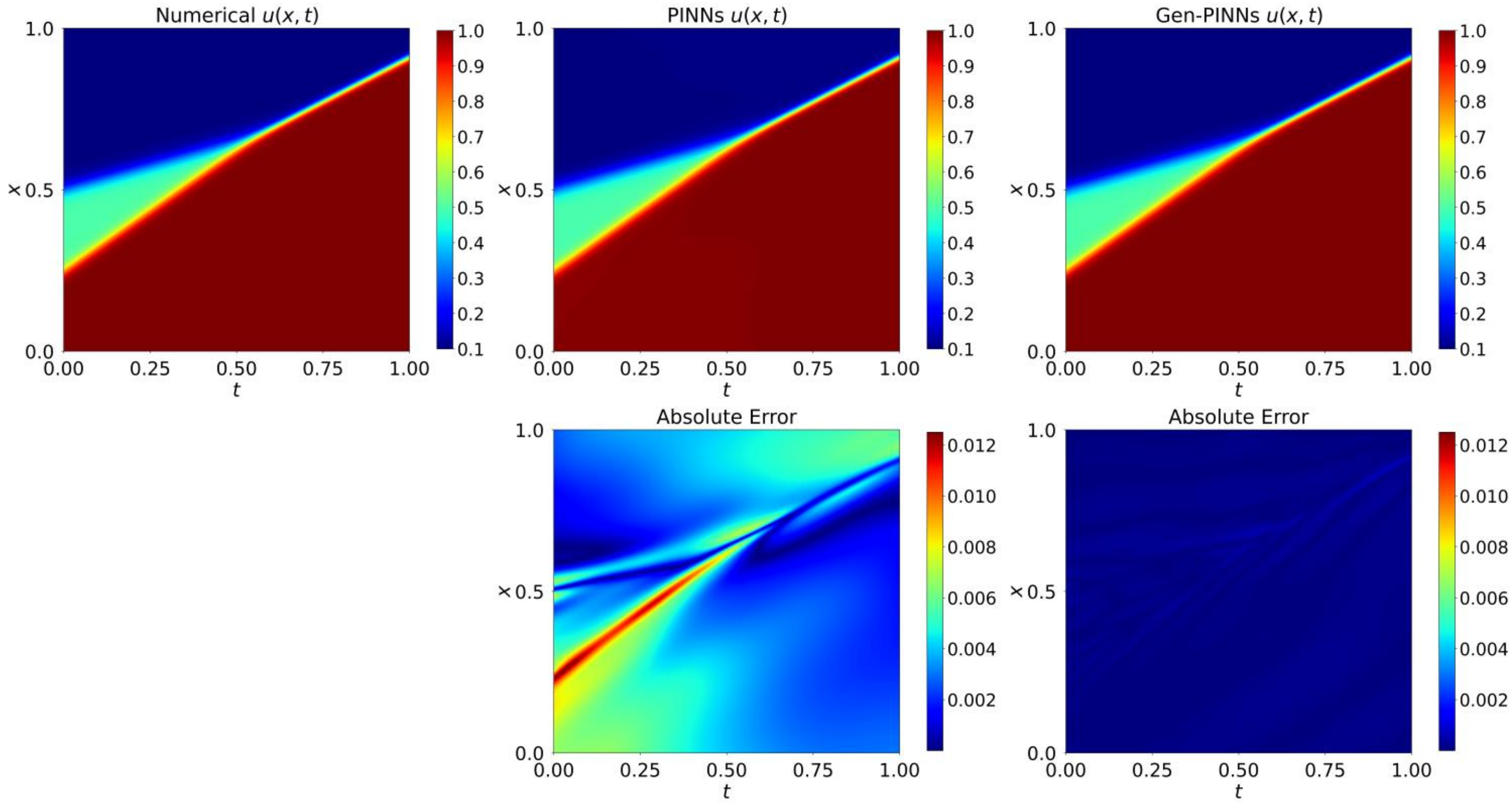

*Figure 9: Top Left: Numerical solution, Top Center: PINNs prediction, Top Right: Gen-PINNs prediction, Bottom Center: Absolute error in PINN prediction, Bottom Right: Absolute error in Gen-PINNs prediction for case 3 of the Burgers' PDE*

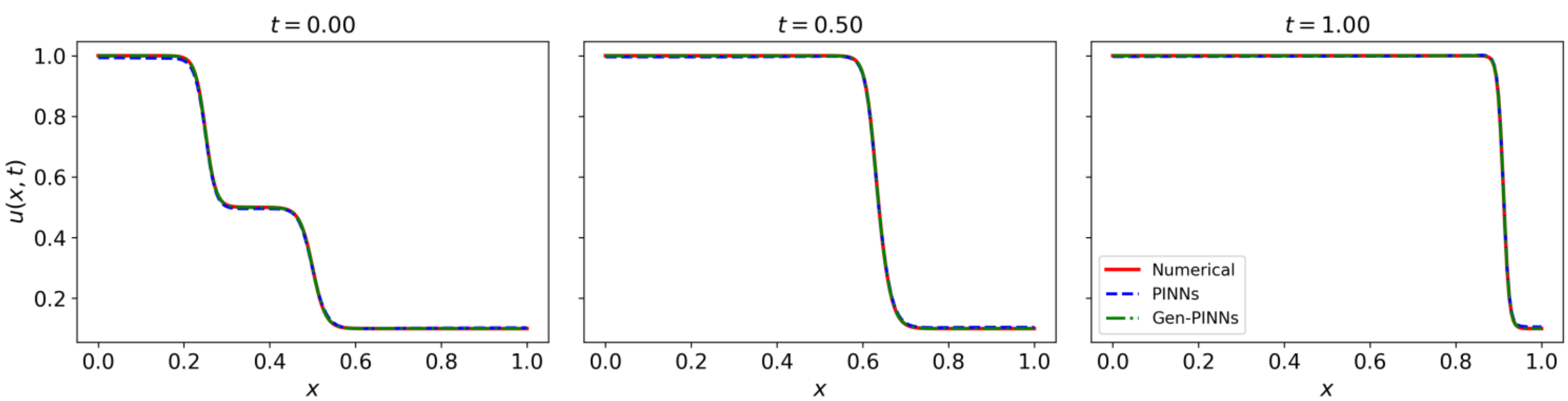

*Figure 10: Left: t=0, Center: t=0.5, Right: t=1 line plot comparison of the numerical solution (red), PINNs prediction (blue) and Gen-PINNs prediction (green) for case 3 of the Burgers' PDE*

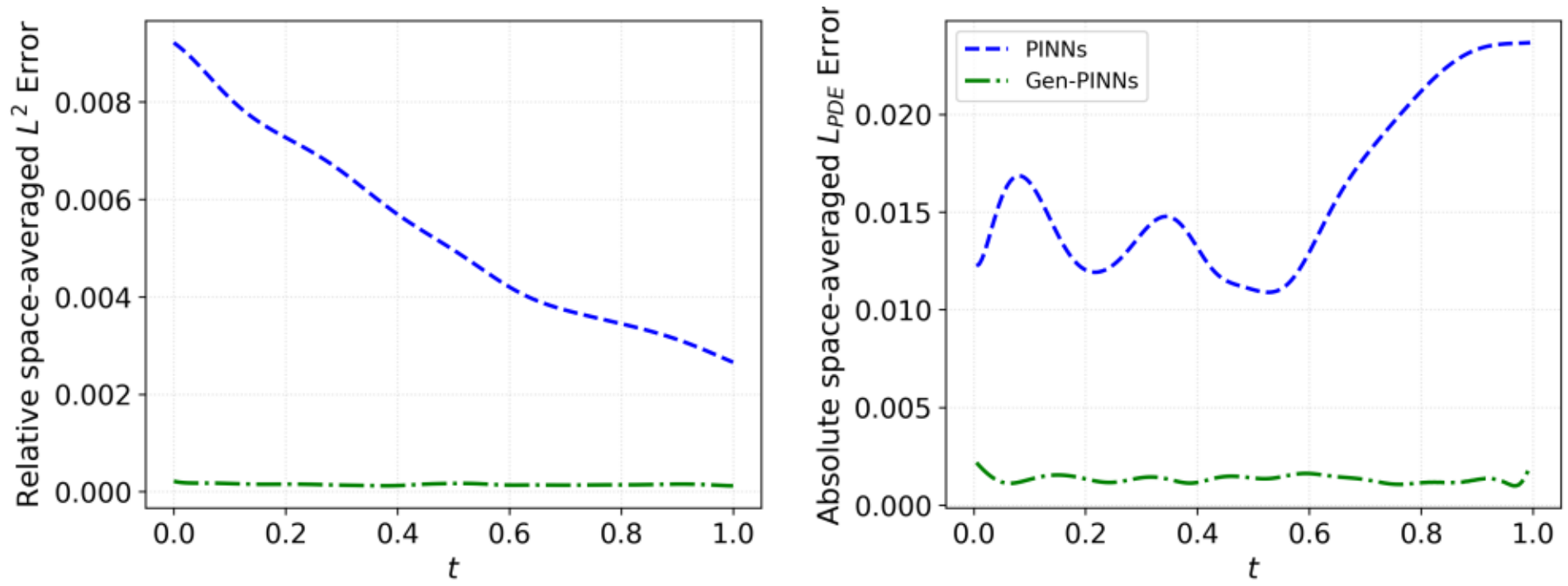

*Figure 11: Space-averaged error plots for standard PINNs (blue) and Gen-PINNs (green)*

## 4.3. Allen Cahn PDE

To further explore the different hyperparameter settings of Gen-PINNs we solve a pedagogical case of the Allen-Cahn PDE originally solved in Raissi et al. (2019) on a limited domain scale (Eq. 73-76). The PDE is challenging to solve with the very small diffusion coefficient, 0.0001, resulting in sharp gradient changes and thin interface layers. Several others accordingly solved the PDE with varying degrees of accuracy using different techniques, including Wight and Zhao (2020), McClenny and Neto (2020) & Mattey and Ghosh (2022). Most recently, Wang et al. (2024) showed significant improvements in modeling the PDE using a multi-NN causal setup by an order of magnitude over previous attempts. We solve the PDE numerically over the domain $t \in [0,1], x \in [-1,1]$ and use standard PINNs to train the model for $2\times10^5$ iterations with a fixed learning rate of $1\times10^{-3}$ and a large NN of the shape [2,128,128,128,128,1]. The relative $L^2$ error over the entire domain is 51.91%. The weights in this implementation are set to unity, no multiplicative support to improve ICs match with $\lambda_{\mathrm{IC}} = \lambda_{\mathrm{BC}} = \lambda_{\mathrm{PDE}} = 1$, which leads to a higher testing loss in standard PINNs.

$$\frac{\partial u}{\partial t} - 0.0001\frac{\partial^2 u}{\partial x^2} + 5u^3 - 5u = 0 \tag{73}$$

$$u(x,0) = x^2\cos(\pi x) \tag{74}$$

$$u(t,-1) = u(t,1) \tag{75}$$

$$\frac{\partial u}{\partial x}(t,-1) = \frac{\partial u}{\partial x}(t,1) \tag{76}$$

To investigate the performance of Gen-PINNs in modeling the Allen-Cahn PDE, we conduct a series of structured component and hyperparameter evaluations. Given the strongly causal nature of the solution, we observe a different pattern compared to the previous PDEs. In one of the runs, all the dynamic weights are turned on with $w_{IC}^{base} = w_{PDE-EXP}^{base} = 1 \times 10^2$, $w_{BC}^{base} = w_{ADV}^{base} = 1 \times 10^{-1}$ and $w_{PDE}^{base} = 1 \times 10^1$. A value of 0.9 is used for $\alpha_{\mathrm{VIMSE}}$, $\alpha_{\mathrm{exp}}$ is set to $1 \times 10^2$ , and only the PDE adversarial loss, $\beta_1 = 1$, is used. After training for $2\times10^5$ iterations, split equally between ADAM followed by the L-BFGS phase, only a few percentage points improvements is noted with a relative $L^2$ error of 47.9% (Appendix E.1). Despite several attempts at overcoming this error, it is realized that global PDE loss makes it difficult for the optimizer to converge, even if kept small. Keeping $w_{PDE}^{base} = 0$, we experiment with the values of $\alpha_{\mathrm{exp}}$ and observe that using large values greater than or equal to $1 \times 10^3$ for this PDE slows down training as the optimizer is too focused at resolving earlier time PDE losses without sufficiently resolving later time solutions. Using a value of $1 \times 10^1$ or smaller, on the other hand, significantly reduces the causal effect preventing sufficient convergence. As only a single NN is used in Gen-PINNs over the domain, it is critical to select an appropriate value that is kept constant $1 \times 10^2$ for further grid search.

Next, the PDE is run with $w_{PDE}^{base} = 0$ while keeping all remaining hyperparameters constant. This results in a reduction in the relative $L^2$ error to 21.2% with improvements in feature modeling (Appendix E.2). However, it is posited that the unrestricted causal update leads to optimizer pathology as the gradient pathway is too involved; the optimizer can manipulate the weights and weighted time residuals simultaneously and circumvent optimal gradient descent. This occurs when the weights are adjusted in such a way that initial time steps are weighted heavily with a low residual and later times consistently receive a lower weight to make the $\mathcal{L}_{\mathrm{PDE-EXP}}$ loss appears low while in reality the PDE loss is high. Using the stabilized causal pathway described in Section 3.1, the Gen-PINNs model is rerun. Even though only a single generator NN is used to predict the solution over the spatiotemporal domain, the stabilized causal approach improves optimization reducing the relative error to 5.2% (Appendix E.3).Finally, the Gen-PINNs model is again rerun with additionally all the adversarial PDE loss components now turned on where $\beta_1 = 1$ and $\beta_2, \beta_3, \beta_4 = 1 \times 10^{-2}$ .This results in a large improvement in the prediction with a relative $L^2$ error of 2.06% (Fig. 12-13). Within the broader Allen–Cahn sensitivity study, the final step is informative of the contribution of the residual adversarial component as the causal-gradient treatment and selected optimization settings improvement from previous run remain constant.

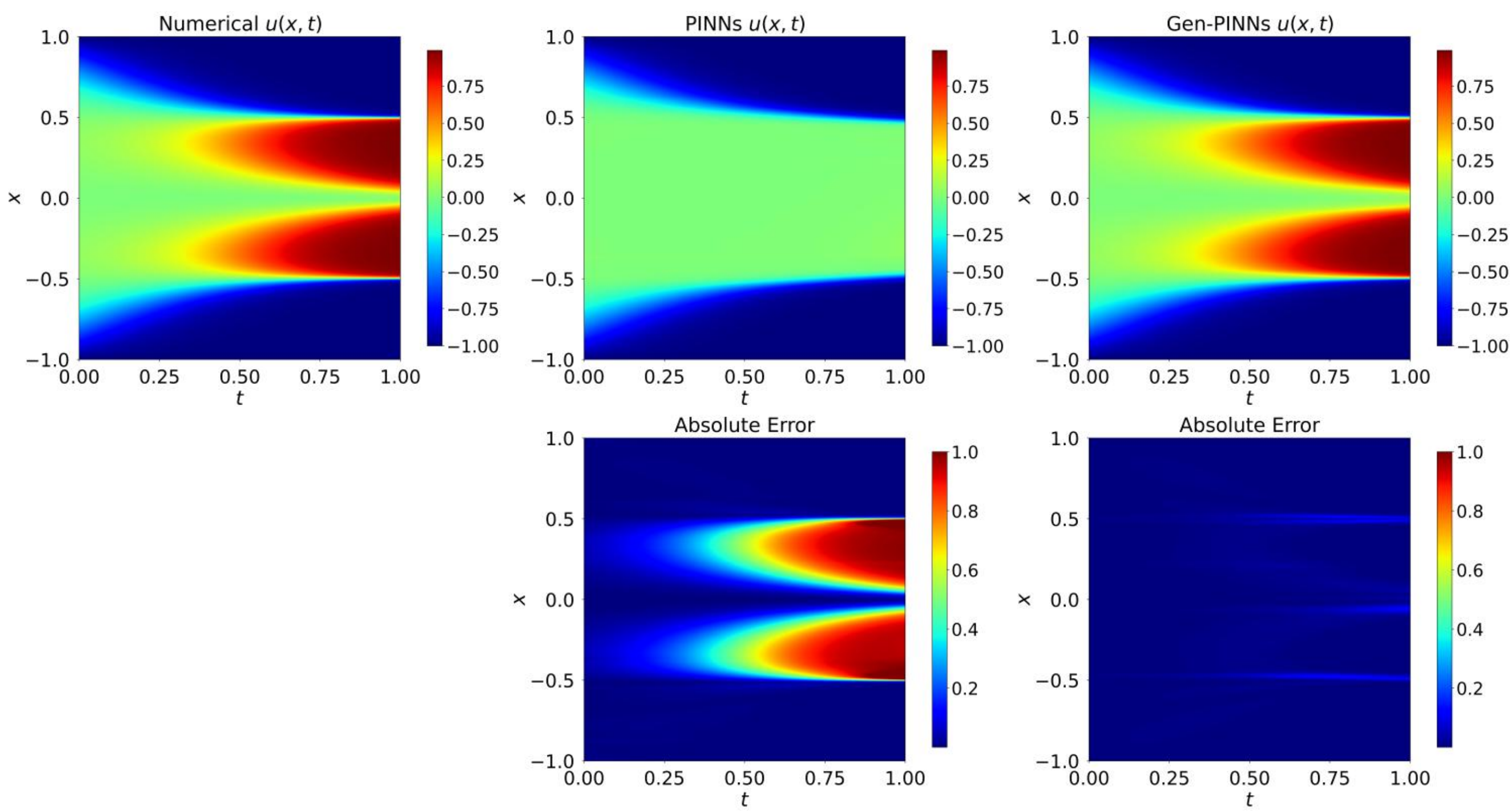

*Figure 12: Top Left: Numerical solution, Top Center: PINNs prediction, Top Right: Gen-PINNs prediction, Bottom Center: Absolute error in PINN prediction, Bottom Right: Absolute error in Gen-PINNs prediction for the Allen-Cahn PDE*

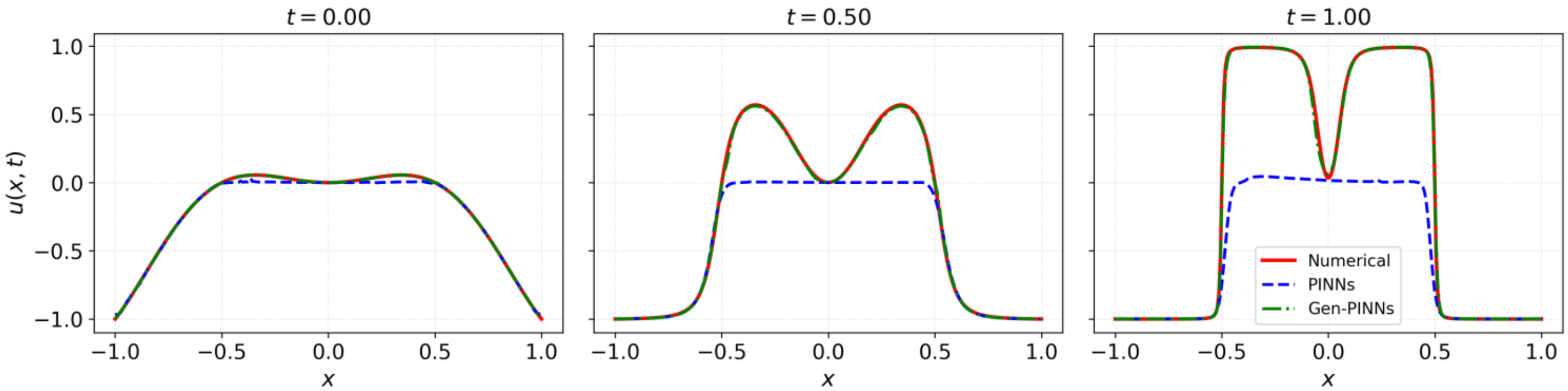

*Figure 13: Left: t=0, Center: t=0.5, Right: t=1 line plot comparison of the numerical solution (red), PINNs prediction (blue) and Gen-PINNs prediction (green) for the Allen-Cahn PDE*

## 4.4. Kuramoto-Sivashinsky PDE

The Kuramoto-Sivashinsky (KS) PDE is investigated next. The KS PDE is used to study and model non-linear spatiotemporal dynamics, including flame front propagation, reaction-diffusion systems, and thin film flow (Chang, 1994; Kuramoto, 1978; Michelson and Sivashinsky, 1977; Sivashinsky, 1977) (Eq. 77). The terms are defined as follows where α is a nonlinear convection coefficient, β is a linear diffusion coefficient and ε is a linear hyperviscosity coefficient. The Exponential Time Differencing Runge-Kutta 4th order method (ETDRK4) in native JAX is implemented to solve the PDE with a fixed inner time step of $10^{-4}$ with α = 1, β = $10^{-2}$, and ε = $10^{-4}$ over the domain $t \in [0,1], x \in [-1,1]$ and specified ICs and BCs such that the solution trend is similar to the 1st Burgers' PDE case in section 2 (Eq. 78-79). The 4th order derivative, $u_{xxxx}$, makes it particularly challenging to solve as the small linear hyperviscosity term is hard to resolve using the PDE residual in standard PINNs (Raissi, 2018).

$$u_t + \alpha u u_x + \beta u_{xx} + \varepsilon u_{xxxx} = 0 \tag{77}$$

$$u(x,0) = -\sin(\pi x) \tag{78}$$

$$u(t,-1) = u(t,1) \tag{79}$$

To investigate the KS PDE, a standard PINNs model is trained for 100,000 epochs. A relative $L^2$ error of 2.93% is reached. Next, a large grid search is performed to tune hyperparameters for the Gen-PINNs model. After several attempts at improving the equation modeling the following trends are observed. First, when the tertiary causal PDE loss is included, the $G$ NN fails to reduce the global PDE loss during training and subsequently results in a larger error in prediction. This is observed when the global PDE loss is also used or when the dynamic weights are assigned at the start of training. Also, if the $\alpha_{\text{exp}}$ is kept too high, which in this PDE case is $1 \times 10^2$, the solutions fail to propagate across time and only converge over a period of the time domain (Appendix F. 1). As $\alpha_{\text{exp}}$ is reduced, the match improves although still performs worse than standard PINNs even as the number of training iterations is increased.

The grid search is applied across many possible combinations of hyperparameters. However, despite many attempts to significantly lower the error, only marginal improvements are seen. For the optimized model, the same number of training iterations the Gen-PINNs model reaches a relative $L^2$ error of 2.52%, primarily from improved modeling of the solution at the boundaries (Fig.14,15). In this case, the tertiary causal exponential PDE loss is removed and $\beta_2$= 1×10$^{-2}$ and $\beta_1, \beta_3, \beta_4 = 0$ which is a type of Sobolev-weight loss only enforcement through the adversarial component. The adversarial component is also held constant throughout with a small multiplier, while the rest of the weights are kept dynamic(see Table. 5, Appendix. F.2 for summary of sensitivity study). We posit that the 4$^{th}$ order linear hyperviscosity term $\varepsilon u_{xxxx}$ may result in very different values over the domain, which is difficult to resolve through gradient changes calculated with backpropagation through the activation functions. To overcome this, we implement the truncated Taylor polynomial estimate of gradient values using the JAX.jet module. Attempting this only results in small changes in the error. Moreover, the Gen-PINNs model uses a single NN which seems to struggle when capturing higher-order terms in the entire time domain, even when the causal loss is used. It is suggested that using a multi-GAN approach with time-based binning may work to improve the case modeling. In such a model, the final output of each generator, except for the generator at ICs, would be used as an initial condition to the next generator in an autoregressive way, such that each generator only needs to model a small portion of the final solution, possibly leading to a better prediction.

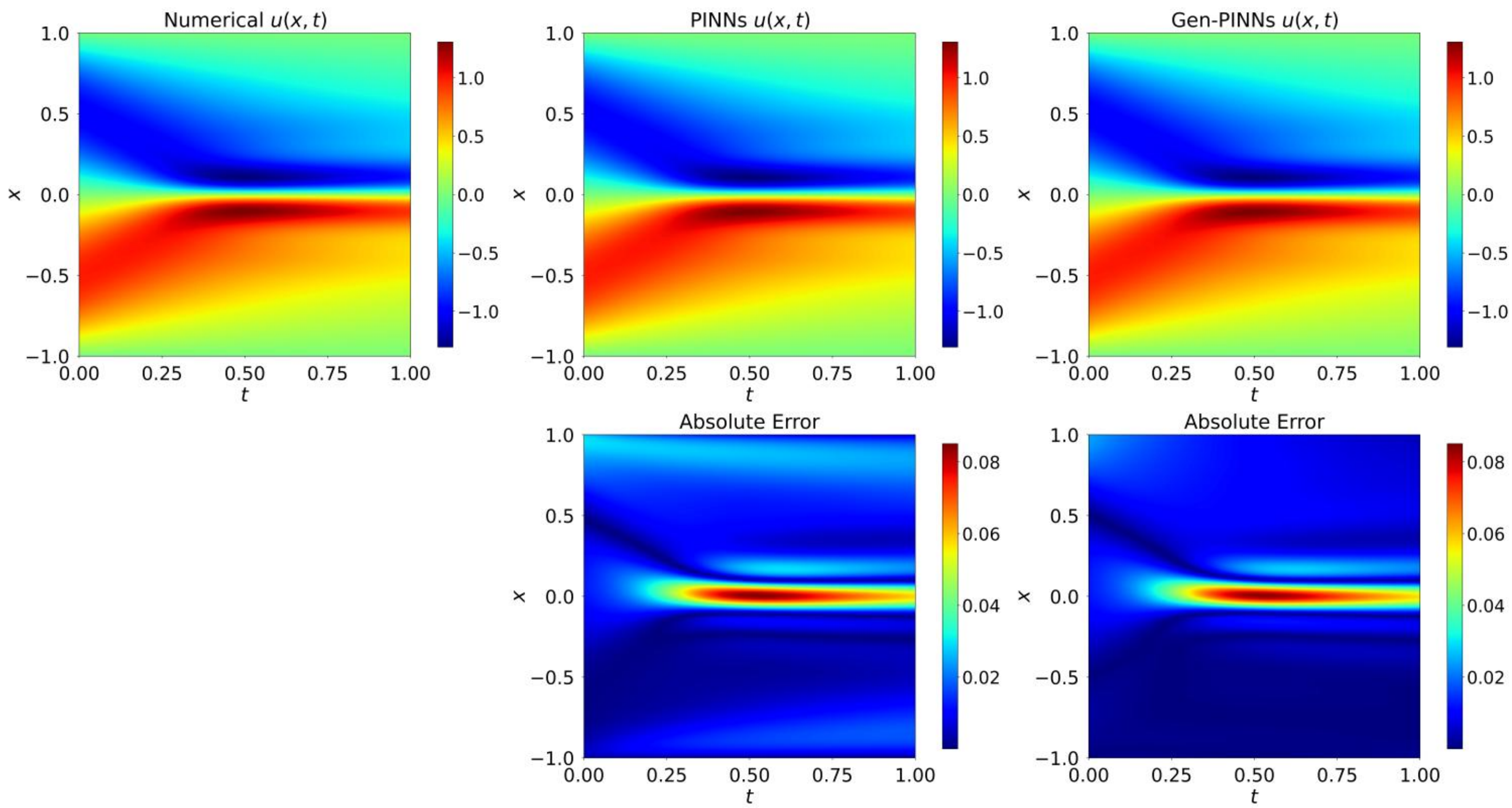

*Figure 14: Top Left: Numerical solution, Top Center: PINNs prediction, Top Right: Gen-PINNs prediction, Bottom Center: Absolute error in PINN prediction, Bottom Right: Absolute error in Gen-PINNs prediction for the Kuramoto-Sivashinsky PDE*

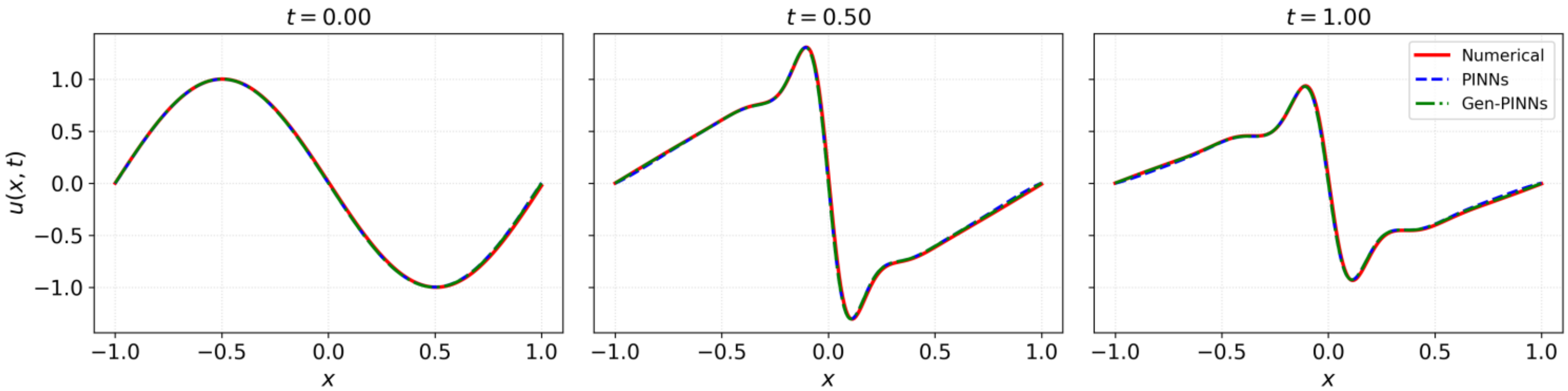

*Figure 15: Left: t=0, Center: t=0.5, Right: t=1 line plot comparison of the numerical solution (red), PINNs prediction (blue) and Gen-PINNs prediction (green) for the Kuramoto-Sivashinsky PDE*

## 4.5 Gen-PINNs component ablation and sensitivity analysis

As observed across the Gen-PINNs experiments, some of which are discussed above, the framework is evaluated through the series of sequential ablation and sensitivity tests. As listed in Table. 1 there are around 24 framework and configuration settings which can take multiple values or are relevant to specific PDE solution characteristics. Performing a complete factorial ablation study would therefore require an impractically large number of combinations including many configurations that are not relevant for particular PDEs. Therefore, the current analysis focuses on detailing a summary of how many of the established PINN strategies are integrated within the residual adversarial framework of Gen-PINNs and outlining under which solution characteristics they perform best. These are based on the theoretical and observed performance improvements in addressing spectral, temporal, loss-balancing and optimization developed for the Gen-PINNs current formulation. The tests are aimed at evaluating individual component effects or small group of related settings while retaining the remaining configuration as best as possible and thus allowing the effects of the principial representation, loss, adversarial, causal, and optimization choices to be assessed under different PDE solution conditions. Table. 2 is constructed to consolidate the most important observations in these evaluations and provides a direct summary for configuration change and the resulting solution accuracy. A more extensive sensitivity analysis for the KS equation is additionally provided in Appendix F.2, where Table. 5 summarizes 15 representative configurations selected from a few hundred runs.

*Table 2: Summary of component ablation and sensitivity evaluations for Gen-PINNs*

| PDE case | Component evaluated | Configuration change | Relative $L^2$ error (%) | | Findings |
|---|---|---|---|---|---|
| | | | Before | After | |
| Burgers' case 2 | Training duration | 1,000 to 2,000 iterations; all else fixed | 1.42 | 0.91 | Selected configuration continues to improve with additional optimization runs |
| Burgers' case 2 | Adversarial residual features | Add auxiliary residual features at low weights | 0.91 | 0.88 | Balanced residual-feature augmentation provide an additional accuracy gain |
| Burgers' case 2 | Adversarial feature weights | Increase auxiliary feature weights to unity | 0.88 | 1.38 | Overweighting auxiliary residual features degrades convergence |
| Burgers' case 2 | Extended optimization | 2,000 to 1×10^5 iterations; restore lower auxiliary weights | 0.88 | 0.77 | Longer optimization produces further late-stage accuracy gains as observed earlier |
| Burgers' case 3 | Causal weighting | Transfer causal configuration from Burgers case 2 | - | 11 | Causal weighting creates early-time bias and inhibits later-time propagation compared to standard PINNs (2.02% relative $L^2$ error) under similar settings |
| Burgers' case 3 | Temporal, spectral, and adversarial configuration | Causal OFF; Fourier ; direct PDE residual retained | 11 | 0.015 | Problem-informed reconfiguration removes temporal bias and sharply improves accuracy |
| Allen–Cahn | Global PDE loss | $\mathcal{L}_{\text{PDE}}$ OFF; all else constant | 47.9 | 21.2 | Global and causal PDE objectives compete in this strongly causal problem, removing the residual loss allows causal loss to improve |
| Allen–Cahn | Causal gradient pathway | Unrestricted to Stabilized causal gradients | 21.2 | 5.2 | The stabilized causal pathway helps to reduce the optimization pathology observed with the unrestricted pathway when solving strongly causal PDEs. |
| Allen–Cahn | Adversarial residual features | All adversarial residual features added after causal stabilization | 5.2 | 2.06 | Residual-adversarial features provide a substantial final accuracy gain |
| KS | Causal weighting strength | Reduce $\alpha_{\text{exp}}$; all else fixed | 45.3 | 11.2 | Excessive causal weighting suppresses temporal propagation, reducing effects thus improves solution |
| KS | Temporal and adversarial configuration | Causal OFF; retain low-weight Sobolev-type adversarial feature | 11.2 | 2.52 | A noncausal, Sobolev-focused configuration better resolves the fourth-order problem with slight improvements |
| KS | Higher-order derivative treatment | Conventional AD to the truncated Taylor JAX.jet evaluation of derivatives | - | Minor change | Derivative evaluation alone is not the dominant limitation, multi Gen-PINNs models are a possible direction for improving match in KS-type problems |

As seen across the Burgers' cases, the ablation results demonstrate that the effectiveness of the type of PDE residual used in the loss is strongly affected by the underlying temporal and spectral characteristics of the solution. For example, using a causal weight improves the convergence in Burgers' case 2 where the final solution is strongly causal however, when applied to Burgers' case 3 results in an early time bias preventing later time solution propagation. In Burgers' case 3 removing the causal loss term and using the conventional PDE residual along with Fourier feature embeddings extended in space and time thus improved the overall solution. These observations are consistent with the expected role of causal weighting as a potential temporal optimization mechanism rather than a universally beneficial loss component always.

Next, for the Allen-Cahn PDE it is observed that removing the competing PDE residual greatly helps overcome optimization stagnation. In this case, as the PDE is strongly causal, the unrestricted causal gradient pathway in the exponential PDE residual results in a large initial drop in the relative error but does not fully resolve the later-time errors. To address this, the stabilized causal pathway is applied and results in a significant further error reduction even when using a single NN over the entire spatiotemporal domain. Finally, once the dominant optimization pathologies are addressed, adding the adversarial-feature PDE loss further reduces the relative error by more than half in the final solution.

Finally, KS experiments provide a useful challenging limiting case for the current Gen-PINNs implementation. The sensitivity studies show that causal weighting can hinder temporal propagation in this higher-order PDE, and reducing its strength improves convergence but remains insufficient to outperform standard PINNs. In the best performing configuration, the causal PDE loss is therefore removed and a small Sobolev-type adversarial contribution is retained, consistent with the rapidly evolving structure of the $4^{th}$ order PDE. This helped marginally reduce the relative error below standard PINNs.

We also tested JAX.jet formulation to reduce overhead of derivative calculations however, that had minimal changes in the final error. The current single-generator NN structure of Gen-PINNs is therefore identified as a limitation for this case and may benefit from a multi-generator, temporally decomposed formulation in which each generator resolves a smaller portion of the time domain. Across the broader case studies, the residual-adversarial pathway plays a pivotal role in the learned physics feedback mechanism, while the temporal, representation, and loss configuration components support the framework under different PDE characteristics. Accordingly, the ablation and sensitivity studies are interpreted as evidence of component behavior within the current Gen-PINNs framework, rather than as establishing causal contribution or universal applicability of individual components across all possible PDE cases.

## 4.6. Performance comparison for Standard PINNs and Gen-PINNs

To train the models we use an .ipynb notebook running on a single NVIDIA A100 40GB GPU on Google Colaboratory. In general, the Standard PINNS models achieve a higher training throughput under same hardware settings as compared to Gen-PINN. The additional computational cost in the Gen-PINNS model stems from many factors including residual-adversarial evaluations, discriminator forward and backward passes and the additional optimization components set activated for a given PDE. Despite the larger run training times the Gen-PINNs model consistently perform better reducing the relative $L^2$ errors in testing. The average per epoch cost of the runs in each model is thus used to compare against improvements in error reduction. For most PDEs a greater factor in reduction of error compared to the additional computational cost. The recorded training throughput and corresponding relative $L^2$ errors for the final configurations of each PDE are summarized in Table. 3.

*Table 3: Computational performance of the final standard PINN and Gen-PINNs configurations for the evaluated PDE cases.*

| PDE case | Standard PINNs avg. epochs/s | Gen-PINNs avg. epochs/s | Standard PINNs relative $L^2$ error | Gen-PINNs relative $L^2$ error |
|---|---|---|---|---|
| Burgers case 1 | 31.83 | 2.55 | 35.41% | 2.99% |
| Burgers case 2 | 51.36 | 3.07 | 50.63% | 1.42% |
| Burgers case 3 | 65.67 | 25.13 | 0.508% | 0.0147% |
| Allen-Cahn | 120.18 | 6.14 | 51.9% | 2.06% |
| KS | 29.21 | 6.69 | 2.93% | 2.52% |

The results show that across the Burgers and Allen–Cahn PDEs, the lower training throughput of Gen-PINNs is accompanied by substantial reductions in the relative $L^2$error. For Burgers' case 1, the average training throughput decreases from 31.83 epochs/s to 2.55 epochs/s, while the relative $L^2$ error decreases from 35.41% to 2.99%, corresponding to an 11.84-fold reduction in error. For Burgers' case 2, the throughput decreases from 51.36 epochs/s to 3.07 epochs/s, while the relative $L^2$ error decreases from 50.63% to 1.42%, corresponding to a 35.65-fold reduction. In Burgers' case 3, a smaller reduction in throughput is observed, from 65.67 epochs/s to 25.13 epochs/s, while the relative $L^2$ error decreases substantially from 0.508% to 0.0147%, corresponding to a 34.56-fold reduction. Similarly, for the Allen–Cahn PDE, the training throughput decreases from 120.18 epochs/s to 6.14 epochs/s, while the relative $L^2$ error decreases from 51.9% to 2.06%, corresponding to a 25.19-fold reduction. Consequently, although Gen-PINNs requires greater computational effort in terms of training duration, the substantial improvements in solution accuracy offer a favorable computational trade-off for PDEs specially in cases where standard PINNs may experience significant optimization difficulty even when using large training budgets.

The KS PDE results outline a limiting returns component of using the Gen-PINNs model. The average training throughput decreases from 29.21 epochs/s for standard PINNs to 6.69 epochs/s for Gen-PINNs, while the relative $L^2$ error only improves slightly from 2.93% to 2.52% despite an extensive optimization effort. In this case, the additional computational cost produces a smaller accuracy gain, consistent with the current limitations of the single generator Gen-PINNs formulation discussed in Section 4.4 with suggested remedies for future models. Overall, the computational comparison indicates that the additional cost of the Gen-PINNs model is most beneficial for PDE cases where the standard PINN stagnates at high training errors or shows significant optimization pathologies as corroborated by earlier PINN extension studies.

## 5. Conclusion

Modeling PDEs using PINNs is an important frontier in machine learning albeit a challenging task as convergence is strongly dependent on the underlying PDE solution structure. We identified several causes of PINN modeling failures when solving a variety of PDEs. These include an inherent strong spectral bias towards resolving less complex low frequency instead of more complex high frequency solution features commonly encountered in the solution of stiff-PDEs. Standard PINNs use fixed weights over multiple loss terms which introduce bias and prevents resolution of each of the terms. This leads to possible oscillations in training and slower convergence when using first order optimizers along with the standard MSE loss function. We posit that the combined effect of each of the standard PINN limitations prevents smooth convergence when solving PDEs. To resolve these multiple training pathologies of standard PINNs we propose a unified residual-adversarial framework incorporating multiple supporting PINN strategies. The residual adversarial pathway enables multiple learned physics based feedback signals, while the supporting components address specific representation and optimization challenges. Using this approach, the Gen-PINNs models demonstrate a large improvement over the standard PINNs approach reaching on average an order of magnitude reduction when measured using the relative $L^2$ error metric. Overall, Gen-PINNs constructs a unified residual-adversarial approach that improves PDE solution accuracy across nonlinear, stiff, sharp-front, and higher-order PDEs considered in the study. The results show that combining adversarial residual learning with structured inclusion of temporal, representation, weighting, and optimization strategies can produce more stable and accurate convergence than standard PINNs across several different solution regimes.

Equations with sharp gradient changes are considered, including Burgers'-type cases, the Allen-Cahn equation and the Kuramoto-Sivashinsky PDE. These problems are solved using the standard PINNs and Gen-PINNs framework under similar network architecture and total training iterations. The proposed Gen-PINNs model consistently performs better across low and high-order PDEs, accurately capturing different scale structures in space and time. For lower-order PDEs, it is observed that the standard PINNs model eventually converges by using a large number of training iterations. However, for higher-order PDEs with stiff dynamics the standard PINNs model struggles to learn the underlying solution even if trained over many iterations. The proposed Gen-PINNs model addresses this by incorporating multi-component optimization pathways that can effectively model PDEs with shock-fronts, thin interfaces and sharp gradient changes. For example, in the Allen-Cahn PDE, the standard PINN reaches a relative $L^2$ error of 51.9% compared to 2.06% using Gen-PINNs. Admittedly, the single NN structure used in Gen-PINNs performs only marginally better than standard PINNs for the KS equation where the $4^{th}$ order term with very small values is challenging to resolve. We posit that using a multi-NN approach using an autoregressive technique may help overcome this modeling constraint.

Although the residual-adversarial framework of Gen-PINNs is flexible, depending on the problem, the forward equation solver may require structured configuration evaluations to reach an accurate solution for each PDE. This sometimes makes optimization challenging given the number of tunable components including the hyperparameters listed in Table.1 and additional settings described in section 3.1. As a result, Gen-PINNS may be computationally demanding for large-scale problems. Furthermore, the preferred configuration of the Gen-PINNs model depends on the underlying solution structure of the PDE. If the reference solution is unknown, selecting the problem-informed components becomes more challenging because the relevant spectral, temporal, boundary and thus the residual characteristics of the PDE solution may not be known as a priori. Table.1 and Algorithm.1 serve as structured guidance for making configuration settings based on available PDE solution characteristics and observable training behavior; however, automated component and hyperparameter selection remains an important direction for future work to reduce computational configuration effort of the Gen-PINNS model. .

Despite these limitations, the proposed modeling paradigm offers a useful basis for improving physics-informed machine learning methods where standard PINNs require prolonged training or fail to converge when solving a PDE. The modular structure of Gen-PINNS framework also enables further development in areas such as automated hyperparameter tuning, adaptive collocation strategies, and multi-dimensional modeling. Further work could explore ways of developing a single loss term that combines the known data at ICs and BCs with the PDE residual using a unified approach, as the current modified version of GradNorm combines loss terms, each measuring a different aspect of the solution error. One possible area of such exploration is training solely in the Fourier domain, where the effect of each of the PDE terms, such as the $4^{th}$ order linear hyperviscosity term, is encoded in the same frequency-based space. When combined with information at the ICs and BCs, such a formulation may provide a more consistent spectral measure for optimizing higher order PDEs. Within the nonlinear, stiff, sharp-fronts and higher order PDE regimes evaluated in this study the Gen-PINNs model demonstrates improved accuracy and convergence compared to standard PINNs.. Overall, the results establish the potential of deterministic residual-adversarial learning as an effective framework for physics-informed PDE solutions where complementary PINN stabilization strategies providing additional flexibility for problem-specific training challenges. Future work will focus on extending the framework to multi-dimensional PDEs, developing automated problem-informed configuration strategies, and reducing the associated computational costs associated with the comprehensive optimization process We believe that Gen-PINNs is a high fidelity framework and is an important step towards developing models that are interpretable, accurate, and robust for SciML and in practical scientific applications and engineering.

**Acknowledgements**

The authors thank the Reservoir Simulation Joint Industry Project (RSJIP) and the Digital Rock Characterization Technology (DIRECT) Industry Affiliate Program at the University of Texas at Austin for supporting this work.

**CRediT authorship contribution statement**

**Muhammad Muneeb Akmal:** Conceptualization, Methodology**,** Investigation, Formal analysis, Writing – original draft, Writing – review & editing, Software, Visualization, Validation, Data Curation. **Kamy Sepehrnoori:** Conceptualization, Writing – review & editing, Methodology, Visualization, Supervision, Project administration, Investigation, Formal analysis, Conceptualization, Funding acquisition. **Michael J. Pyrcz:** Conceptualization, Writing – review & editing, Methodology, Visualization, Supervision, Project administration, Investigation, Formal analysis, Conceptualization, Funding acquisition.

**Declaration of competing interest**

The authors declare that they have no known competing financial interests or personal relationships that could have appeared to influence the work reported in this paper.

**Data availability**

Data will be made available on request.

# Appendix

A. Summary of hyperparameters used in PDE experiments are listed below in Table. 4

*Table 4: Hyperparameter table for the final PDE experiments*

| | PDE | | | |
|---|---|---|---|---|
| Hyperparameter | Burgers' 2nd PDE | Burgers' 3rd PDE | Allen-Cahn PDE | KS PDE |
| Weight Initialization | Xavier | Xavier | Xavier | Xavier |
| Learning rate | $1\times10^{-3}$ | $1\times10^{-3}$ | $1\times10^{-3}$ | $1\times10^{-3}$ |
| Activation type | Tanh | Tanh | Tanh | Tanh |
| Number of Fourier frequencies in $\boldsymbol{G}$ | 10 | 8 | 8 | 4 |
| Fourier Dimensions | $x$ | $x, t$ | $x$ | $x, t$ |
| $\boldsymbol{G}$ NN structure | [22,128,128,128,128,1] | [34, 128,128,128,128,1] | [18,128,128,128,128,1] | [18,128,128,128,128,1] |
| $\boldsymbol{D}$ NN structure | [3,64,64,64,1] | [3,64,64,64,1] | [3,128,128,128, 128,1] | [3,64,64,64,1] |
| $\boldsymbol{G}:\boldsymbol{D}$ update frequency | 1:10 | 1:10 | 1:1 | 1:10 |
| Optimizer Stage 1 | ADAM | ADAM | ADAM | ADAM |
| Number of iterations Stage 1 | 250 | 5000 | 100000 | 90000 |
| Optimizer Stage 2 | L-BFGS | L-BFGS | L-BFGS | L-BFGS |
| Number of iterations Stage 2 | 250 | 5000 | 100000 | 10000 |
| History size | 5 | 10 | 5 | 5 |
| $\boldsymbol{\alpha_{VIMSE}}$ | 0.95 | 0.9 | 0.9 | 0.9 |
| Dynamic Weights | ON | OFF | OFF | OFF |
| Exponential PDE Loss | ON | OFF | ON | OFF |
| Gradient pathway | Unrestricted Causal | N/A | Stabilized Causal | N/A |
| $\boldsymbol{\alpha}_{\mathbf{exp}}$ | $1\times10^{3}$ | 0 | $1\times10^{2}$ | 0 |
| $\boldsymbol{w}_{\boldsymbol{IC}}^{\text{base}}$ | $1\times10^{1}$ | $1\times10^{2}$ | $1\times10^{2}$ | $1\times10^{2}$ |
| $\boldsymbol{w}_{\boldsymbol{BC}}^{\text{base}}$ | $1\times10^{-1}$ | $1\times10^{0}$ | $1\times10^{-1}$ | $1\times10^{0}$ |
| $\boldsymbol{w}_{\boldsymbol{PDE}}^{\text{base}}$ | $1\times10^{0}$ | $1\times10^{1}$ | 0 | $1\times10^{1}$ |
| $\boldsymbol{w}_{\boldsymbol{ADV}}^{\text{base}}$ | $1\times10^{-1}$ | $1\times10^{-2}$ | $1\times10^{-1}$ | $1\times10^{-2}$ |
| $\boldsymbol{w}_{\boldsymbol{PDE-EXP}}^{\text{base}}$ | $1\times10^{1}$ | 0 | $1\times10^{2}$ | 0 |
| $[\boldsymbol{\beta}_1, \boldsymbol{\beta}_2, \boldsymbol{\beta}_3, \boldsymbol{\beta}_4]$ | [1,0,0,0] | [1,0,0,0] | $[1, 1\times10^{-2}, 1\times10^{-2}, 1\times10^{-2}]$ | $[0, 1\times10^{-2}, 0, 0]$ |

B. Mathematical formulation of the deterministic residual adversarial formulation

The mathematical interpretation of the residual-adversarial formulation introduced in Section 3.1 is analyzed here. The analysis focuses on the sensitivity of the PDE residual transformations defined in Eq. 42-45 and the respective contribution to the generator gradient, and the physics-informed approach of the LS-GANs formulation defined in Eq. 46-48 and used in the Gen-PINNs model.

B.1. Generator parameters update sensitivity to residual features

The residual features as listed in Eq. 42-45 provide different transformations applied to the standard PDE residual $f_\theta(x,t)$. At an interior collocation point $\left(x_f^i, t_f^i\right)$ the residual is calculated as

$$f_i = f_\theta\left(x_f^i, t_f^i\right) \tag{80}$$

The local sensitivities with respect to $f_i$ for the squared, cubic, and quartic residual features are calculated as a simple first derivative. The corresponding gradients with respect to the generator parameters, $\theta$, is thus calculated as

$$\nabla_\theta f_i^2 = 2 f_i \nabla_\theta f_i \tag{81}$$
$$\nabla_\theta f_i^3 = 3 f_i^2 \nabla_\theta f_i \tag{82}$$
$$\nabla_\theta f_i^4 = 4 f_i^3 \nabla_\theta f_i \tag{83}$$

Comparing the squared residual to the cubic and quartic transformations the sensitivity of the update thus increases as $|f_i|$ increases. Similarly, the $H^1$ residual feature acts on spatial variation in the residual and its generator gradient is slightly different and calculated as

$$\nabla_\theta (\partial_x f_i)^2 = 2(\partial_x f_i)\nabla_\theta(\partial_x f_i) \tag{84}$$

The four residual features thus enable different aspect sensitivities to the PDE residual in terms of magnitude, sign, and spatial variation to affect the generator parameter update which is very limited if only the standard PDE residual is used. Furthermore, as the higher-order features vary significantly with residual scale, the respective relative contributions are controlled through the coefficients $\beta_s$ multiplier described in Eq. 46.

B.2. Generator gradient of the residual adversarial objective

Next, during a generator update, the discriminator parameters are held fixed. Applying the chain rule to Eq. 47, with $\gamma_i = \gamma\left(x_f^i, t_f^i\right)$ defined for compactness leads to

$$\nabla_\theta L_{\mathrm{ADV}} = 2\omega \sum_{s=1}^{S} \sum_{i=1}^{N_f} \left[D_s\left(\beta_s F_s\big(G_\theta(\gamma_i)\big)\right) - 1\right] \frac{\partial D_s\left(\beta_s F_s\big(G_\theta(\gamma_i)\big)\right)}{\partial\left(\beta_s \phi_s[u_\theta]\left(x_f^i, t_f^i\right)\right)} \beta_s \nabla_\theta \phi_s[u_\theta]\left(x_f^i, t_f^i\right) \tag{85}$$

Where $D_s\left(\beta_s F_s\big(G_\theta(\gamma_i)\big)\right)$ is the discriminator output. The terms in Eq. 85 follow from the discriminator input $\left[\beta_s \phi_s[u_\theta]\left(x_f^i, t_f^i\right), x_f^i, t_f^i\right]$ as defined in Eq. 46. Next, the discriminator output is differentiated against the generator parameters update by applying the chain rule with respect to $\theta$. Ast the discriminator is composed of three components the derivative is written as

$$\nabla_\theta D_s\left(\beta_s F_s\left(G_\theta(\gamma_i)\right)\right) = \frac{\partial D_s}{\partial\left(\beta_s \phi_s[u_\theta]\left(x_f^i, t_f^i\right)\right)} \nabla_\theta\left(\beta_s \phi_s[u_\theta]\left(x_f^i, t_f^i\right)\right) + \frac{\partial D_s}{\partial x_f^i} \nabla_\theta x_f^i + \frac{\partial D_s}{\partial t_f^i} \nabla_\theta t_f^i \tag{86}$$

However as $x_f^i, t_f^i$ are fixed input locations and do not change when the generator parameters are updated the effective contribution of the 2nd and 3rd derivative is zero, $\nabla_\theta x_f^i = 0,\ \nabla_\theta t_f^i = 0$. Equivalently, only the residual feature, $\beta_s \phi_s[u_\theta]\left(x_f^i, t_f^i\right)$, changes during the generator update. As $\beta_s$ is a constant Eq. 86 becomes

$$\nabla_\theta D_s\left(\beta_s F_s\left(G_\theta(\gamma_i)\right)\right) = \frac{\partial D_s}{\partial\left(\beta_s \phi_s[u_\theta]\left(x_f^i, t_f^i\right)\right)} \beta_s \nabla_\theta \phi_s\left[u_\theta\left(x_f^i, t_f^i\right)\right] \tag{87}$$

Next, define $e_{s,i} = D_s\left(\beta_s F_s\left(G_\theta(\gamma_i)\right)\right) - 1$ as the adversarial mismatch measuring how far the discriminator prediction is from generator target of 1. Similarly, define the local sensitivity of the discriminator the residual feature as $q_{s,i}$ where

$$q_{s,i} = \frac{\partial D_s}{\partial\left(\beta_s \phi_s[u_\theta]\left(x_f^i, t_f^i\right)\right)} \tag{88}$$

Then the generator gradient in Eq. 85 can be written as

$$\nabla_\theta L_{\mathrm{ADV}} = 2\omega \sum_{s=1}^{S} \sum_{i=1}^{N_f} e_{s,i}\ q_{s,i}\, \beta_s \nabla_\theta \phi_s[u_\theta]\left(x_f^i, t_f^i\right) \tag{89}$$

Therefore, following Eq. 89 it is clear that the contribution of each residual feature to the generator parameter update is controlled by four terms that include: the current adversarial mismatch $e_{s,i}$, the local discriminator sensitivity $q_{s,i}$, the respective weight multiplier of each residual feature influence of each physics feature $\beta_s$ and the residual feature gradient $\nabla_\theta \phi_s[u_\theta]$ at the specific coordinate $\left(x_f^i, t_f^i\right)$. An effective local adversarial weighting can thus be written as

$$\lambda_{s,i}^{\mathrm{ADV}} = 2\omega \beta_s e_{s,i} q_{s,i} \tag{90}$$

Where the residual adversarial generator gradient becomes

$$\nabla_\theta L_{\mathrm{ADV}} = \sum_{s=1}^{S} \sum_{i=1}^{N_f} \lambda_{s,i}^{\mathrm{ADV}}\, \nabla_\theta \phi_s[u_\theta]\left(x_f^i, t_f^i\right) \tag{91}$$

Unlike in the standard PDE loss, where the residual gradient is controlled by a predefined fixed weight, the adversarial gradient is controlled by the dynamic learned responses of the corresponding discriminators used. As the $D_s$ also receive the coordinates $\left(x_f^i, t_f^i\right)$, the local sensitivity $q_{s,i}$ and therefore $\lambda_{s,i}^{\mathrm{ADV}}$ can vary throughout the spatiotemporal domain. This allows the generator updates to place a larger focus on the residual features and the respective coordinates where the physics informed loss is higher as opposed to applying the same penalty at all the collocation points.

B.3. Physics consistent adversarial target

The learned adversarial weight $\lambda_{s,i}^{\text{ADV}}$ determines the contribution of each residual feature to the generator update at a given collocation point. The direction of the adversarial update is set to minimize $\mathcal{L}_{\text{ADV}}$ (Eq. 47), where the generator seeks to trick the discriminator prediction to approach the real label probability of 1. In the deterministic residual adversarial formulation of Gen-PINNs, the real target state corresponds to the zero residual physics state as defined in Eq. 48.

Accordingly, as the training progresses the generator prediction increasingly satisfies the governing PDE at all interior collocation points, $f_\theta\left(x_f^i, t_f^i\right) \to 0,\ i = 1, \dots, N_f$, where all active residual features defined in Eq. 42-45 also reach the respective physics consistent values, $\phi_s[u_\theta]\left(x_f^i, t_f^i\right) \to 0,\ s = 1, \dots, 4,\ i = 1, \dots, N_f$. Moreover, the generated residual feature input in Eq. 46 approaches the same zero residual state used as the real discriminator input in Eq. 48. Consequently, the adversarial gradient defined in Eq. 91 provides a discriminator dependent pathway for guiding the generated residual features and thus the generator parameters toward target values as guided by multiple physics consistent states.

The $H^1$ has a slightly different role as the ideal residual state, $(\partial_x f_\theta)^2 = 0$, only requires the derivative of the residual to be spatially constant. This is useful in scenarios where the direct adversarial features are inactive and the $H^1$ discriminator allows for additional spatial feedback while the standard PDE loss continues enforcing the optimization of the residual in the main objective to minimize. Concurrently, the Gen-PINNs approach uses the zero PDE residual as the physical optimization target while simultaneously leveraging additional pathways for optimization through the multiple residual transformation and the learned discriminator dependent weighting. Therefore, the type of physics information that is supplied to the generator is controlled by the residual feature sensitivity while the effect of the contribution of that physics residual feature at each collocation point is controlled by the adversarial mismatch and the discriminator sensitivity.

## C.1. Effect of increased iterations on the convergence of PINN models (Fig. 16-17).

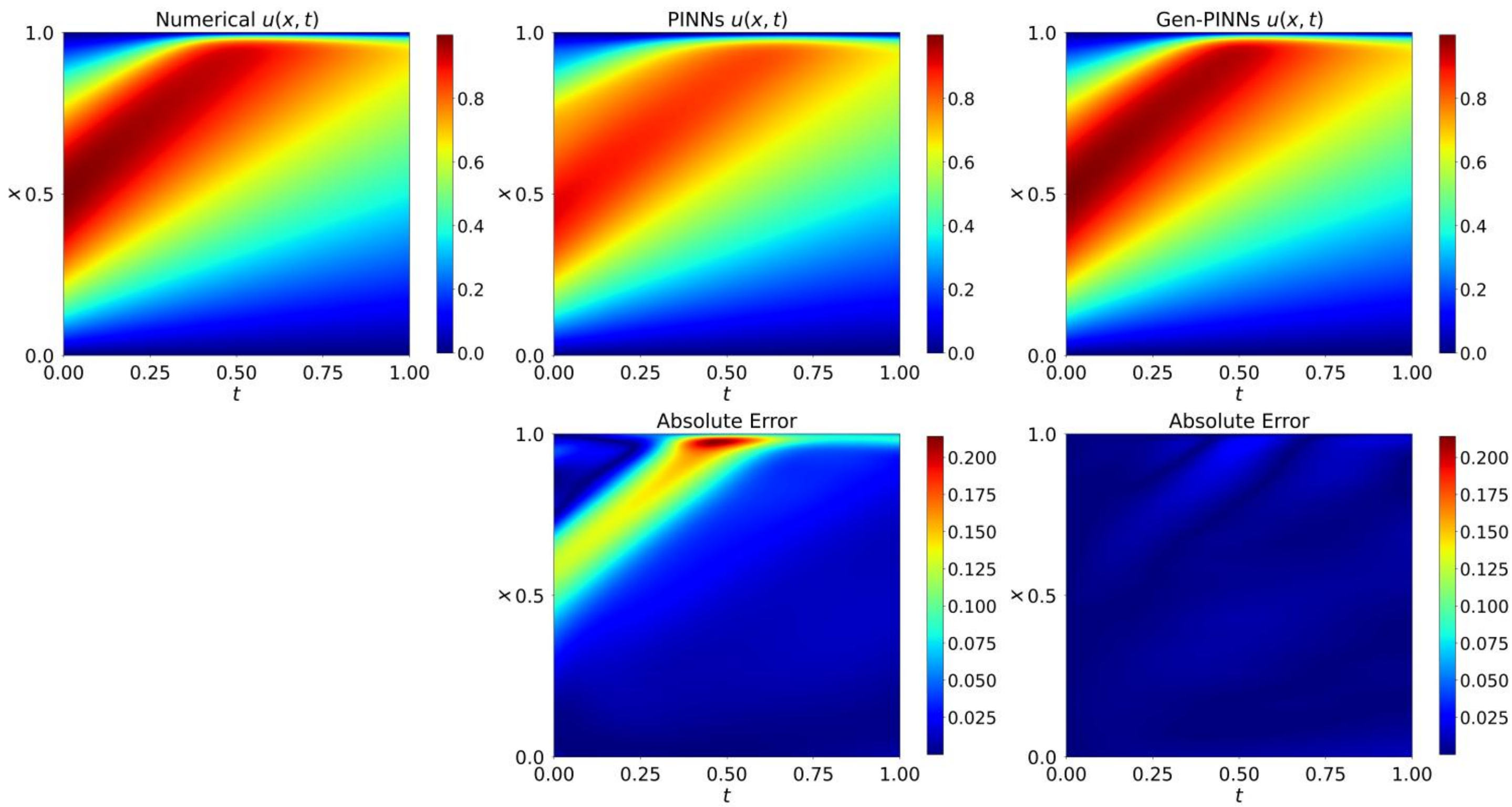

*Figure 16: Top Left: Numerical solution, Top Center: PINNs prediction, Top Right: Gen-PINNs prediction, Bottom Center: Absolute error in PINN prediction, Bottom Right: Absolute error in Gen-PINNs prediction for case 2 of the Burgers' PDE*

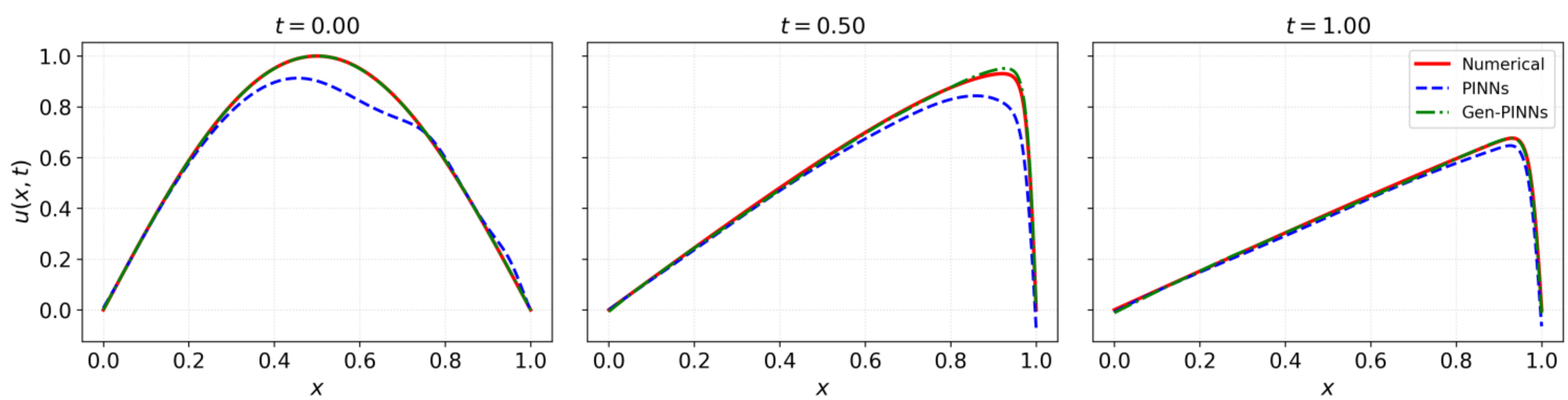

*Figure 17: Left: t=0, Center: t=0.5, Right: t=1 line plot comparison of the numerical solution (red), PINNs prediction (blue) and Gen-PINNs prediction (green) for case 2 of the Burgers' PDE*

## C.2. Effect of adding all physics components to the Adversarial Loss (Fig. 18-19)

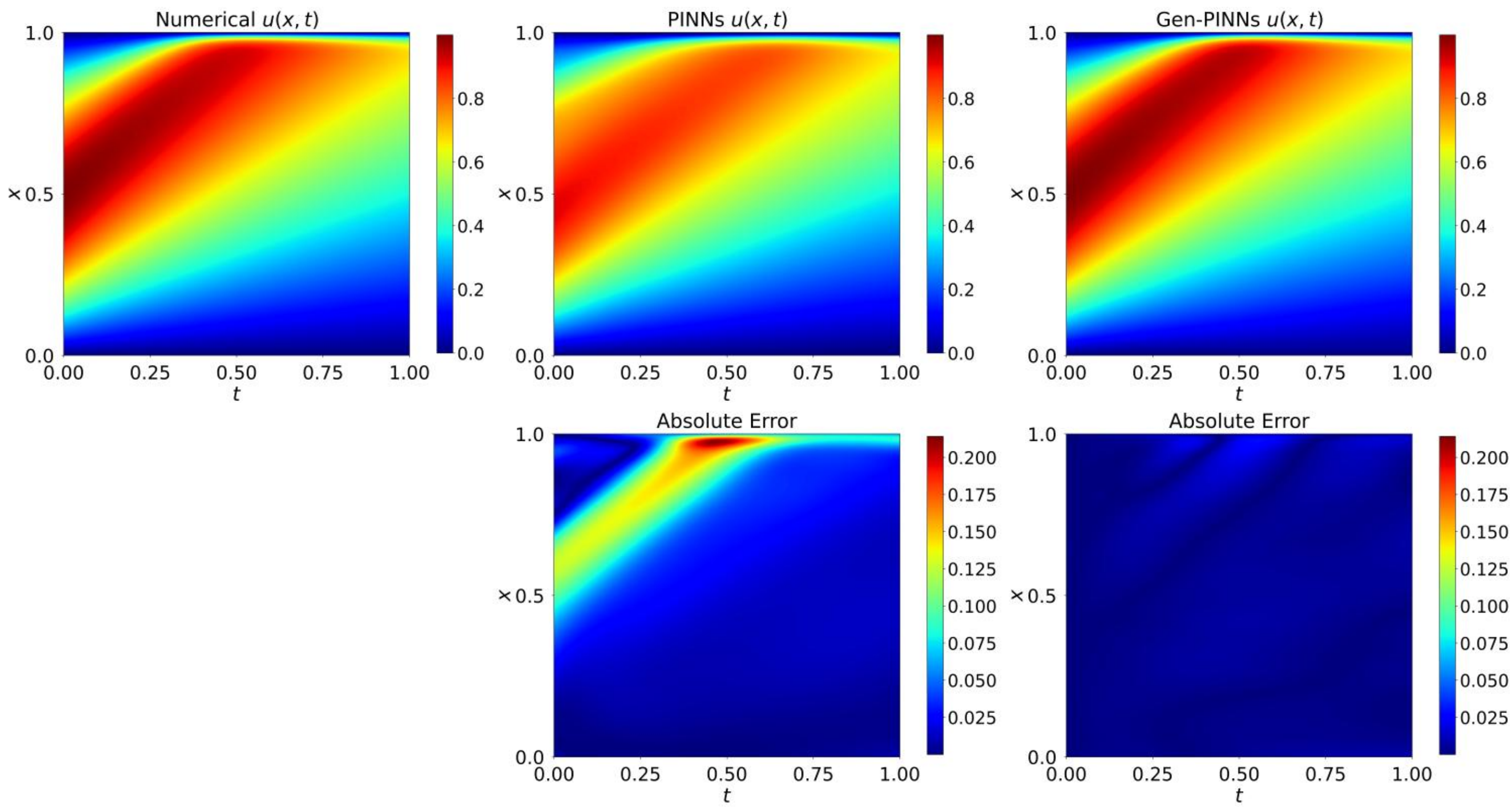

*Figure 18: Top Left: Numerical solution, Top Center: PINNs prediction, Top Right: Gen-PINNs prediction, Bottom Center: Absolute error in PINN prediction, Bottom Right: Absolute error in Gen-PINNs prediction for case 2 of the Burgers' PDE*

.

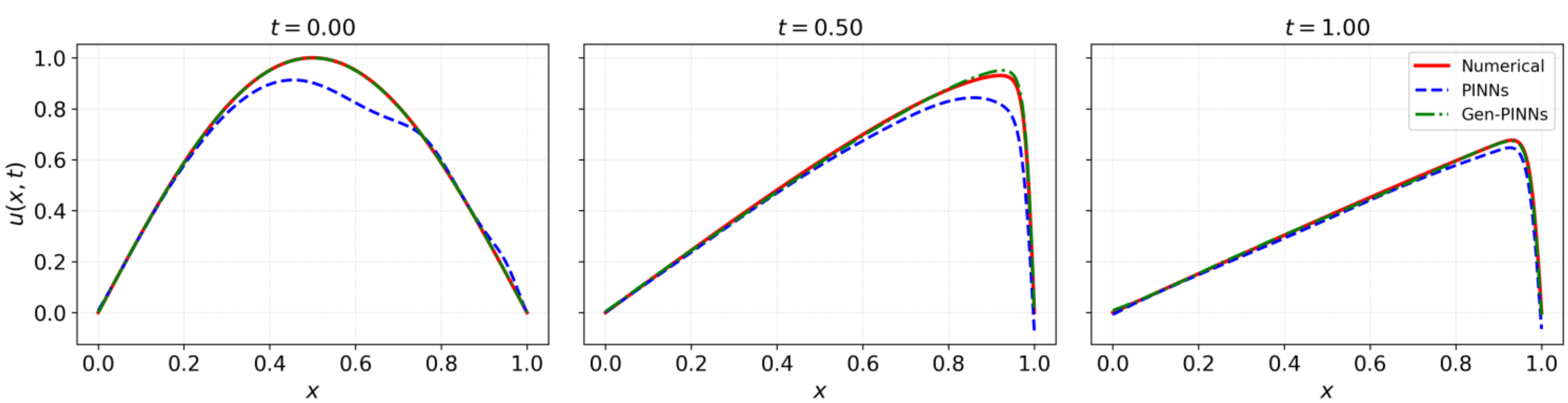

*Figure 19: Left: t=0, Center: t=0.5, Right: t=1 line plot comparison of the numerical solution (red), PINNs prediction (blue) and Gen-PINNs prediction (green) for case 2 of the Burgers' PDE*

D.1. Causal PDE weights turned on. The standard PINNs records an error of 2.02% while Gen-PINNs in this configuration records an error of 11% (Fig. 20-21).

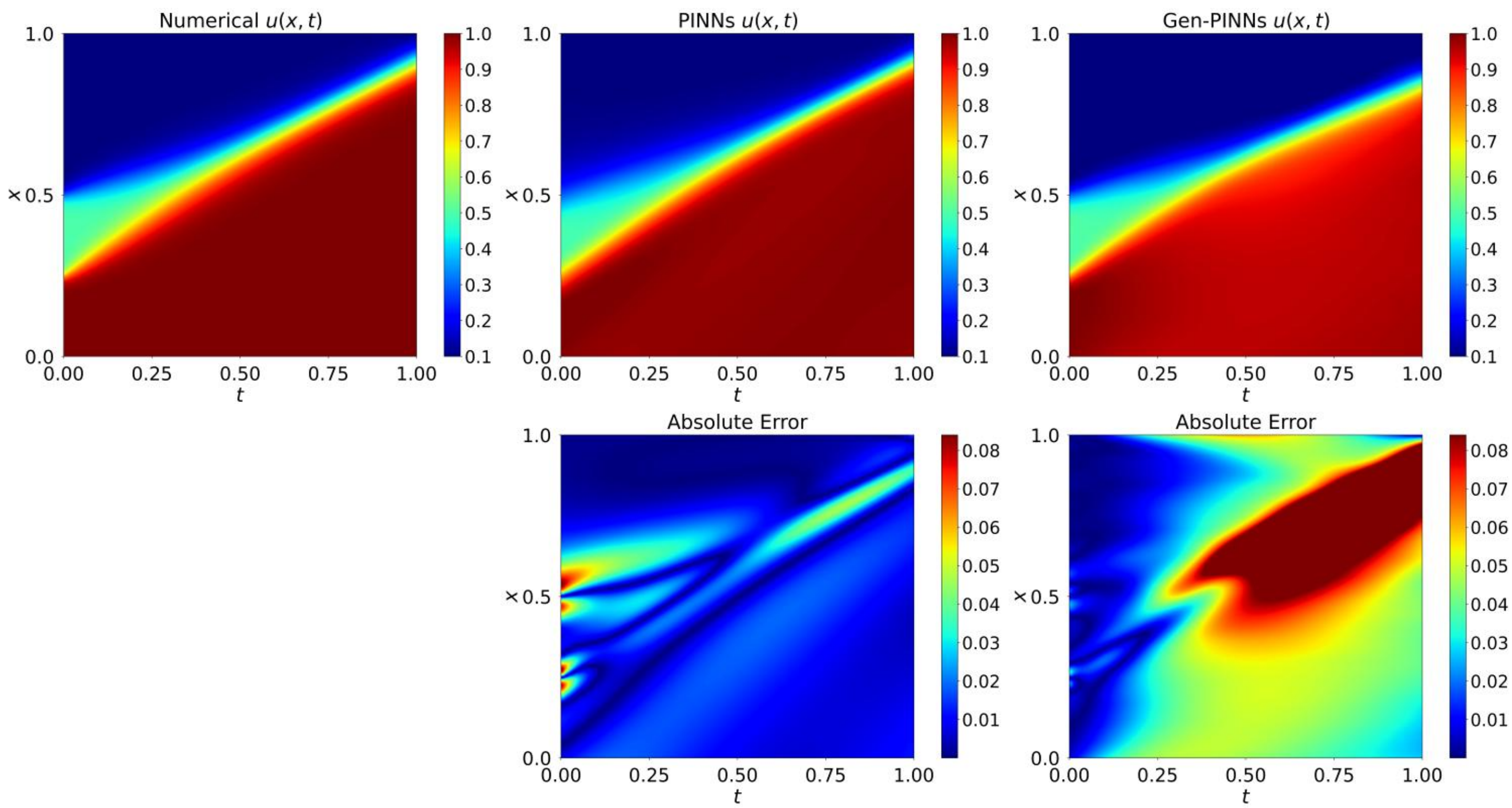

*Figure 20: Top Left: Numerical solution, Top Center: PINNs prediction, Top Right: Gen-PINNs prediction, Bottom Center: Absolute error in PINN prediction, Bottom Right: Absolute error in Gen-PINNs prediction for case 3 of the Burgers' PDE*

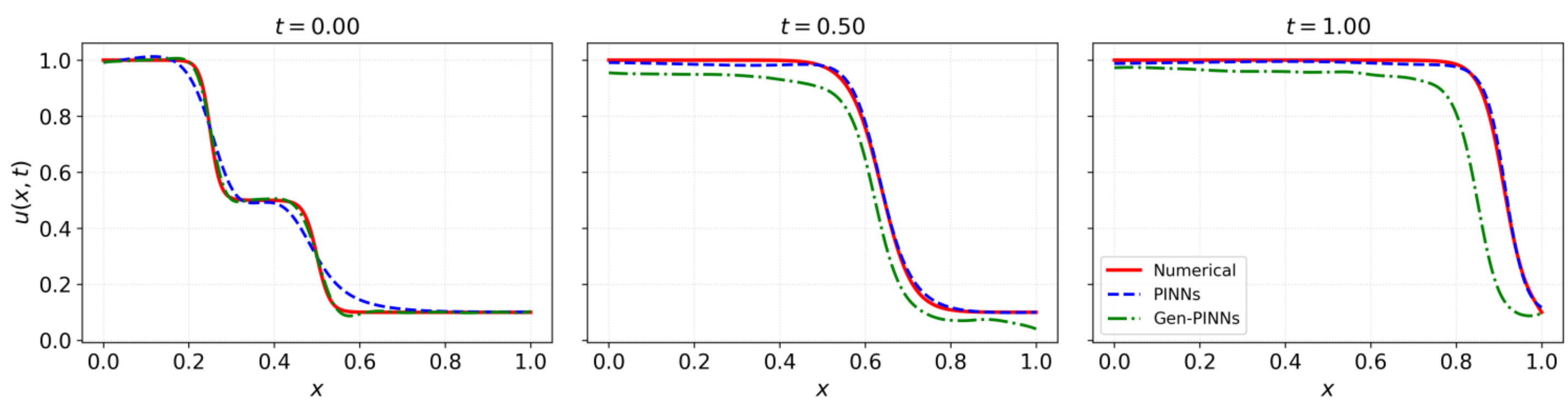

*Figure 21: Left: t=0, Center: t=0.5, Right: t=1 line plot comparison of the numerical solution (red), PINNs prediction (blue) and Gen-PINNs prediction (green) for case 3 of the Burgers' PDE*

D.2. The space averaged relative $L^2$ error, $\mathcal{E}_{REL-L^2}$, and relative PDE $L^2$ error, $\mathcal{E}_{\text{PDE}}$, are calculated using Eq. 92-93 for standard and Gen-PINNs. The errors are then plotted against the time. The plots show a large increase in the actual and PDE error over time for Gen-PINNs where the causal PDE loss is turned on while relatively steady error profiles for the standard PINNs model (Fig. 22). This indicated that the global PDE loss is sometimes sufficient for PDE shock front modeling specially in cases where the varying BCs impose a strong influence on the solution and the time-based causal effect is affected in both space and time.

$$\mathcal{E}_{REL-L^2}(t) = \frac{\sqrt{\frac{1}{N_x}\sum_{i=1}^{N_x} \left(u_\theta(x_i,t) - u(x_i,t)\right)^2}}{\sqrt{\frac{1}{N_x}\sum_{i=1}^{N_x} u(x_i,t)^2}} \tag{92}$$

$$\mathcal{E}_{\text{PDE}}(t) = \left(\frac{1}{N_x}\sum_{i=1}^{N_x} \left(f_\theta(x_i,t)\right)^2\right)^{1/2} \tag{93}$$

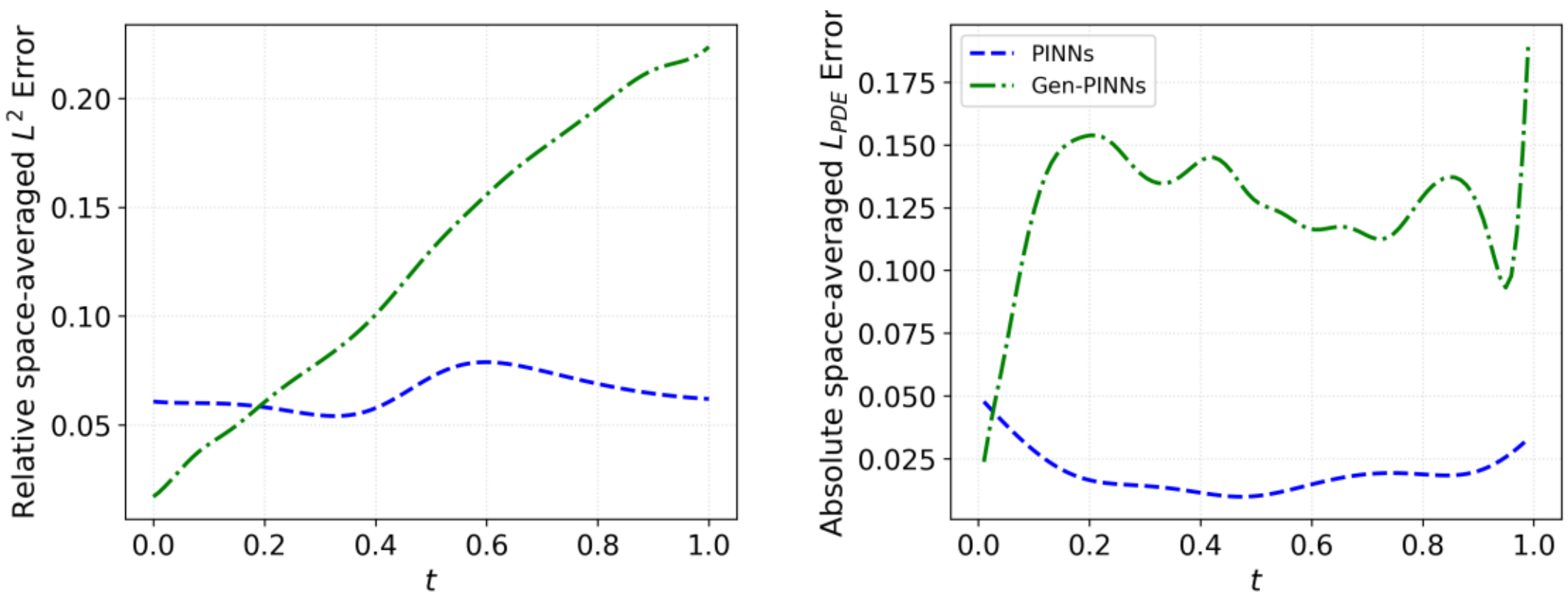

*Figure 22: Space-averaged error plots for standard PINNs (blue) and Gen-PINNs (green)*

D.3. The added learnable Fourier features in $x$ and $t$ significantly improve the match in Burgers PDE with non-standard boundary conditions. Expanding Eq. 36, the input features are now represented with random frequencies, $w^{(x)}$ and $w^{(t)}$ (Eq. 94).

$$\gamma_\theta(x,t) = \left[x, t, \sin\left(\omega_1^{(x)}x\right), \cos\left(\omega_1^{(x)}x\right), \dots, \sin\left(\omega_m^{(x)}x\right), \cos\left(\omega_m^{(x)}x\right)\right. \\ \left.\sin\left(\omega_1^{(t)}t\right), \cos\left(\omega_1^{(t)}t\right), \dots, \sin\left(\omega_m^{(t)}t\right), \cos\left(\omega_m^{(t)}t\right)\right]^{\mathsf{T}} \tag{94}$$

Following the sequence of feature mapping and induced kernel, Eq. 37-39, a slightly more flexible version of the Fourier input embedding expansion is reached where the kernel is now represented as

$$k_{\gamma_\theta} = x_1x_2 + t_1t_2 + \sum_{k=1}^{m} \cos\left(\omega_k^{(x)}(x_1 - x_2)\right) + \sum_{k=1}^{m} \cos\left(\omega_k^{(t)}(t_1 - t_2)\right) \tag{95}$$

Comparing Eq. 95 with Eq. 36, this approach adds an additional input data in the form of the embedded time dimension. Improvements in heuristic testing are noticed, and the approach is therefore used in the Burgers' 3rd case where non-standard BCs occur in the PDE solution. It is important to note that the mathematical implications of this implementation are beyond the scope of this work and are left for further studies.

E.1. Training Gen-PINNs using the direct PDE losses simultaneously, including the global PDE loss and the tertiary causal PDE loss, results in poor convergence. This is observed regardless of the initial value of direct PDE loss weights, $w_{PDE-EXP}^{base}$ and $w_{PDE}^{base}$, or if dynamic weighting is turned on. Regular space wise banding is observed in the absolute error plot where the causal loss seems to resolve errors in certain bands, while ignoring other bands resulting in noisy output at all time steps (Fig. 23-24). The global PDE loss, even if kept small, is unable to help the optimizer from converging. It is therefore recommended to only use the tertiary causal PDE loss in cases of strong PDE causality as the global PDE either fails to converge or results in optimization pathologies.

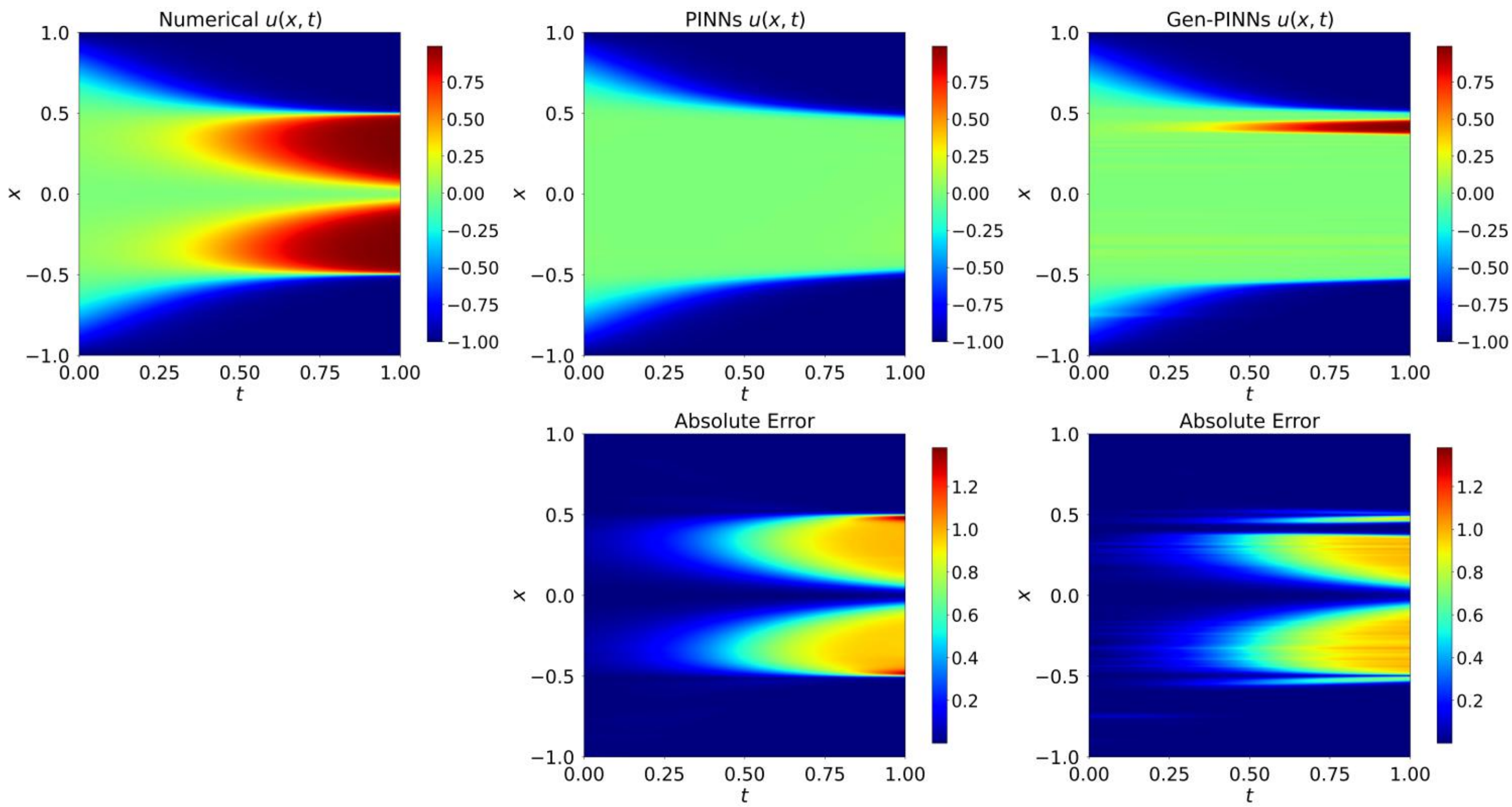

*Figure 23: Top Left: Numerical solution, Top Center: PINNs prediction, Top Right: Gen-PINNs prediction, Bottom Center: Absolute error in PINN prediction, Bottom Right: Absolute error in Gen-PINNs prediction for the Allen-Cahn PDE*

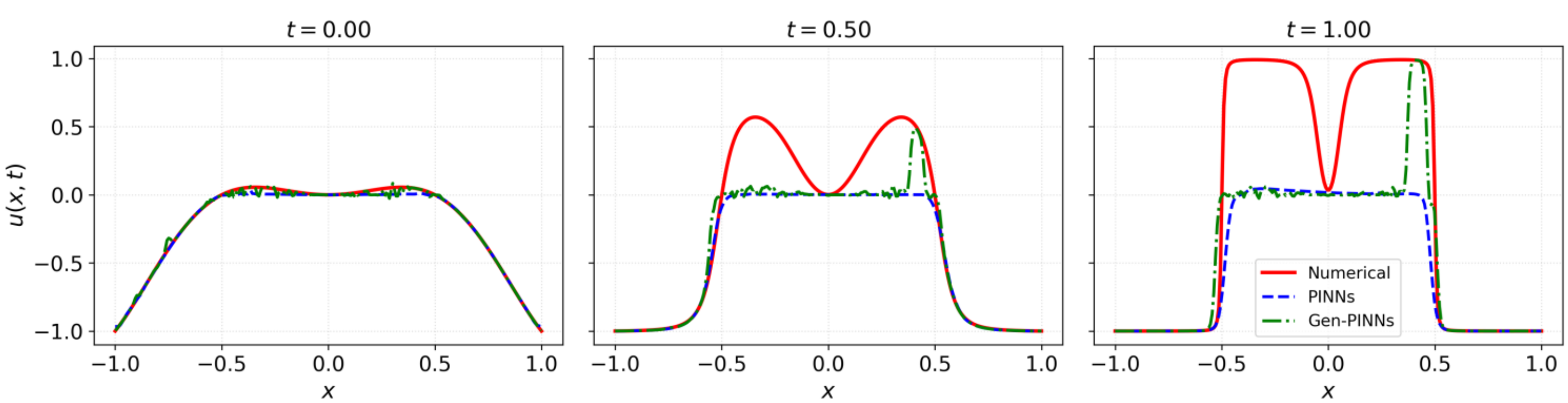

*Figure 24: Left: t=0, Center: t=0.5, Right: t=1 line plot comparison of the numerical solution (red), PINNs prediction (blue) and Gen-PINNs prediction (green) for the Allen-Cahn PDE*

E.2. Improvements in PDE modeling are observed when the global PDE loss is turned off (Fig. 25-26). However, with the unrestricted exponential PDE weighting it is posited that there are some unstable pathologies occurring during the training that prevents sufficient convergence in time. Using the exact same hyperparameters and changing the gradient pathway to use only weighted residuals in the graph reductions in error is observed.

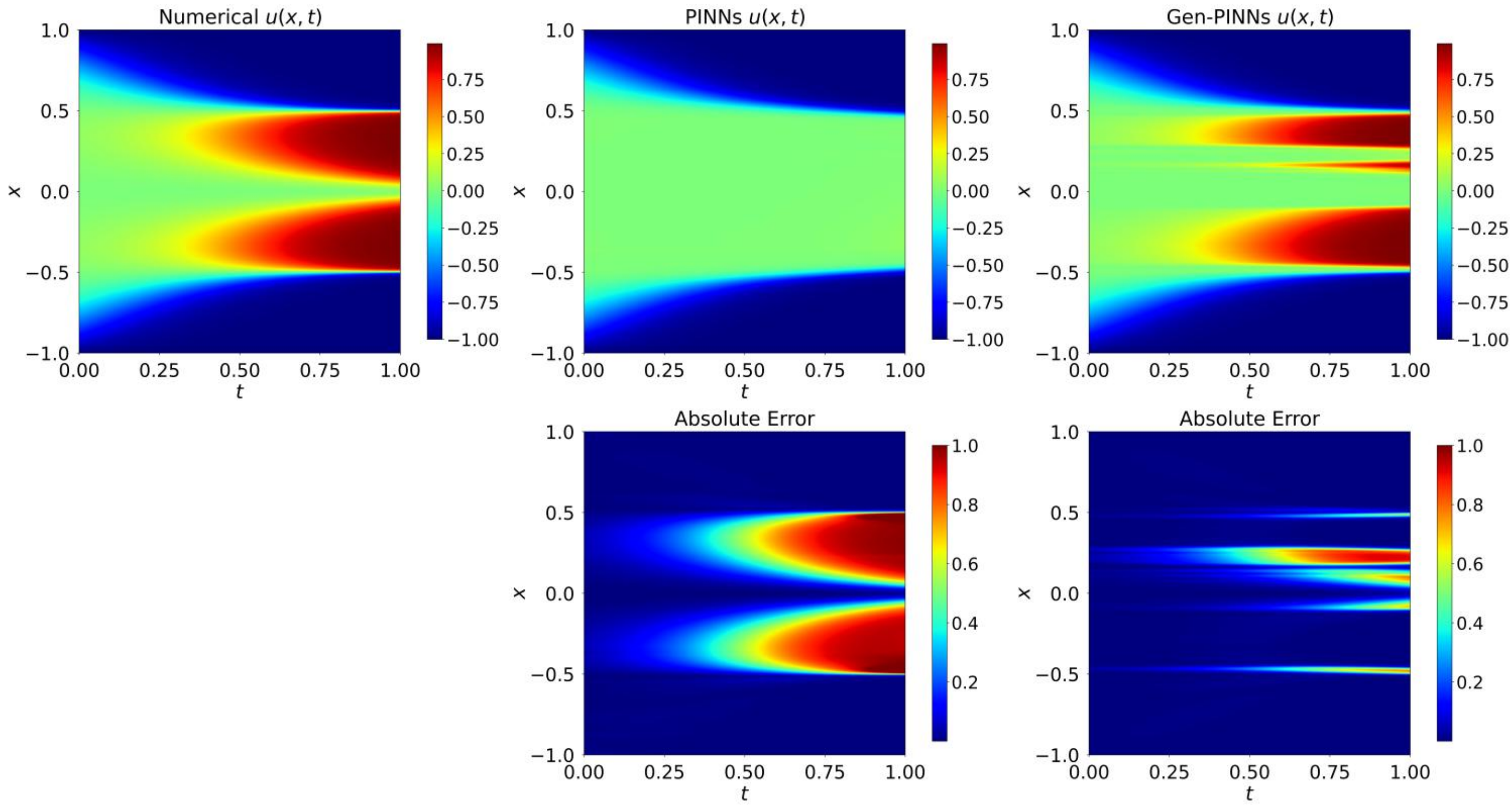

*Figure 25: Top Left: Numerical solution, Top Center: PINNs prediction, Top Right: Gen-PINNs prediction, Bottom Center: Absolute error in PINN prediction, Bottom Right: Absolute error in Gen-PINNs prediction for the Allen-Cahn PDE*

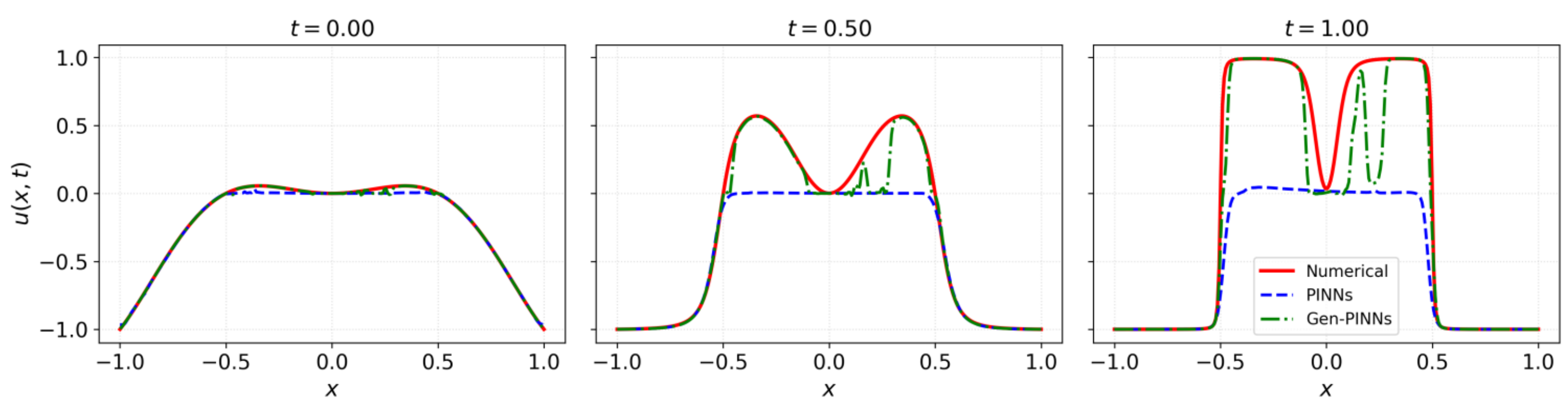

*Figure 26: Left: t=0, Center: t=0.5, Right: t=1 line plot comparison of the numerical solution (red), PINNs prediction (blue) and Gen-PINNs prediction (green) for the Allen-Cahn PDE*

E.3. Using the Stabilized Causal update scheme allows for a much improved convergence to the optimal solution (Fig. 27-28).

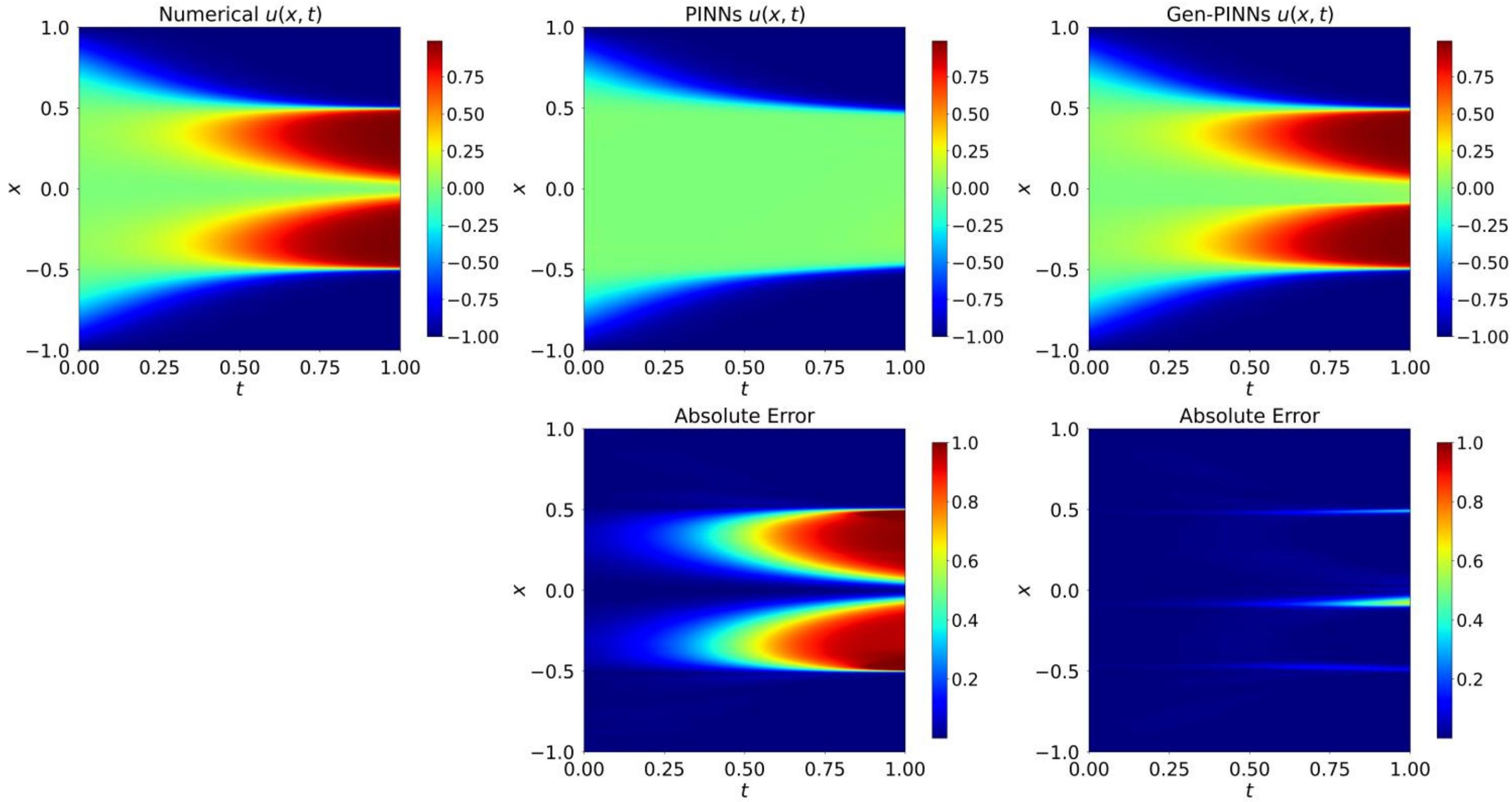

*Figure 27: Top Left: Numerical solution, Top Center: PINNs prediction, Top Right: Gen-PINNs prediction, Bottom Center: Absolute error in PINN prediction, Bottom Right: Absolute error in Gen-PINNs prediction for the Allen-Cahn PDE*

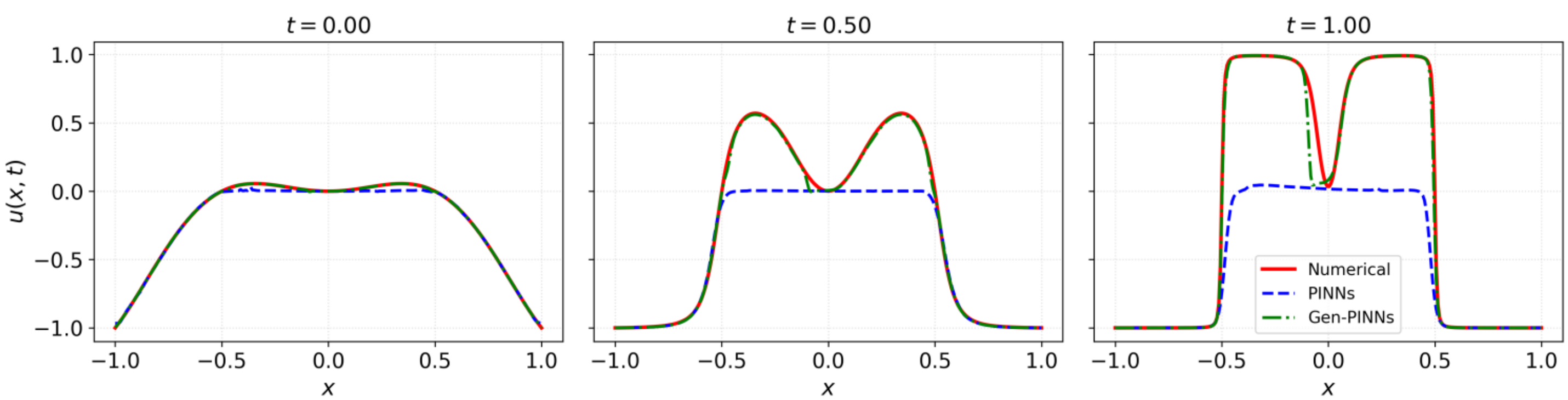

*Figure 28: Left: t=0, Center: t=0.5, Right: t=1 line plot comparison of the numerical solution (red), PINNs prediction (blue) and Gen-PINNs prediction (green) for the Allen-Cahn PDE*

F.1. Gen-PINNs solutions fails to propagate sufficiently in time if $\alpha_{\text{exp}}$ is set too high at $1 \times 10^2$ resulting in a cut-off in time with a relative $L^2$ error of 45.3% (Fig. 29). Repeating the training while keeping all hyperparameters constant and reducing $\alpha_{\text{exp}}$ to $1 \times 10^1$ results in an improved match with a relative $L^2$ error of 11.2% (Fig. 30). As Gen-PINNs is based on a single NN approach from the start to the end time of prediction, it is possible that improvements in modeling are possible by using a multi generator prediction structure as used in causal PINN extension in Wang et al. (2024). This case was run for 10000 epochs as increasing number of training iterations only slightly improved the solution.

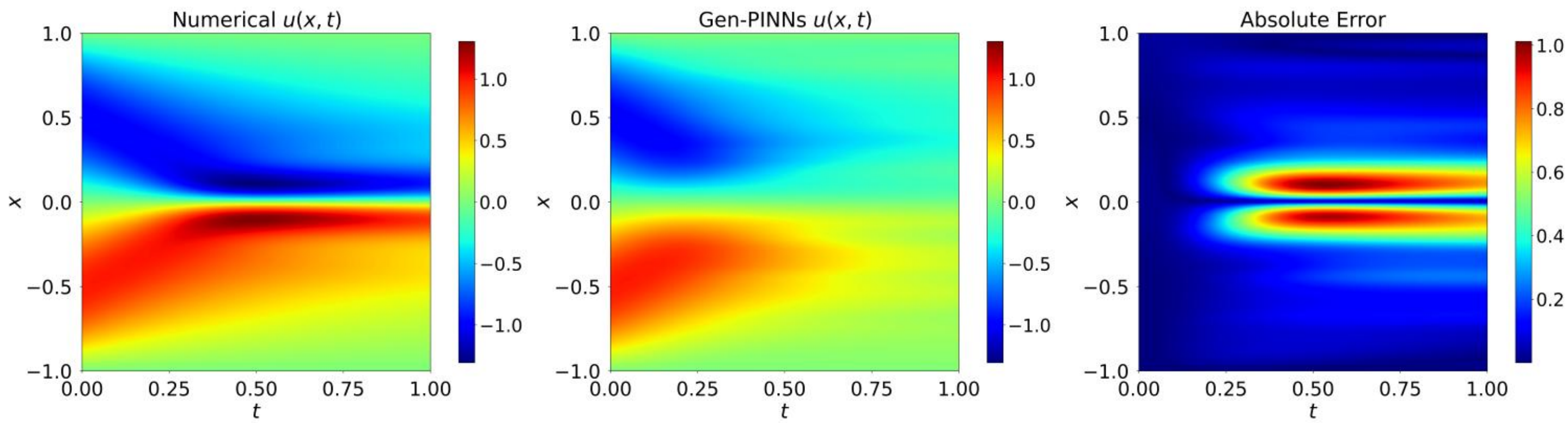

*Figure 29: Left: Numerical solution, Center: Gen-PINNs prediction, Right: Absolute Error plot with $\alpha_{exp} = 1 \times 10^2$*

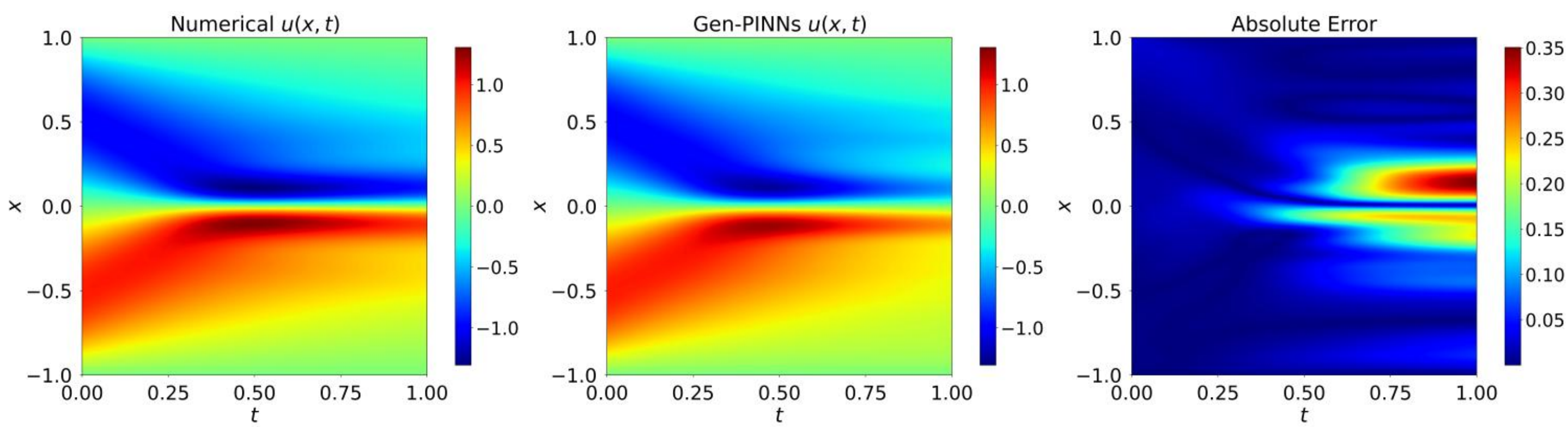

*Figure 30:Left: Numerical solution, Center: Gen-PINNs prediction, Right: Absolute Error plot with $\alpha_{exp} = 1 \times 10^1$*

## F.2. Hyperparameter analysis for KS PDE

To investigate the comparatively modest improvement for the KS equation, 15 configuration and sensitivity runs were extracted and examined from a few hundred runs and the representative results are summarized in Table 5.

*Table 5: Select KS configuration sensitivity runs*

| Trial* | Configuration changes | ADAM / L-BFGS | Relative $L^2$ Error |
|---|---|---|---|
| 1 | $n_{\text{f}} = 4$; $\alpha_{\text{VIMSE}} = 0.9$; $w_{\text{PDE-EXP}}^{\text{base}} = 10^1$; $w_{\text{PDE}}^{\text{base}} = 0$; $w_{\text{IC}}^{\text{base}} = 10^1$; $w_{\text{BC}}^{\text{base}} = 10^{-1}$; $w_{\text{ADV}}^{\text{base}} = 10^{-1}$; $\alpha_{\text{exp}} = 10^3$ | 10k / 10k | 66.60% |
| 2 | $n_{\text{f}} = 4$; $\alpha_{\text{VIMSE}} = 0.9$; $w_{\text{PDE-EXP}}^{\text{base}} = 10^2$; $w_{\text{PDE}}^{\text{base}} = 10^{-1}$; $w_{\text{IC}}^{\text{base}} = 10^{-1}$; $w_{\text{BC}}^{\text{base}} = 10^{-1}$; $w_{\text{ADV}}^{\text{base}} = 10^{-1}$; $\alpha_{\text{exp}} = 10^3$ | 9k / 1k | 62.48% |
| 3 | $n_{\text{f}} = 4$; $\alpha_{\text{VIMSE}} = 0.9$; $w_{\text{PDE-EXP}}^{\text{base}} = 10^{-1}$; $w_{\text{PDE}}^{\text{base}} = 0$; $w_{\text{IC}}^{\text{base}} = 10^2$; $w_{\text{BC}}^{\text{base}} = 10^{-1}$; $w_{\text{ADV}}^{\text{base}} = 0$; $\alpha_{\text{exp}} = 10^3$ | 9k / 1k | 56.53% |
| 4 | $n_\mathrm{f} = 4$; $\alpha_\mathrm{VIMSE} = 0.9$; $w_\mathrm{PDE-EXP}^\mathrm{base} = 10^0$; $w_\mathrm{PDE}^\mathrm{base} = 10^0$; $w_\mathrm{IC}^\mathrm{base} = 10^1$; $w_\mathrm{BC}^\mathrm{base} = 10^{-1}$; $w_\mathrm{ADV}^\mathrm{base} = 10^{-1}$; $\alpha_\mathrm{exp} = 10^3$ | 30k / 50k | 53.11% |
| 5 | $n_\mathrm{f} = 4$; $\alpha_\mathrm{VIMSE} = 0.9$; $w_\mathrm{PDE-EXP}^\mathrm{base} = 10^{-1}$; $w_\mathrm{PDE}^\mathrm{base} = 0$; $w_\mathrm{IC}^\mathrm{base} = 10^2$; $w_\mathrm{BC}^\mathrm{base} = 10^{-1}$; $w_\mathrm{ADV}^\mathrm{base} = 10^{-1}$; $\alpha_\mathrm{exp} = 10^3$ | 80k / 20k | 40.10% |
| 6 | $n_\mathrm{f} = 4$; $\alpha_\mathrm{VIMSE} = 0.9$; $w_\mathrm{PDE-EXP}^\mathrm{base} = 10^0$; $w_\mathrm{PDE}^\mathrm{base} = 10^{-1}$; $w_\mathrm{IC}^\mathrm{base} = 10^0$; $w_\mathrm{BC}^\mathrm{base} = 10^{-1}$; $w_\mathrm{ADV}^\mathrm{base} = 10^{-1}$; $\alpha_\mathrm{exp} = 10^3$ | 9k / 1k | 33.02% |
| 7 | $n_\mathrm{f} = 8$; $\alpha_\mathrm{VIMSE} = 0.95$; $w_\mathrm{PDE-EXP}^\mathrm{base} = 10^4$; $w_\mathrm{PDE}^\mathrm{base} = 10^{-2}$; $w_\mathrm{IC}^\mathrm{base} = 10^1$; $w_\mathrm{BC}^\mathrm{base} = 10^{-1}$; $w_\mathrm{ADV}^\mathrm{base} = 10^{-1}$; $\alpha_\mathrm{exp} = 10^3$ | 9k / 1k | 25.79% |
| 8 | $n_\mathrm{f} = 4$; $\alpha_\mathrm{VIMSE} = 0.9$; $w_\mathrm{PDE-EXP}^\mathrm{base} = 10^{-1}$; $w_\mathrm{PDE}^\mathrm{base} = 10^{-1}$; $w_\mathrm{IC}^\mathrm{base} = 10^2$; $w_\mathrm{BC}^\mathrm{base} = 10^{-1}$; $w_\mathrm{ADV}^\mathrm{base} = 10^{-1}$; $\alpha_\mathrm{exp} = 10^3$ | 80k / 20k | 25.07% |
| 9 | $n_\mathrm{f} = 4$; $\alpha_\mathrm{VIMSE} = 0.9$; $w_\mathrm{PDE-EXP}^\mathrm{base} = 10^{-1}$; $w_\mathrm{PDE}^\mathrm{base} = 10^{-1}$; $w_\mathrm{IC}^\mathrm{base} = 10^2$; $w_\mathrm{BC}^\mathrm{base} = 10^{-1}$; $w_\mathrm{ADV}^\mathrm{base} = 10^{-1}$; $\alpha_\mathrm{exp} = 10^3$ | 80k / 20k | 12.66% |
| 10 | $n_\mathrm{f} = 10^1$; $\alpha_\mathrm{VIMSE} = 0.95$; $w_\mathrm{PDE-EXP}^\mathrm{base} = 10^2$; $w_\mathrm{PDE}^\mathrm{base} = 0$; $w_\mathrm{IC}^\mathrm{base} = 10^2$; $w_\mathrm{BC}^\mathrm{base} = 10^0$; $w_\mathrm{ADV}^\mathrm{base} = 10^{-2}$; $\alpha_\mathrm{exp} = 10^2$ | 20k / 10k | 7.28% |
| 11 | $n_\mathrm{f} = 4$; $\alpha_\mathrm{VIMSE} = 0.9$; $w_\mathrm{PDE-EXP}^\mathrm{base} = 10^1$; $w_\mathrm{PDE}^\mathrm{base} = 10^0$; $w_\mathrm{IC}^\mathrm{base} = 10^0$; $w_\mathrm{BC}^\mathrm{base} = 10^{-1}$; $w_\mathrm{ADV}^\mathrm{base} = 10^{-1}$; $\alpha_\mathrm{exp} = 10^2$ | 10k / 10k | 6.93% |
| 12 | $n_\mathrm{f} = 4$; $\alpha_\mathrm{VIMSE} = 0.9$; $w_\mathrm{PDE-EXP}^\mathrm{base} = 10^1$; $w_\mathrm{PDE}^\mathrm{base} = 10^0$; $w_\mathrm{IC}^\mathrm{base} = 10^0$; $w_\mathrm{BC}^\mathrm{base} = 10^{-1}$; $w_\mathrm{ADV}^\mathrm{base} = 10^{-1}$; $\alpha_\mathrm{exp} = 10^2$ | 10k / 10k | 6.93% |
| 13 | $n_\mathrm{f} = 4$; $\alpha_\mathrm{VIMSE} = 0.95$; $w_\mathrm{PDE-EXP}^\mathrm{base} = 0$; $w_\mathrm{PDE}^\mathrm{base} = 10^1$; $w_\mathrm{IC}^\mathrm{base} = 10^2$; $w_\mathrm{BC}^\mathrm{base} = 10^0$; $w_\mathrm{ADV}^\mathrm{base} = 10^{-2}$; $\alpha_\mathrm{exp} = 0$ | 1k / 0.5k | 3.01% |
| 14 | $n_\mathrm{f} = 8$; $\alpha_\mathrm{VIMSE} = 0.95$; $w_\mathrm{PDE-EXP}^\mathrm{base} = 0$; $w_\mathrm{PDE}^\mathrm{base} = 10^1$; $w_\mathrm{IC}^\mathrm{base} = 10^2$; $w_\mathrm{BC}^\mathrm{base} = 10^0$; $w_\mathrm{ADV}^\mathrm{base} = 10^{-2}$; $\alpha_\mathrm{exp} = 0$ | 9k / 0 | 2.61% |
| 15 | $n_\mathrm{f} = 8$; $\alpha_\mathrm{VIMSE} = 0.99$; $w_\mathrm{PDE-EXP}^\mathrm{base} = 0$; $w_\mathrm{PDE}^\mathrm{base} = 10^1$; $w_\mathrm{IC}^\mathrm{base} = 10^2$; $w_\mathrm{BC}^\mathrm{base} = 10^0$; $w_\mathrm{ADV}^\mathrm{base} = 10^{-2}$; $\alpha_\mathrm{exp} = 0$ | 90k / 10k | 2.52% |

**The trials summarize selected configurations from the broader KS hyperparameter grid search without including intermediate sensitivity tests. Multiple settings may thus vary between Trials listed here.*

The evaluations identify a consistent configuration trend as expected of the contributions in overcoming PINN pathologies using the various components. The strongest solutions are obtained when the temporally weighted PDE objective is removed. A potential reason is the 4$^{\mathrm{th}}$ order derivative in this PDE which the current Gen-PINNs poorly represents in later time. The Burger case 1 with a similar PDE solution pattern has a much lower relative error when trained for longer times as it is only second order. Furthermore it is observed that final accuracy is obtained in considerably shorter runs with well balanced configurations. The final Gen-PINNs model reaches a relative error of 2.52%, compared with 2.93% for standard PINNs.